\documentclass{ecai} 
\usepackage{subfiles}
\usepackage{latexsym}
\usepackage{amssymb}
\usepackage{amsmath}
\usepackage{amsthm}
\usepackage{booktabs}
\usepackage{enumitem}
\usepackage{graphicx}
\usepackage{color}
\usepackage{times}
\usepackage{soul}
\usepackage{url}
\usepackage[hidelinks]{hyperref}
\usepackage[utf8]{inputenc}
\usepackage[small]{caption}
\usepackage{algorithm}
\usepackage{subfigure}
\usepackage{algorithmic}
\usepackage[switch]{lineno}
\newcommand{\BibTeX}{B\kern-.05em{\sc i\kern-.025em b}\kern-.08em\TeX}

\def\argmax{\displaystyle\operatornamewithlimits{arg\,max}}
\def\argmin{\displaystyle\operatornamewithlimits{arg\,min}}
\begin{document}

%%%%%%%%%%%%%%%%%%%%%%%%%%%%%%%%%%%%%%%%%%%%%%%%%%%%%%%%%%%%%%%%%%%%%%%%

\begin{frontmatter}

%%% Use this command to specify your submission number.
%%% In doubleblind mode, it will be printed on the first page.

\paperid{948} 

%%% Use this command to specify the title of your paper.

\title{Cluster Exploration using Informative Manifold Projections}

%%% Use this combinations of commands to specify all authors of your 
%%% paper. Use \fnms{} and \snm{} to indicate everyone's first names 
%%% and surname. This will help the publisher with indexing the 
%%% proceedings. Please use a reasonable approximation in case your 
%%% name does not neatly split into "first names" and "surname".
%%% Specifying your ORCID digital identifier is optional. 
%%% Use the \thanks{} command to indicate one or more corresponding 
%%% authors and their email address(es). If so desired, you can specify
%%% author contributions using the \footnote{} command.

\author[A]{\fnms{Stavros}~\snm{Gerolymatos}\thanks{Corresponding Author. Email: s.gerolymatos@liverpool.ac.uk.}}
\author[B,C]{\fnms{Xenophon}~\snm{Evangelopoulos}}
\author[A]{\fnms{Vladimir V.}~\snm{Gusev}} 
\author[A]{\fnms{John Y.}~\snm{Goulermas}}

\address[A]{Department of Computer Science, University of Liverpool, Liverpool, UK}
\address[B]{Department of Chemistry, University of Liverpool, Liverpool, UK}
\address[C]{Leverhulme Research Centre for Functional Materials Design, University of Liverpool, Liverpool, UK}
\address{{\{s.gerolymatos, evangx, vladimir.gusev, j.y.goulermas\}@liverpool.ac.uk}}

%%% Use this environment to include an abstract of your paper.

\begin{abstract}
Dimensionality reduction (DR) is one of the key tools for the visual exploration of high-dimensional data and uncovering its cluster structure in two- or three-dimensional spaces. The vast majority of DR methods in the literature do not take into account any prior knowledge a practitioner may have regarding the dataset under consideration. We propose a novel method to generate informative embeddings which not only factor out the structure associated with different kinds of prior knowledge but also aim to reveal any remaining underlying structure. To achieve this, we employ a linear combination of two objectives: firstly, contrastive PCA that discounts the structure associated with the prior information, and secondly, kurtosis projection pursuit which ensures meaningful data separation in the obtained embeddings. We formulate this task as a manifold optimization problem and validate it empirically across a variety of datasets considering three distinct types of prior knowledge. Lastly, we provide an automated framework to perform iterative visual exploration of high-dimensional data.
\end{abstract}

\end{frontmatter}

%%%%%%%%%%%%%%%%%%%%%%%%%%%%%%%%%%%%%%%%%%%%%%%%%%%%%%%%%%%%%%%%%%%%%%%%

\section{Introduction}

\label{Introduction}
Data exploration focuses on identifying informative patterns to discover new insight and knowledge about a collection of data. The often high-dimensional nature of such data renders the visual exploration process intractable for the human eye, and therefore specialized data manipulation of the original samples is essential in practice. Dimensionality reduction methods have been at the forefront of this challenge \cite{10.5555/1162264} aiming to recover lower-dimensional embeddings of the original data that facilitate the identification of underlying data cohorts and help understand better the problem at hand.

One of the most well known dimensionality reduction approaches perhaps is principal component analysis (PCA) \cite{Hotelling1933AnalysisOA}, an efficient linear method aiming to maximizing the variance along the projection vectors, which in practice appears insufficient for meaningful separation of cohorts. A variety of non-linear methods have also been proposed that conversely focus on locally preserving the structure of the data such as Isomap \cite{tenenbaum_global_2000}, LLE \cite{article}, t-SNE \cite{tsne}, UMAP \cite{UMAP}, TriMap \cite{2019TRIMAP} and LargeVis \cite{Tang_2016}, etc. Projection pursuit (PP) \cite{pp} defines a family of dimensionality reduction methods that can enable various embedding effects depending on a suitably selected criterion. The kurtosis index \cite{934071} is one specific PP example that specializes in identifying ``interesting'' projections. Its minimization particularly penalizes the normality of the data distribution, promoting thus more meaningful separability when searching for clusters. Other approaches that promote non-normality (non-Gaussianity) include negentropy and mutual information~\cite{Hyvrinen2000IndependentCA} which belong in the broader family of independent component analysis methodologies.

The above approaches nevertheless share the same attribute of offering a single static projection that does not consider any prior knowledge a practitioner may have regarding the high-dimensional latent structure. Such projections can be uninformative as they tend to illustrate the trends of the most evident features which are often already known by the reader. In practise, it has been shown \cite{10.1109/TVCG.2018.2864477} that an interactive or dynamic exploration of the available data can capture better their high-dimensional structure, especially when knowledge from a continual analysis of the data or cohort distribution can be factored into the analysis. Recently, a number of similar methods were introduced to dynamically generate new embeddings guided by the user~\cite{Senanayake2019SelfOrganizingNG}. Data exploration tasks require studying the patterns and cluster structure of the less dominant features in a dataset. Contrastive PCA (cPCA) \cite{cpca} computes data projections along which the variance of high-dimensional data is maximized (as in PCA), while the variance of some selected data is minimized. As a result, patterns and variation of the selected data can be removed from the projections. cPCA can be particularly useful when certain subsets of the data convey high variations, which might be obscuring the formation of alternative meaningful patterns in the dataset and thus may hinder its proper exploration. 

In this work we focus on computing data projections which filter out any cluster structure associated with some, usually unwanted, prior data knowledge. This knowledge reflects the practitioner's information about the data and by discarding this information from the projections, we hope to uncover unknown patterns. To achieve this, we jointly optimize a projection pursuit index that promotes data separability together with a variance-based objective to remove unwanted background patterns. Our method is applicable in three different prior (background) knowledge case scenarios, where the prior knowledge is represented by a dataset whose structure we wish to remove from the obtained embeddings. Specifically, the cases include: 
\begin{itemize}
    \item \textbf{Attribute-based prior}. In this case, the prior data consist of a subset of attributes of the high-dimensional dataset. We want to obtain embeddings that reveal the structure associated with the remaining attributes by discounting the structure of the attributes of the prior dataset. A naive approach to achieve this would be to discard this subset of features from the dataset. However, removing any features from a dataset may not necessarily factor out their variation from the lower dimensional embeddings, as different features are often correlated \cite{ctSNE}.
    \item \textbf{Sample-based prior}. We wish to visually explore a complex dataset that consists of the combination of a reference dataset (e.g., FMNIST) and a background dataset (e.g., MNIST). The background data modify and/or corrupt the reference dataset and we want to obtain embeddings that remove the background structure and disclose the reference one.
    \item \textbf{Subset-based prior}. The prior dataset consists of a subset of the original high-dimensional samples which are known to be similar to each other. By removing the structure associated with these samples, we can learn embeddings that reveal the structure of the remaining data points. Iterative visual exploration can take place in this setting by repeatedly updating the prior to include the most distinct cluster of each obtained projection.
%    employing our proposed structure extraction and update framework. 
\end{itemize}

We specifically propose an efficient and effective optimization on the Stiefel manifold \cite{Stiefel1935}, which appears to empirically perform better compared to other related methods and helps circumvent data-related numerical issues. Our main contributions are the following:
\begin{itemize}
    \item A novel objective function which when optimised computes projections that factor out different types of prior knowledge while also revealing previously unknown underlying structure.
    \item Manifold optimization modeling of the complex loss function to achieve numerical stability and fast convergence to a desirable solution.
    \item An iterative framework that can be applied to produce multiple informative projections for the visual data exploration of high-dimensional data.
\end{itemize}

The rest of the paper is organised as follows. Section~\ref{Related Work} expands upon related literature and in Section~\ref{Methodology} we give a detailed description of our proposed method. Section~\ref{Experimental Setup} provides a quantitative and qualitative analysis of our results and comparison with related approaches. Finally, Section ~\ref{Conclusion}
concludes our work.

%%%%%%%%%%%%%%%%%%%%%%%%%%%%%%%%%%%%%%%%%%%%%%%%%%%%%%%%%%%%%%%%%%%%%%%%
\section{Related Work} 
\label{Related Work}
A few works on interactive DR have been proposed in the last few years. Contrastive PCA \cite{cpca} computes data projections which highlight the salient structure of some reference data while discarding the structure of some background data. Conditional t-SNE (ct-SNE) \cite{ctSNE} is a generalisation of t-SNE which considers some prior knowledge in order to construct informative 2D embeddings. Prior knowledge corresponds to some information that the user is already aware of and can be represented by a set of labels assigned to the data samples. The labels are either available before any analysis or can be inferred by clustering a set of embeddings. To discount the known factor of the labels, new embeddings are then generated which can provide insight on any underlying unknown structure. In similar fashion, Fair Neighbor Retrieval Visualizer (FAIR-NeRV) \cite{peltonen2023fair} computes low-dimensional embeddings which preserve the high-dimensional neighborhood relationships while discarding the reliance on some sensitive information. This information can be specific attributes or class labels and is removed from the embeddings which can then reveal some complementary structure.

Unlike ct-SNE and FAIR-NeRV, which produce non-linear embeddings, SIDE \cite{puol1} is a linear approach which takes as input some prior knowledge in terms of a background distribution of points. These points are known to be similar to each other either \textit{a priori} or after some analysis. Projections that promote the maximal difference between the data and the background distribution are then computed. Another recently published method \cite{JMLR:v22:19-364} allows the user to guide the examination procedure according to their own exploration interests. The users can formulate their prior knowledge as well as their specific interests in terms of relations among a subset of samples and a subset of attributes. These are then introduced to the model to compute the projections. 

Finding lower dimensional embeddings over matrix manifolds has recently become quite popular, mostly due to the flexibility the constraint-free manifold optimization offers~\cite{manopt_book}. The (compact) Stiefel manifold~\cite{Stiefel1935}, i.e., the set of all k-tuples of orthonormal vectors, defines a smooth manifold that is endowed with a tangent space with its own inner product, (can be linearized locally around every point) allowing for fast, unconstrained optimization. Matrix manifold optimization has been employed in various cluster exploration applications~\cite{theis2009,Huang2020}, as well as in general machine learning~\cite{cetingul2009,Smith2022}. %More recently, the Grassmann manifold has been employed for lower dimensional embedding of 3D point clouds~\cite{grassmann2021}.

 \section{Methodology} \label{Methodology}
In this section, we first exemplify the optimization details of our method. Secondly, we introduce a framework for structure extraction and knowledge update with which we can perform iterative visual exploration of high-dimensional data.

\subsection{Objective formulation and optimization}
Let $\mathbf{X} \in \mathbb{R}^{n \times d}$ be the column mean-centered data matrix with each row $\{\mathbf{x}_i\}_{i=1}^n$ corresponding to the coordinates of the $i^\textrm{th}$ data point. Our goal is to generate a set of low dimensional embeddings $\{\mathbf{q}_i\}_{i=1}^n \in \mathbb{R}^k$ (with $k \ll d$) that render meaningful data separation based on prior information to maximize data cohort informativeness. Kurtosis \cite{934071} is a simple and efficient measure of non-normality, which when minimised promotes meaningful separability that tends to reveal informative data cohorts within a dataset. For a set of data projections, the kurtosis is defined as:
\begin{equation}
\label{eq:kurtosis}
kurt(\mathbf{X}) = \frac{n\sum_{i=1}^{n} \big(\mathbf{v}^\top \mathbf{x}_i\mathbf{x}_i^\top\mathbf{v}\big)^2}{\big(\mathbf{v}^\top\mathbf{X}^\top\mathbf{X}\mathbf{v}\big)^2}, 
\end{equation}
where $\mathbf{v} \in \mathbb{R}^{d}$ is the projection vector. Due to its quartic form, the kurtosis index can have multiple local minima and thus its optimization is a challenging task. Constrained gradient-based methods that force the projection vectors to have unit length~\cite{Pena2001} have been used to optimize the index but showed poor scalability. An efficient quasi-power method~\cite{kpp-optimization,sparsekpp,reckpp} has been designed as an alternative to constrained approaches, which however is less stable in practice when the covariance matrix $\mathbf{X}^\top\mathbf{X}$ is singular. Dimensionality reduction of the original samples (e.g., with SVD) is therefore required so that kurtosis can be optimized on the newly embedded points and can often lead to weak representations. To avoid the singularity issue, we instead optimize kurtosis directly over the Stiefel manifold~\cite{Stiefel1935} $St(k,d) \triangleq \{ \mathbf{M} \in \mathbb{R}^{d\times k} : \mathbf{M}^\top\mathbf{M} = \mathbf{I} \}$ which ensures unit-length projections over its surface, where an unconstrained optimization can be realised efficiently using any gradient-based solver~\cite{manopt_book}. We further propose a combination of a kurtosis and a cPCA \cite{cpca} reconstruction loss. cPCA can effectively maintain data subsets of high-variance while discarding subsets of no interest that may be obscuring the formation of alternative meaningful patterns. Nevertheless, cPCA is not explicitly designed to provide low-dimensional projections that exhibit meaningful data segregation. As a result, it often happens that cPCA embeddings are not informative. By jointly optimizing kurtosis, improved data separation that can reveal underlying structure is ensured. Let us assume a set of  $\{\mathbf{y}_i\}_{i=1}^m$ data samples associated with our prior knowledge, organised in matrix format $\mathbf{Y} \in \mathbb{R}^{m \times d}$. We formulate the above requirements in the following optimization and term our proposed method as IMAPCE (Informative MAnifold Projections for Cluster Exploration). \\
\begin{align}
\label{eq:objfun}
 \mathbf{V}^\star = & \argmin_{\mathbf{V} \in \mathbb{R}^{d \times k}} \Biggl\{ f(\mathbf{V}) \triangleq \| \mathbf{X} - \mathbf{X}\mathbf{V}\mathbf{V}^\top \|_F^2 - \alpha \| \mathbf{Y} - \mathbf{Y}\mathbf{V}\mathbf{V}^\top \|_F^2 \nonumber \\
 & \left. + \mu n \sum_{i=1}^{n} \left[ \mathbf{x}_i^\top \mathbf{V} (\mathbf{V}^\top \mathbf{X}^\top \mathbf{X} \mathbf{V})^{-1} \mathbf{V}^\top \mathbf{x}_i \right]^2 \right\} \notag \\
 & \text{s.t.} \quad \mathbf{V}^\top \mathbf{V} = \mathbf{I},
\end{align}
where $\mu$ is a scaling parameter, $\alpha$ (as in cPCA) regulates the trade-off between having a high target variance and a low background data variance and $\mathbf{V} \in \mathbb{R}^{d \times k}$ with $k < d$ is the projection matrix to be computed. Setting $\alpha=0$ corresponds to assuming no prior information (and therefore no prior data). $\mathbf{X}, \mathbf{Y}$ are both standardized before the optimization takes place. The gradient of $f(\mathbf{V})$ is given as
\begin{align}
    \label{eq:grad}
    \nabla_\mathbf{V} f(\mathbf{V}) &= 2\big(\alpha\mathbf{Y}^\top\mathbf{Y} - \mathbf{X}^\top\mathbf{X} \big)\mathbf{V}  \notag\\
    & +4\mu n  \sum_{i=1}^{n} \big(\mathbf{x}_i^{\top} \mathbf{V}\mathbf{A}^{-1}\mathbf{V^\top}\mathbf{x}_i\big)\big[ (\mathbf{x}_i \mathbf{x}_i^\top)\mathbf{V}\mathbf{A}^{-1} \notag\\
    &- \big(\mathbf{X}^\top \mathbf{X} \big)\mathbf{V}\mathbf{A}^{-1} \big( \mathbf{V}^\top \mathbf{x}_i \mathbf{x}_i^\top\mathbf{V} \big) \mathbf{A}^{-1} \big],
\end{align}
where $\mathbf{A} = \mathbf{V}^\top\mathbf{X}^\top\mathbf{X}\mathbf{V}$. For visualization purposes we usually set $k=2,3$ and in practice therefore $\mathbf{A} \in \mathbb{R}^{k\times k}$ is of small size, rendering its condition number $\kappa(\mathbf{A}) = ||\mathbf{A}||||\mathbf{A}^{-1}||$ to be relatively small and well-conditioned. By jointly minimizing both objectives of Eq.~\eqref{eq:objfun} on the Stiefel manifold, we avoid any singularity issues and at the same time obtain more informative separability of cohorts as we will empirically demonstrate later in the experiments. The time complexity of IMAPCE is $\mathcal{O}(dn^2)$ which mainly owes to the retraction step in the manifold optimization.

\subsection{Iterative visual exploration}
In the case of a subset-based prior, we are \textit{a priori} aware that a subset of samples of some high-dimensional data are similar to each other (e.g., they share the same class) and we wish to explore the cluster structure of the remaining data points. To achieve this, we optimize Eq. \eqref{eq:objfun}, where the kurtosis term is calculated over the remaining (subset of unexplored) samples, defined as $\{\mathbf{z}_i\}_{i=1}^{(n-m)} = \{\mathbf{x}_i\}_{i=1}^n \backslash \{\mathbf{y}_i\}_{i=1}^m$. As a result, the obtained projections provide meaningful data separation of the unexplored samples and reveal their cluster structure.  
 
To extract the structure of the unexplored samples \textbf{Z}, we perform the clustering of their embeddings. While the obtained clusters unveil some previously unknown structure, some of them are not as informative as others. We argue that the cluster which is the most separated from the rest (noted as \textit{most distinct}), is the most informative as it is expected to have the greatest probability of consisting of very similar points. The distinctness of a cluster and its separability are two connected notions stemming from the definition of kurtosis and is naturally encouraged by the optimization. While other cluster characteristics such as shape or density could also be considered, we empirically observed that such characteristics are less reliable across different types of data and features, as opposed to separability.

After finding the most distinct cluster, we dynamically update our prior data \textbf{Y} to include the points of this cluster, while also removing them from the unexplored subset \textbf{Z}. Optimization of Eq.~\eqref{eq:objfun} takes then place to compute new 2D embeddings that exhibit data separation of the updated unexplored data where cluster extraction can again be performed. This process continues iteratively until no more data remain unexplored and can efficiently provide gradual exploration of the unknown underlying structure of high-dimensional data.

To cluster the embeddings we used the Bayesian infinite Gaussian mixture model \cite{rasmussen}, \cite{Anderson1991-ANDTAN-7}, \cite{Neal2000MarkovCS}, commonly referred to as Dirichlet Process Gaussian Mixture Model (DPGMM) as it does not require a predefined number of clusters (but rather a maximum cluster number) and also due to its performance quality. More information about the DPGMM is provided in Section \ref{Supp:DPGMM} of the Supplementary Material. 

Given the mean $\mathbf{m}_l$ and covariance matrix $\mathbf{C}_l$ for each cluster $l$, we can extract the most distinct cluster by calculating all pairwise distances of cluster centers and their respecive distributions using the Mahalanobis distance~\cite{mahalanobis1936generalized}
\begin{equation}
    \label{eq:mahalanobis}
    \delta_{lj} = \sqrt{(\mathbf{m}_j-\mathbf{m}_l)^\top\mathbf{C}_l^{-1}(\mathbf{m}_j-\mathbf{m}_l)}.
\end{equation}

To avoid selecting a small deal of outliers as a cluster, we define a minimum acceptable cluster size. All clusters with size less than that are discarded as outliers and their distances are not considered. From Eq.~\eqref{eq:mahalanobis} we define a symmetric pairwise distances matrix for all clusters as $D_{lj} = (\delta_{lj} + \delta_{jl})/2$ and the most distinct cluster is given by

\begin{equation}
    \label{eq:distinctcluster}
    c^\star = \argmax \mathbf{D}\mathbf{1},
\end{equation}
where $\mathbf{1}$ is the vector of all ones. As a special case, if only two clusters of acceptable size are detected, then both are chosen as most distinct. Algorithm~\ref{alg:imapce} outlines the major steps of our proposed iterative framework. 

\begin{algorithm}[tb]
\caption{Iterative Visual Exploration using IMAPCE}
\label{alg:imapce}
\begin{algorithmic}[1]
\REQUIRE original data $\bigl\{ \mathbf{x}_i \bigl\}_{i=1}^{n}$, prior data $\bigl\{ \mathbf{y}_i \bigl\}_{i=1}^{m}$, unexplored data  $\bigl\{ \mathbf{z}_i \bigl\}_{i=1}^{n-m}$, acceptable cluster size $s$ (for termination criterion)
\ENSURE Embeddings $\mathbf{Q}$.
\WHILE{ $(n-m)>s$ } 
\STATE $\mathbf{V}^\star =  \argmin_{\mathbf{V}\in St(k,d)} f(\mathbf{V})$ \qquad \qquad \qquad \qquad \qquad \textcolor{blue}{\# IMAPCE}
\STATE $\mathbf{Q} = \mathbf{Z}\mathbf{V}^\star$ \qquad \qquad \qquad \textcolor{blue}{\# Embeddings of unexplored data}
\STATE $\{\mathbf{m}\}_l^r,\{\mathbf{C}\}_l^r \sim DPGMM\big(\{\mathbf{q}_i\}_{i=1}^{n-m}\big)$ \qquad \quad \textcolor{blue}{\# Clustering}
\FOR{all cluster pairs $(l,j)$} 
\STATE $\delta_{lj} = \sqrt{(\mathbf{m}_j-\mathbf{m}_l)^\top\mathbf{C}_l^{-1}(\mathbf{m}_j-\mathbf{m}_l)}$
\STATE $D_{lj} = (\delta_{lj} + \delta_{jl})/2$
\ENDFOR
\STATE $c^\star = \argmax \mathbf{D}\mathbf{1}$ \qquad \quad \textcolor{blue}{\# Most distinct cluster calculation}
\STATE $\mathbf{X}_c = \{\mathbf{x}_i : \mathbf{x}_i \in c^\star \}$ \quad \textcolor{blue}{\# Points of the most distinct cluster}
\STATE $\mathbf{Y} = \mathbf{Y}\cup\mathbf{X}_c$ \qquad \quad \textcolor{blue}{\# Update of prior data and their size $m$}
\STATE $\mathbf{Z} = \mathbf{Z}\backslash\mathbf{X}_c$ \qquad \qquad \qquad \qquad \textcolor{blue}{\# Update of unexplored data}
\ENDWHILE
\end{algorithmic}
\end{algorithm}

\subsection{Hyperparameter tuning}
Hyperparamaters $\alpha$ and $\mu$ have to be chosen for optimizing Eq.~\eqref{eq:objfun}. We need to select $\alpha$ if there is any prior data (otherwise it is set to zero), while $\mu$ has to be selected regardless of the availabilty of prior data. As a rule of thumb, setting $\alpha = 1$ empirically provides embeddings with a desirable trade-off between high original data variance and low prior data variance. In practice, we observe that the kurtosis term is not greater than 50-100. After computing the cPCA reconstruction error of our original data, $\mu$ is selected as one or two orders of magnitude below the cPCA reconstruction error. In this way, kurtosis can influence but not dominate the optimization. To further study the effect of the kurtosis term on the performance of IMAPCE, we perform an ablation study (Section \ref{Supp:Abb} of the Supplementary Material) on the hyperparameter $\mu$ and demonstrate the effectiveness of the proposed empirical rule for choosing $\mu$ across different scenarios. Finally, parameter $s$ in iterative visual exploration denotes the minimum acceptable size of a cluster and is used for discarding outliers as well as stopping the exploration process. Its choice depends on the size and dimensionality of the original data.

\section{Experimental results}
\label{Experimental Setup}
To showcase that IMAPCE can efficiently factor out different types of prior knowledge, we ran experiments on several datasets for all previously mentioned prior knowledge types. We provide both quantitative and qualitative results and compare IMAPCE's  performance against similar embedding methods that promote some underlying structure of the data while also removing any variation associated with the prior knowledge, namely cPCA, ct-SNE and Fair-NeRV. 

For cPCA, different projection matrices are computed for a fixed user-selected number of $\alpha$'s (trade-off hyperparameter) and spectral clustering is applied to them. Subsequently, the projection matrix which corresponds to the cluster medoid is used to compute the projections. It is noteworthy that ct-SNE and Fair-NeRV require the input of explicit data labels, which form the prior information to be removed from the generated embeddings. Details about the hyperparameters' selection for ct-SNE and Fair-NeRV are provided in Section \ref{Supp:hypers} of the Supplement.

We implemented IMAPCE on Python and used Pymanopt \cite{JMLR:v17:16-177}, a Python toolbox for optimization on Riemannian manifolds. The DPGMM used for clustering is implemented via the Sci-kit learn library of Python \cite{scikit-learn}. Our implementation is provided in \url{https://github.com/StavGer/IMAPCE}. For cPCA, ct-SNE and Fair-NeRV we used the official implementations. Experimental results are organized according to different prior knowledge cases.

\subsection{Attribute-based prior}
The task in this case is to generate embeddings that factor out the structure of one or more selected attributes of the high-dimensional data and reveal the structure of the remaining attributes. We ran experiments using IMAPCE, ct-SNE, cPCA and Fair-NeRV and compare their performance on both synthetic and real-world data. 

\subsubsection{Synthetic data}
The synthetic dataset \cite{revisitedctsne} consists of 1,500 ten-dimensional points. All points are assigned to one of two clusters (with centers sampled from $\mathcal{N}(0,25)$) in the first four dimensions and one of three clusters (with centers sampled from $\mathcal{N}(0,1)$) in dimensions five and six. For each point we add noise from $\mathcal{N}(0,0.01)$. The last four dimensions correspond to samples from $\mathcal{N}(0,1)$. To run IMAPCE (we set $\alpha = 1$, $\mu =200$) and cPCA, we define as prior data the first four dimensions and wish to generate embeddings that reveal the complementary cluster structure of dimensions five and six. With the same goal, we implement ct-SNE and Fair-NeRV using the cluster labels of the first four dimensions as prior. 

The generated embeddings are shown in figures \ref{fig:ex1-a}-\ref{fig:ex1-d} where prior data information is encoded according to shape and complementary structure according to color. We observe that cPCA embeddings are clustered with respect to their shape, indicating that the structure of prior data is not removed. On the contrary, ct-SNE, Fair-NeRV and IMAPCE compute embeddings that factor out the prior information as there is mix of points with different shapes. Fair-NeRV and IMAPCE seem to achieve a slightly better separation with respect to the green and red cluster than ct-SNE, unveiling the complementary structure to a better extent.

The Normalised Laplacian Score (NLS) was proposed \cite{ctSNE} to quantify the presence of some prior label information on a set of embeddings. This score takes values in [0, 1] and measures the label homogeneity within a user-selected neighborhood in an embedding set. Higher NLS values indicate better prior information removal w.r.t some labels from the embeddings. To quantitatively compare how well the embeddings of all methods factor out the labels of dimensions one to four, we calculated the NLS with respect to these labels and for neighborhood sizes of 10, 20, \dots, 100. We report the average NLS over the neighborhood size in Table \ref{table:1}. IMAPCE achieved higher mean NLS (lower homogeneity) than cPCA and ct-SNE, indicating that it removes the prior information more effectively (as also observed qualitatively). We argue that the inclusion of the kurtosis term in IMAPCE achieves the meaningful data separation that reveals some unknown structure missed by cPCA. Fair-NeRV achieves a slightly bigger average NLS than IMAPCE, nevertheless at a higher computational cost as shown later in Section~\ref{subsec:sample}.

\subsubsection{UCI Adult data}
We sampled 1,000 data points from the UCI Adult dataset \cite{UCI_adult} which consists of six features. Age, education level and work hours per week are numeric while ethnicity (white/other), gender (male/female) and income ($>$ 50,000) are binary. 

\begin{figure*}[h]
\centering
\subfigure[cPCA ]{ \rotatebox{90}{\footnotesize \qquad Synthetic}\label{fig:ex1-a}\includegraphics[height=1.2in,width=1.2in]{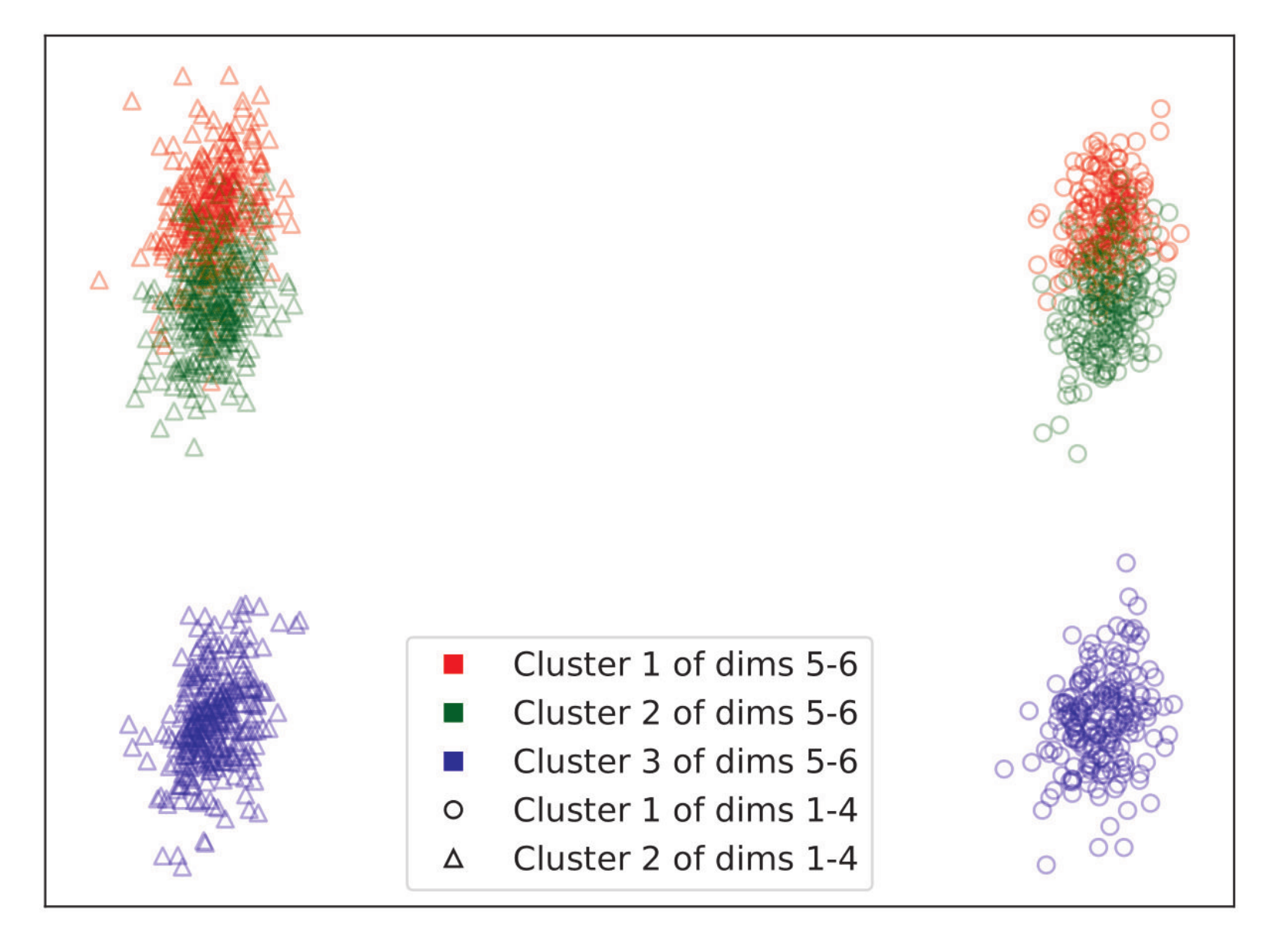}}
\subfigure[ct-SNE ]{\label{fig:ex1-b}\includegraphics[height=1.2in,width=1.2in]{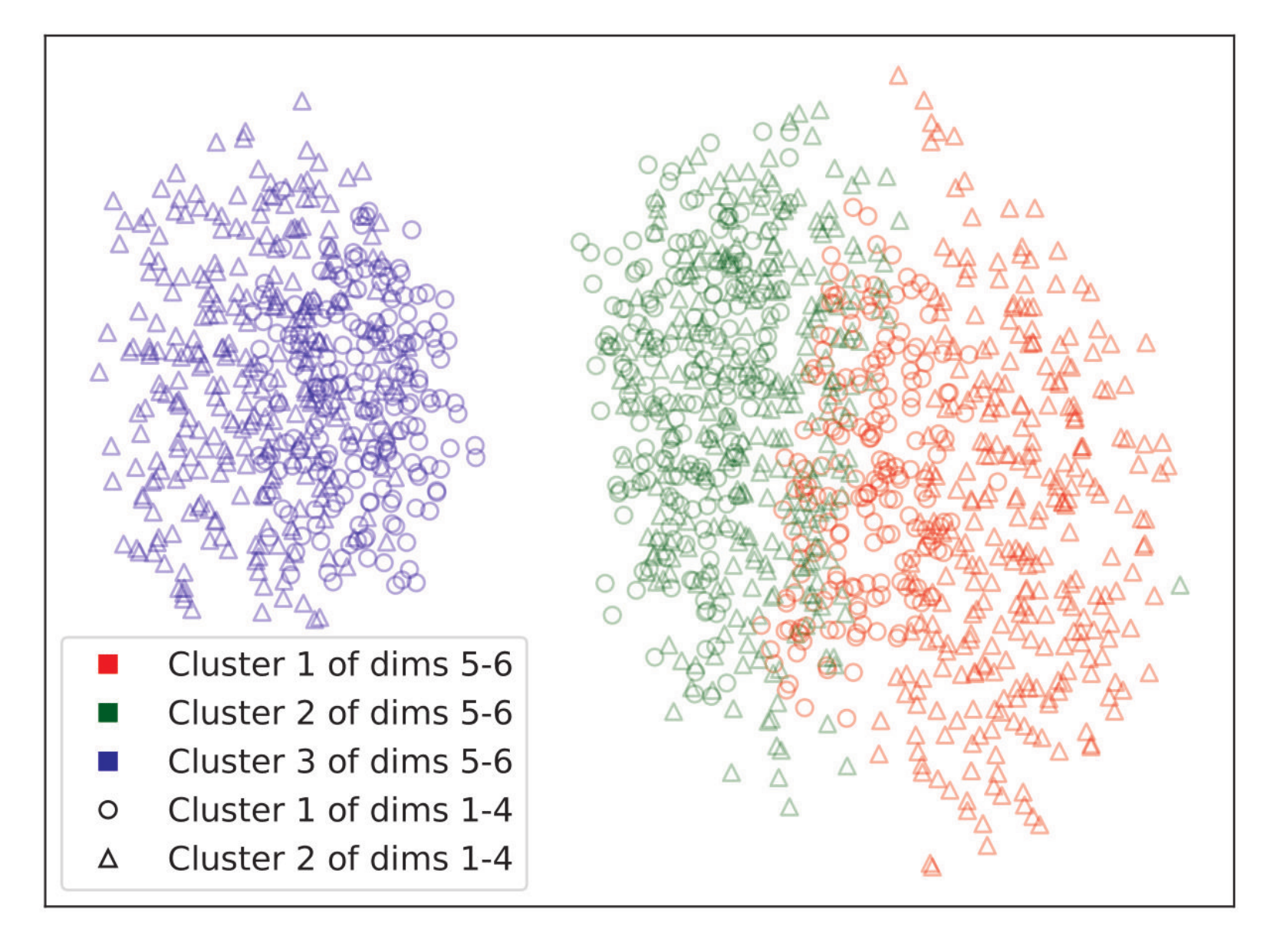}}
\subfigure[Fair-NeRV ]{\label{fig:ex1-c}\includegraphics[height=1.2in,width=1.2in]{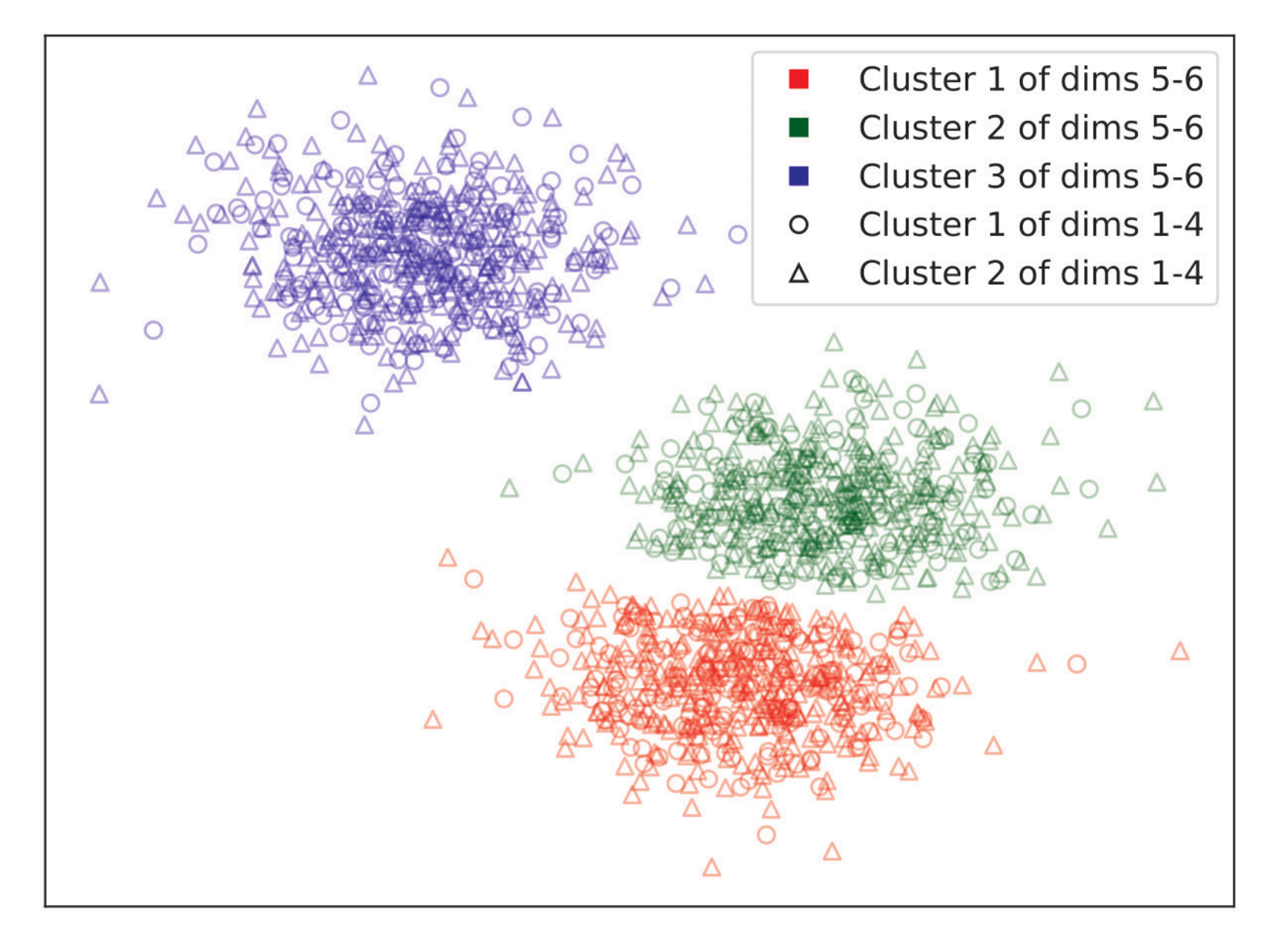}}
\subfigure[IMAPCE ]{\label{fig:ex1-d}\includegraphics[height=1.2in,width=1.2in]{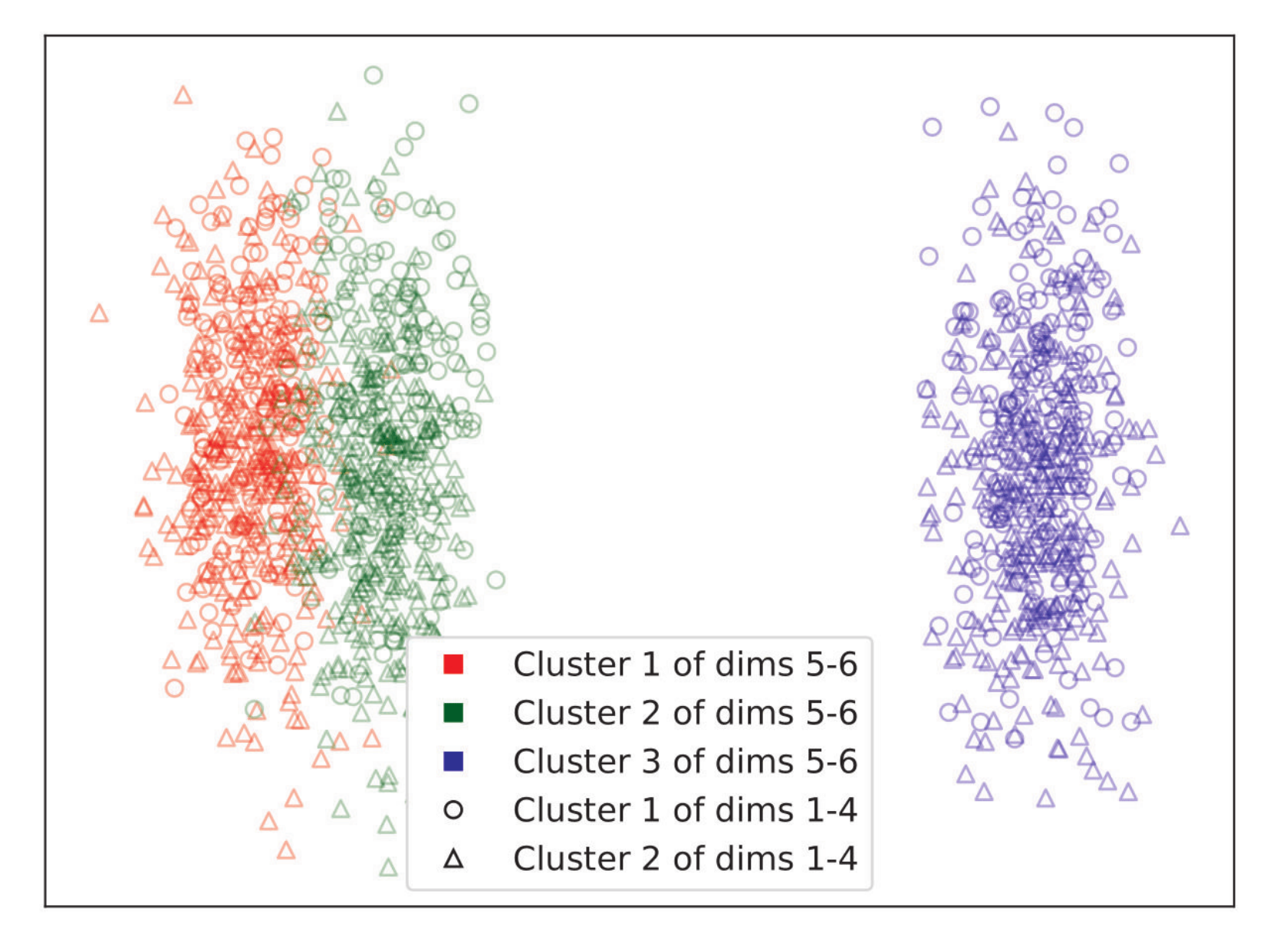}}\\
\subfigure[cPCA]{\rotatebox{90}{\footnotesize \qquad UCI Adult}\label{fig:ex2-a}\includegraphics[height=1.2in,width=1.2in]{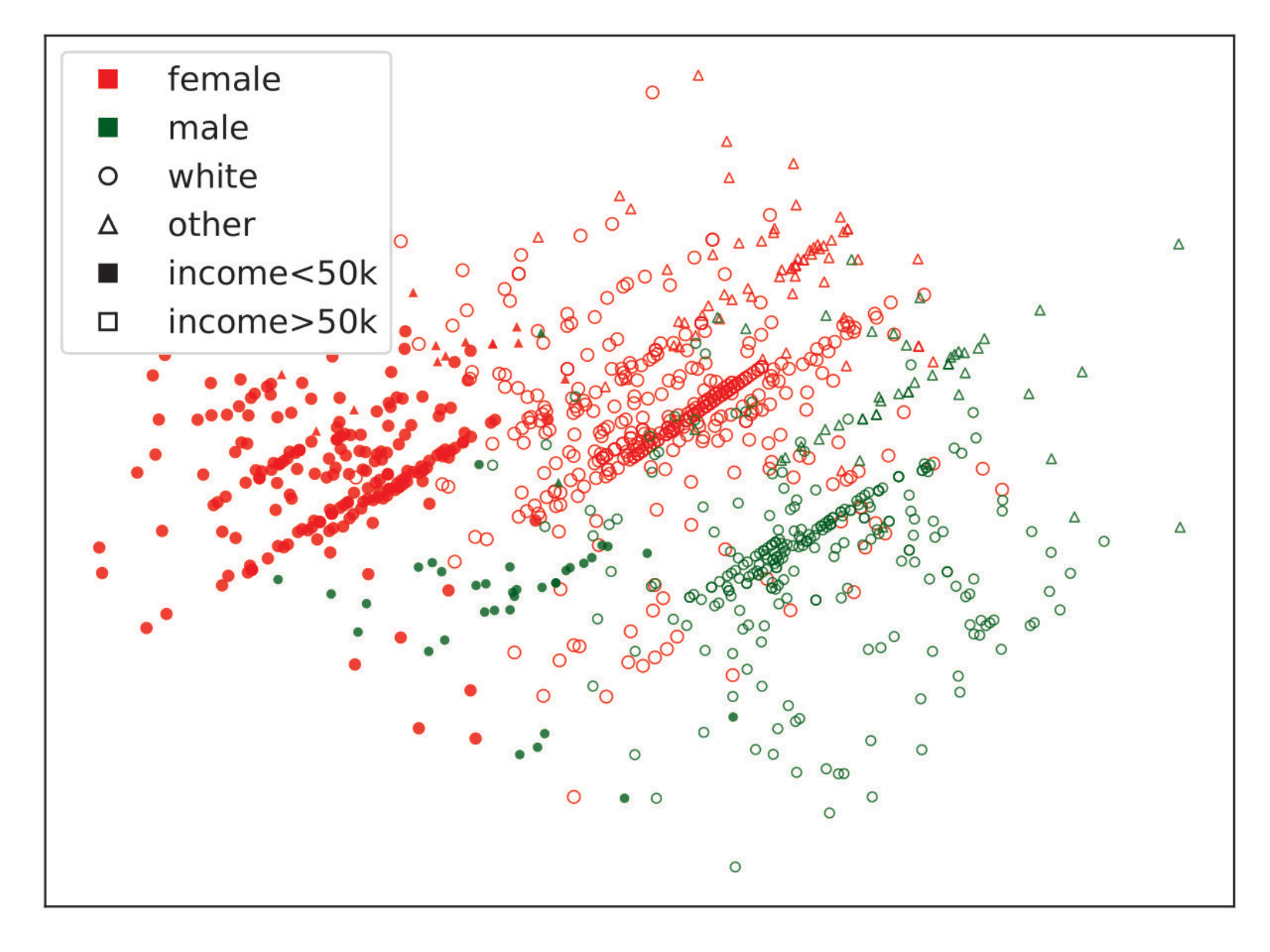}}
\subfigure[ct-SNE]{\label{fig:ex2-b}\includegraphics[height=1.2in,width=1.2in]{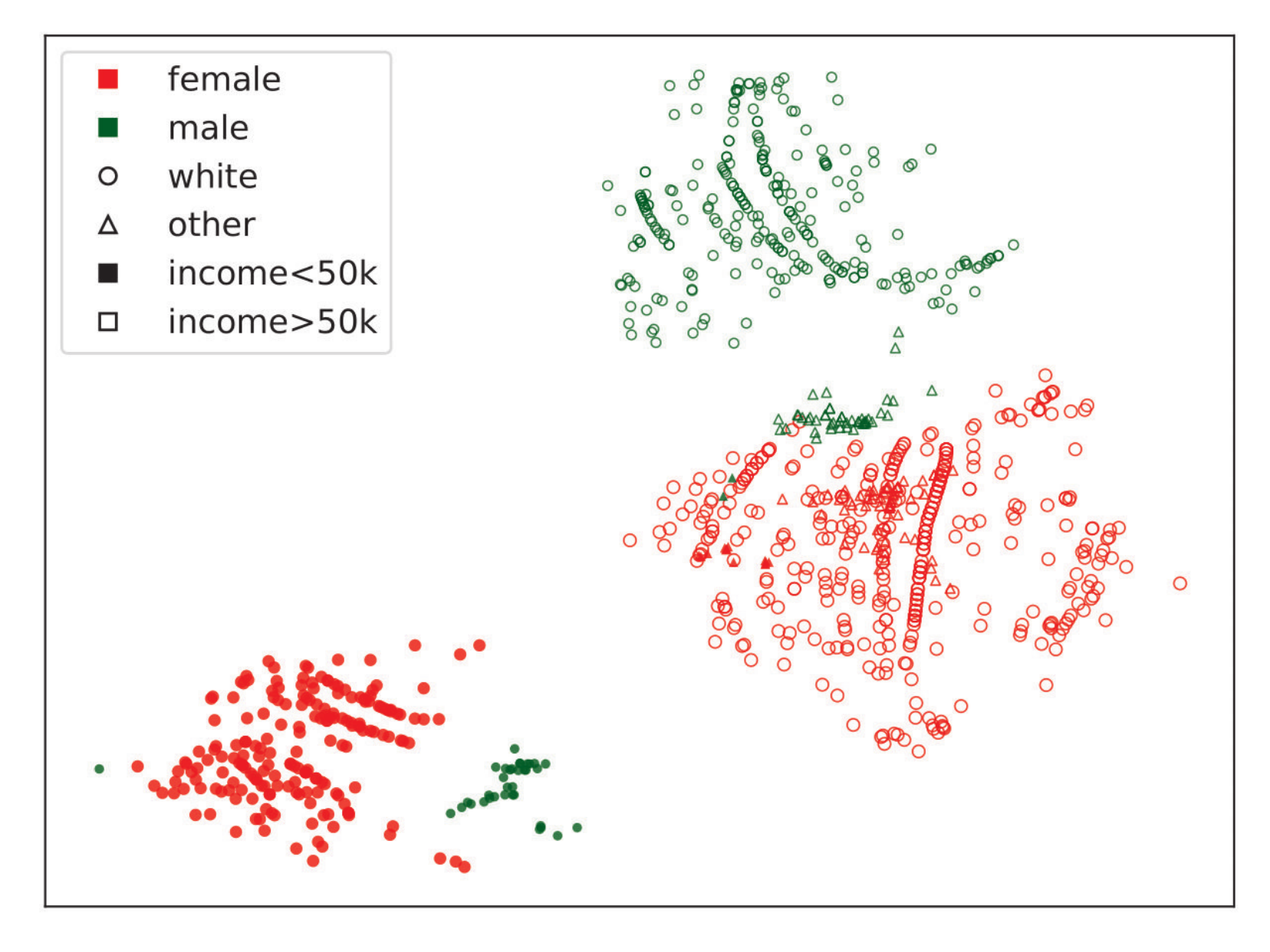}}
\subfigure[Fair-NeRV]{\label{fig:ex2-c}\includegraphics[height=1.2in,width=1.2in]{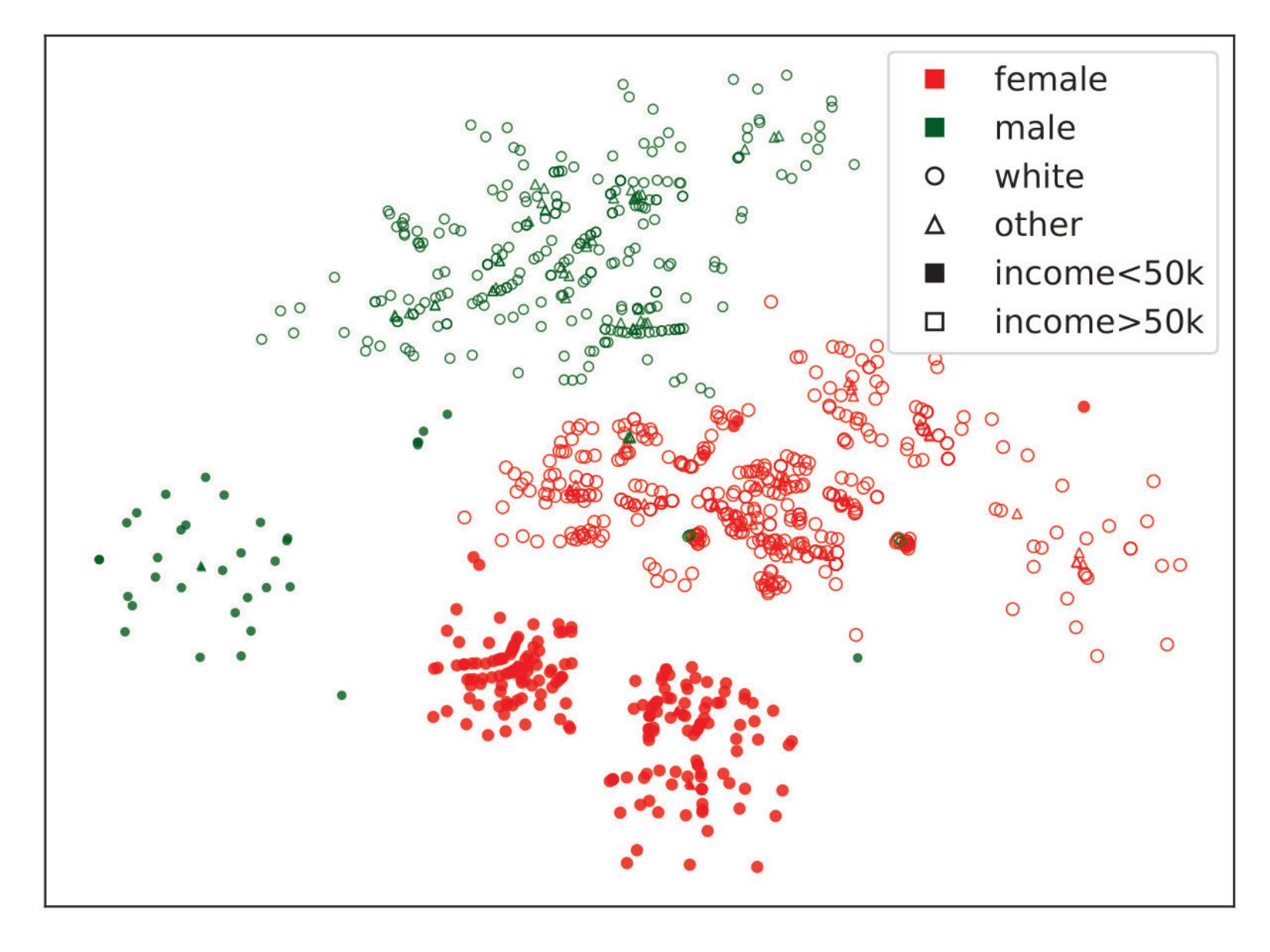}}
\subfigure[IMAPCE]{\label{fig:ex2-d}\includegraphics[height=1.2in,width=1.2in]{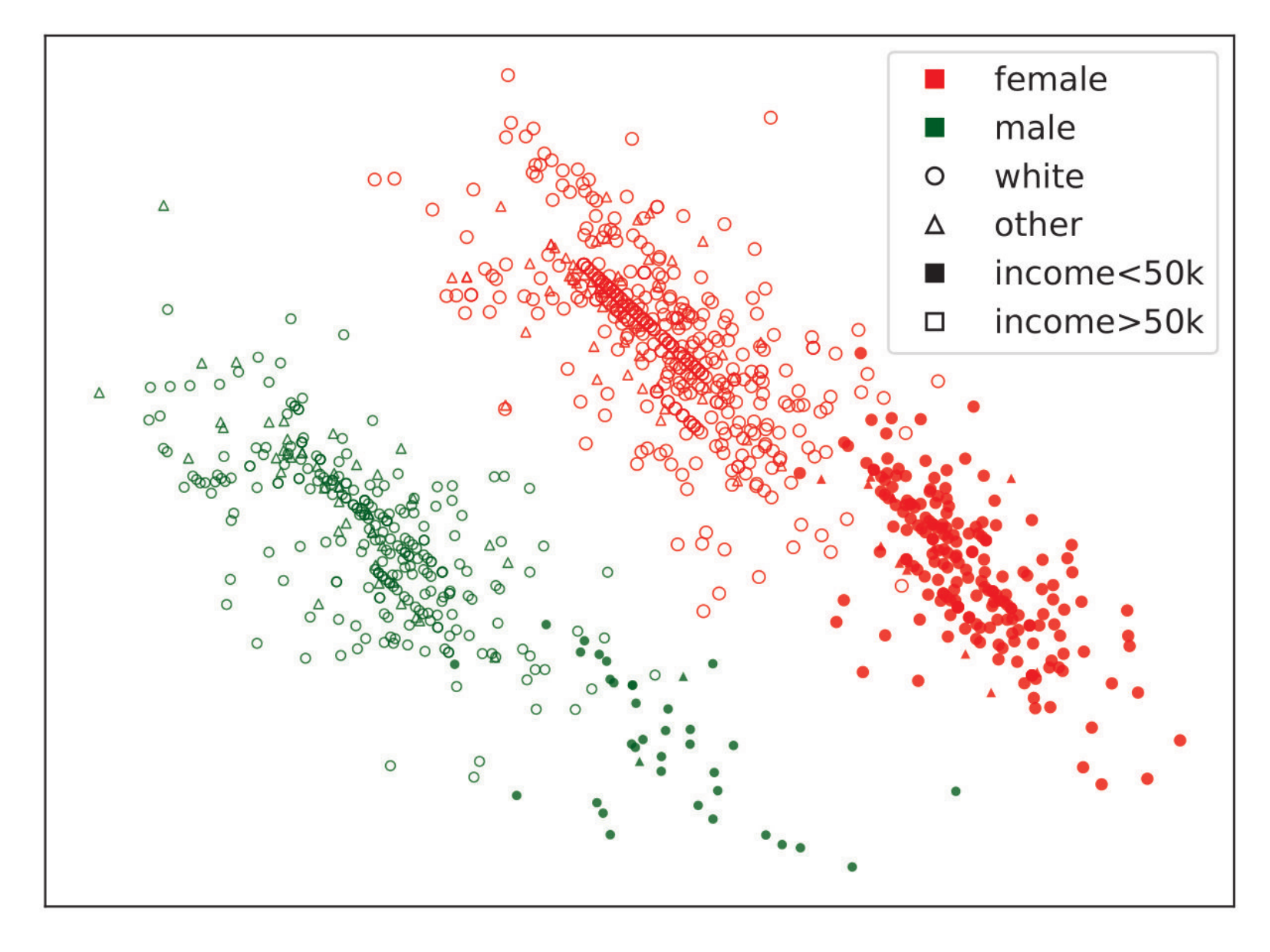}}
\caption{Top row shows synthetic data experiments with information of dimensions one to four as a prior. Bottom row illustrates UCI adult data experiments with ethnicity feature as a prior. (a) cPCA embeddings clustered w.r.t. labels from dimensions one to four. (b) ct-SNE embeddings are clustered w.r.t. labels from dimensions five to six (complementary structure) with some noticeable error (overlap). (c,d) Fair-NeRV and IMAPCE embeddings are clustered w.r.t. labels from dimensions five to six (complementary structure) with clearer separation. (e) cPCA fails to separate embeddings according to their gender and income. (f) ct-SNE clearly separates embeddings w.r.t. to their income and to some extent according to their gender. (g) Fair-NeRV computes embeddings that are mostly separated w.r.t. to gender and income with the exception of some outliers, while (h) IMAPCE perfectly clusters embeddings according to their gender and income (revealing complementary structure).}
\label{fig:UCI_Adult_projections}
\end{figure*}

Using the ethnicity feature as prior, we obtain cPCA, ct-SNE, Fair-NeRV and IMAPCE ($\alpha=1, \mu = 150$) embeddings as shown in Figures \ref{fig:ex2-a} - \ref{fig:ex2-d}. cPCA provides embeddings with mixed ethnicity, gender and income features, failing to exhibit any clear cluster formation. ct-SNE fails to remove the ethnicity information from male instances, but successfully does so for female ones. In addition, the complementary gender structure is not unveiled as there is overlap between red and green points. On the contrary, Fair-NeRV and IMAPCE remove the ethnicity variation by having clusters with mixed ethnicities. Their results are similar as they exhibit clusters of different gender (red-green colors) and income (filled-unfilled markers) attributes. However, IMAPCE slightly outperforms Fair-NeRV which incorrectly clusters together points with different gender and some points with different income labels.

Similar to the synthetic data experiments, we computed the NLS scores of all methods with respect to the ethnicity attribute and for neighborhood sizes of 10, 20, \dots, 100. We report the average NLS over the neighborhood size in Table \ref{table:1}. IMAPCE and Fair-NeRV achieve higher average NLS and thus discount the ethnicity attribute information more effectively than both cPCA and ct-SNE, as also qualitatively confirmed. Similar experiments using gender attribute and the combination of gender and ethnicity attributes as prior are included in Section \ref{Supp:Adult} of the Supplementary Material.

%MAKE HORISONTAL
\begin{table}[ht]
\caption{Mean test accuracy of a Linear classifier over ten different random train-test splits for both types of Complex data and mean NLS over 10, 20, \dots, 100 neighborhood sizes using the specified attribute(s) as prior for synthetic and adult.}
\begin{center}
\begin{tabular}{ccccc} 
 \toprule
Dataset & cPCA & ct-SNE & Fair-NeRV & IMAPCE\\ 
\midrule
Synthetic  & 0.16 & 0.34 & \textbf{0.49} & 0.45\\
\midrule
Adult &  0.11 & 0.13 & \textbf{0.23} & 0.21\\
\midrule 
MNIST+FMNIST &  0.75 & 0.65 & 0.78 & \textbf{0.81} \\
\midrule
CIFAR 100+FMNIST &  0.58 & 0.79 & 0.83 & \textbf{0.95} \\
\bottomrule
\end{tabular}
\label{table:1}
\end{center}
\end{table}

\begin{table}[ht]
\caption{Execution time of all methods with respect to each dataset}
\begin{center}
\begin{tabular}{c c c c c} 
 \toprule
 Dataset & cPCA & ct-SNE & Fair-NeRV & IMAPCE \\ 
 \midrule
Synthetic & \textbf{2 sec} & 30 sec & 5 min & 3 sec\\ 
 \midrule
Adult & \textbf{5 sec} & 70 sec & 5 min & \textbf{5 sec}\\
 \midrule
MNIST+FMNIST & \textbf{21 sec} & 63 sec & 1hr-cutoff & 26 sec\\
\midrule
CIFAR+FMNIST & \textbf{22 sec} & 61 sec & 1hr-cutoff & 25 sec\\
\bottomrule
\end{tabular}
\label{tab:time}
\end{center}
\end{table}

\begin{figure*}[ht]
\centering
\subfigure[MNIST + FMNIST]{\rotatebox[origin=l]{90}{\footnotesize \qquad Complex Data}\label{fig:ex3-a}\includegraphics[height=0.5in,width=1in]{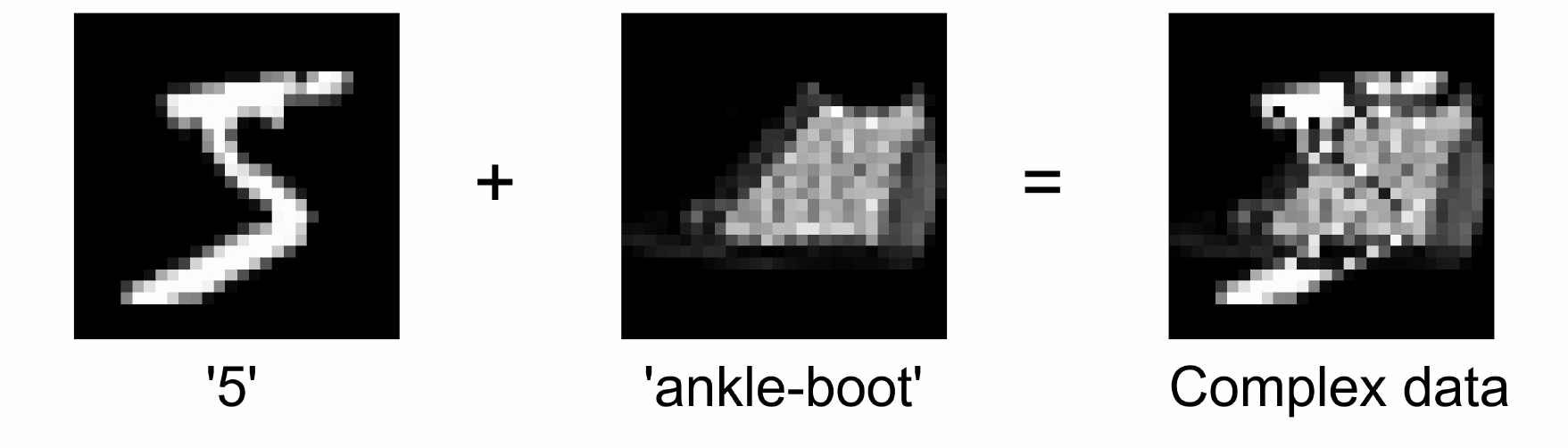}}
\subfigure[cPCA]{\label{fig:ex3-b}\includegraphics[height=1in,width=1in]{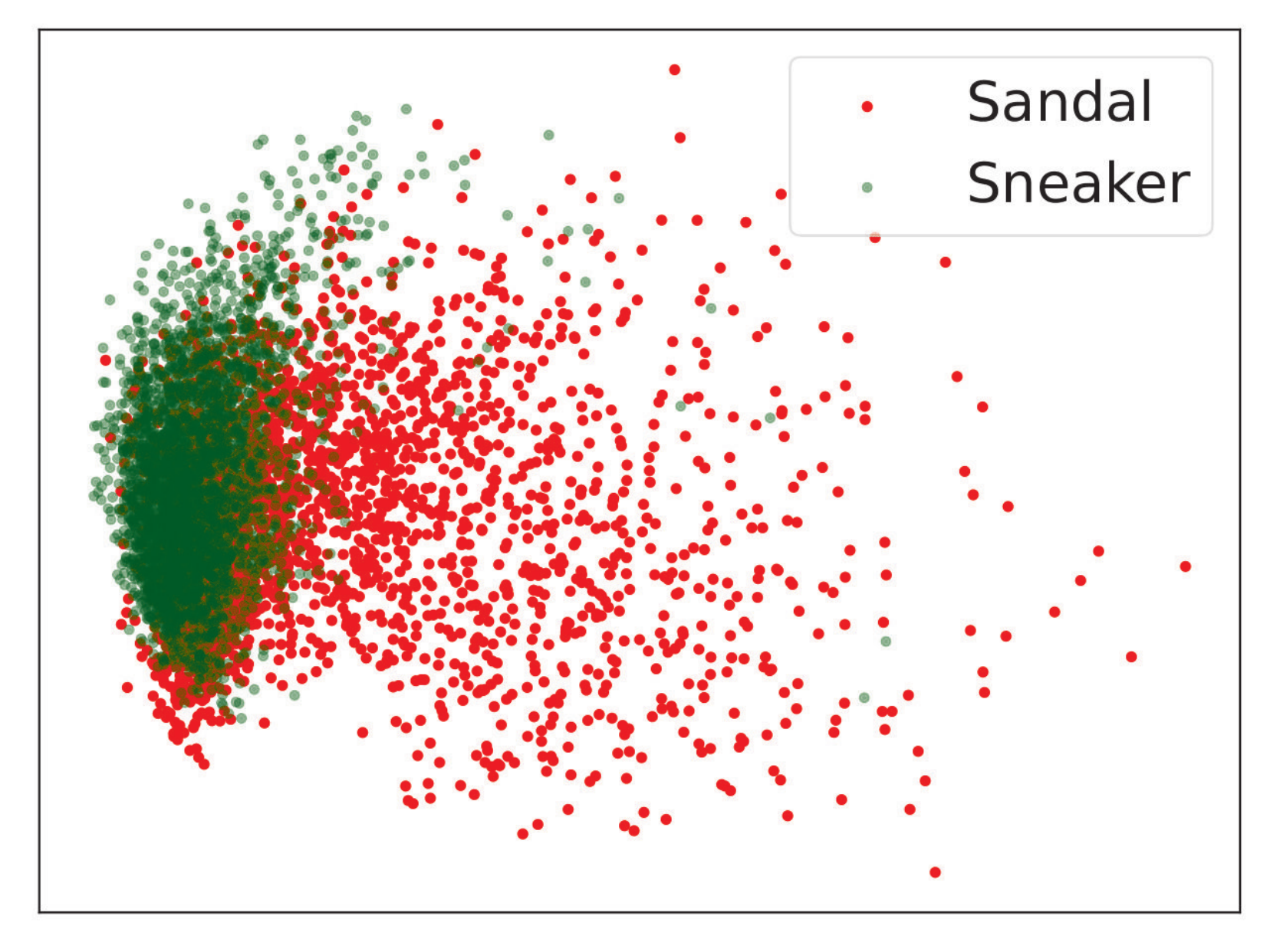}}
\subfigure[ct-SNE]{\label{fig:ex3-c}\includegraphics[height=1in,width=1in]{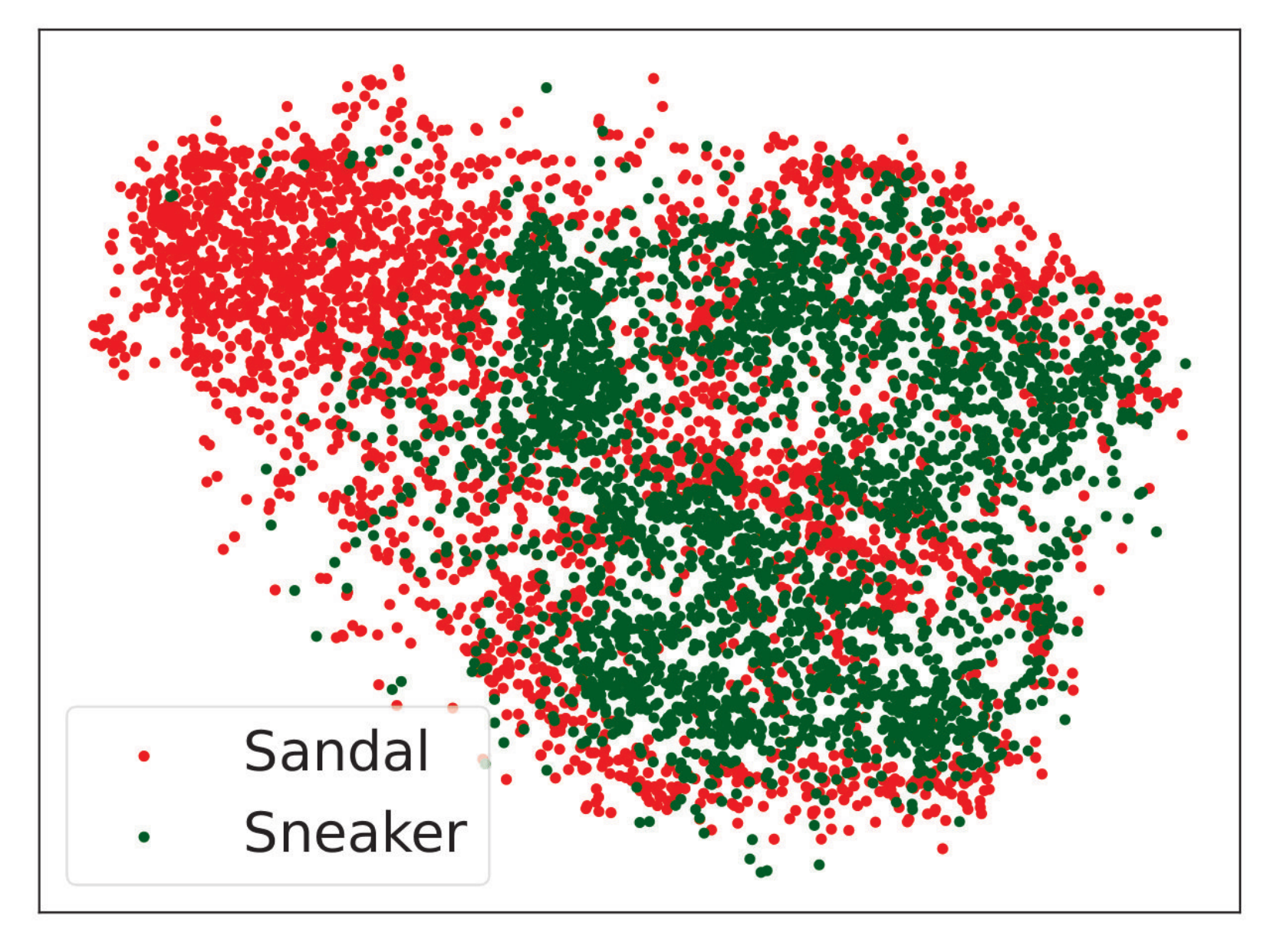}}
\subfigure[Fair-NeRV]{\label{fig:ex3-d}\includegraphics[height=1in,width=1in]{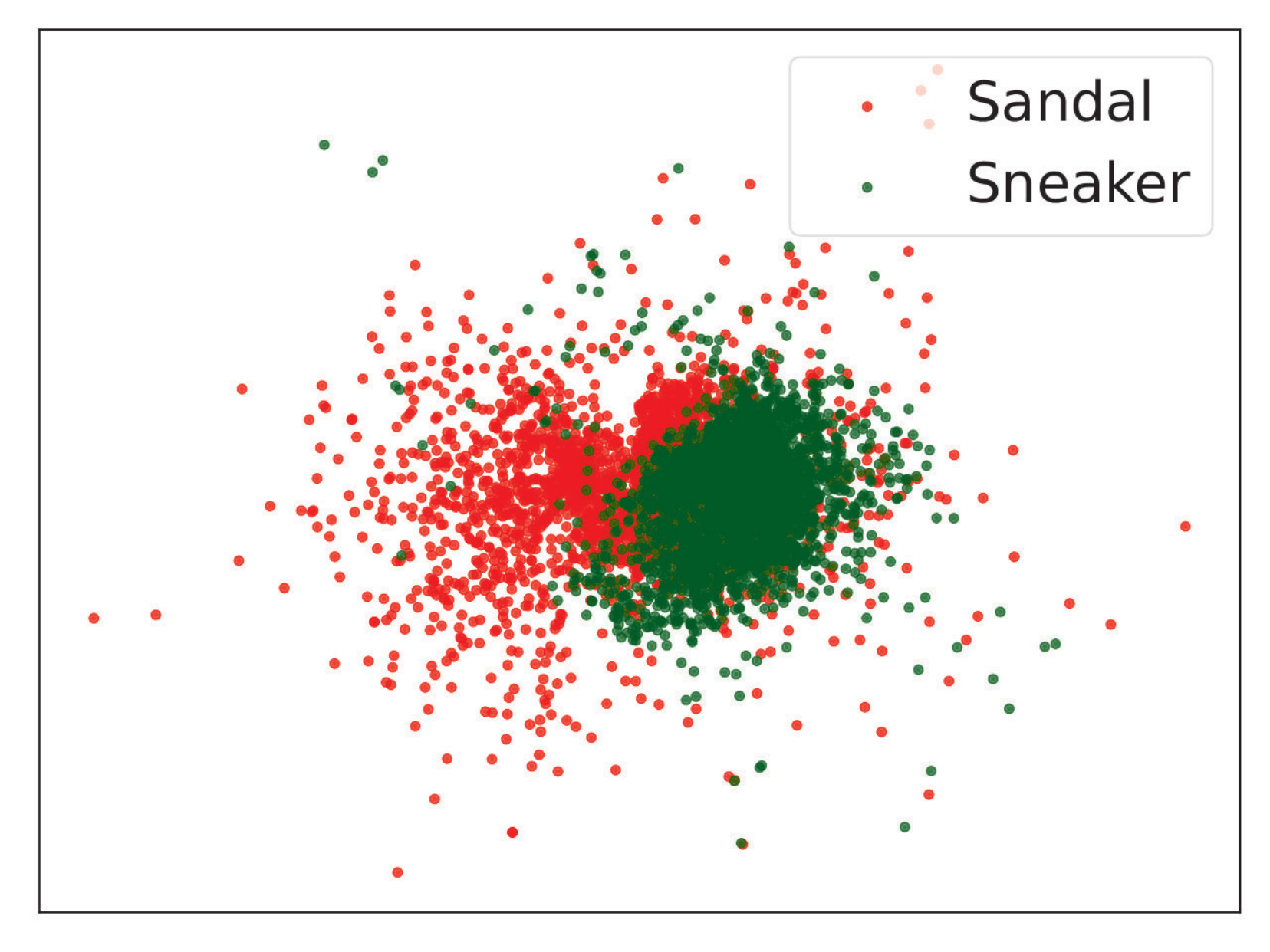}}
\subfigure[IMAPCE]{\label{fig:ex3-e}\includegraphics[height=1in,width=1in]{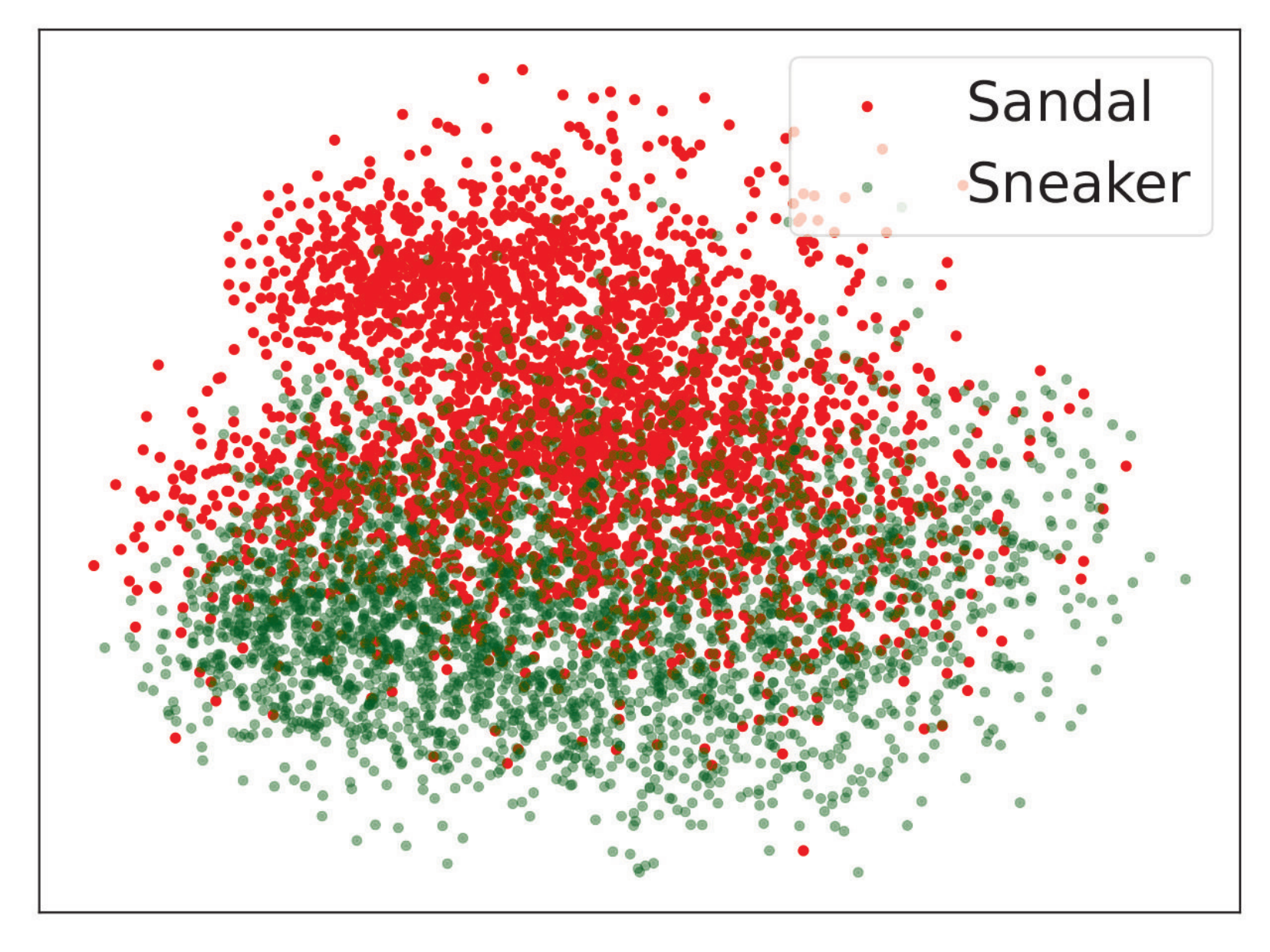}}\\
\subfigure[CIFAR-100 + FMNIST]{\rotatebox[origin=l]{90}{\footnotesize \qquad Complex Data}\label{fig:ex4-a}\includegraphics[height=0.5in,width=1in]{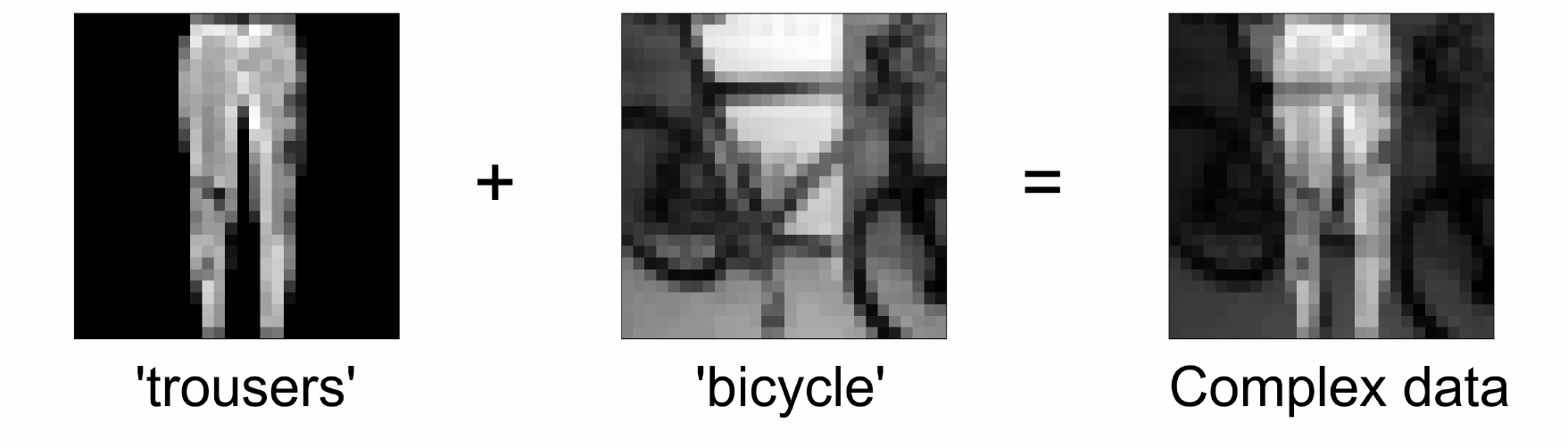}}
\subfigure[cPCA]{\label{fig:ex4-b}\includegraphics[height=1in,width=1in]{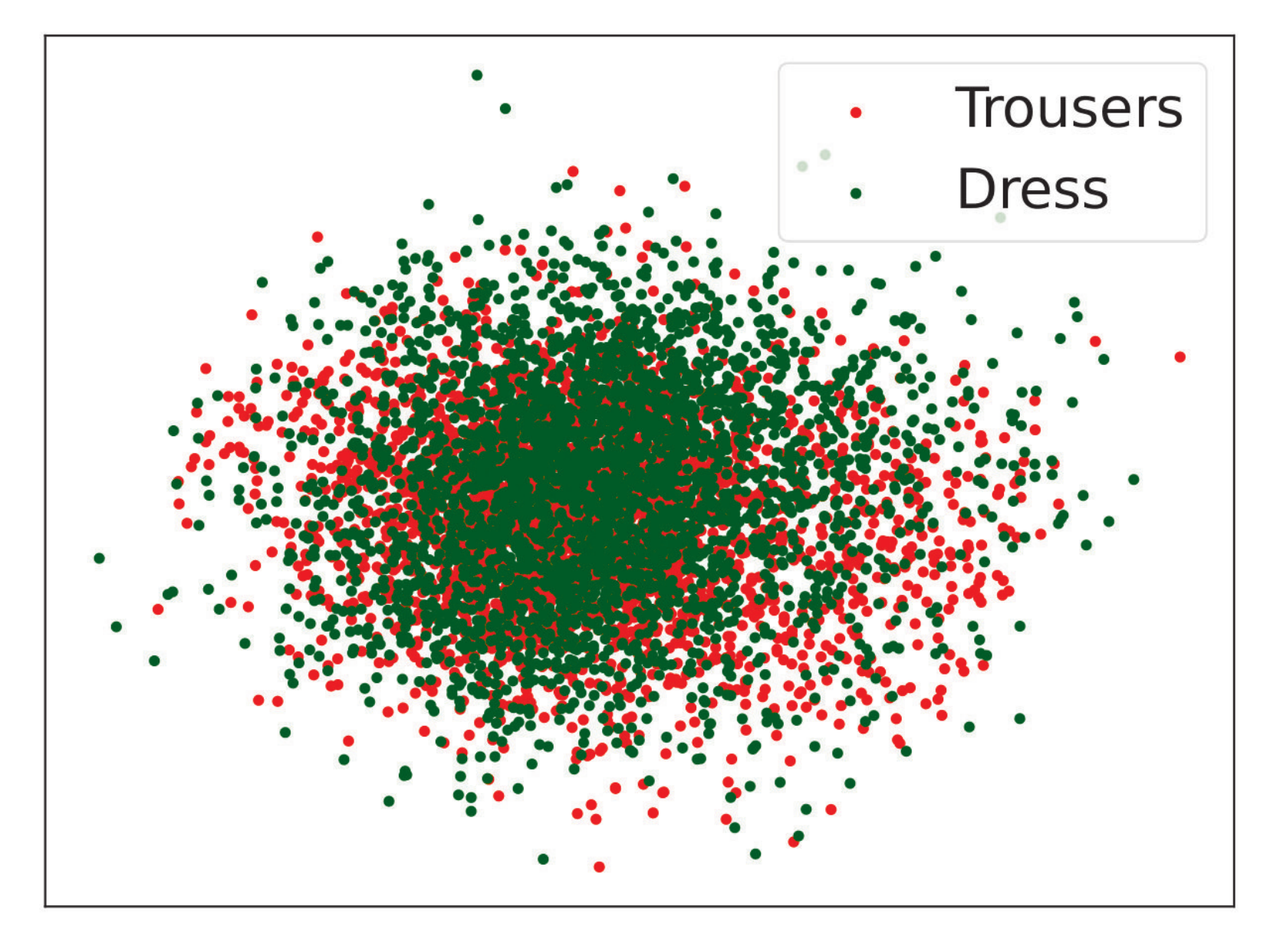}}
\subfigure[ct-SNE]{\label{fig:ex4-c}\includegraphics[height=1in,width=1in]{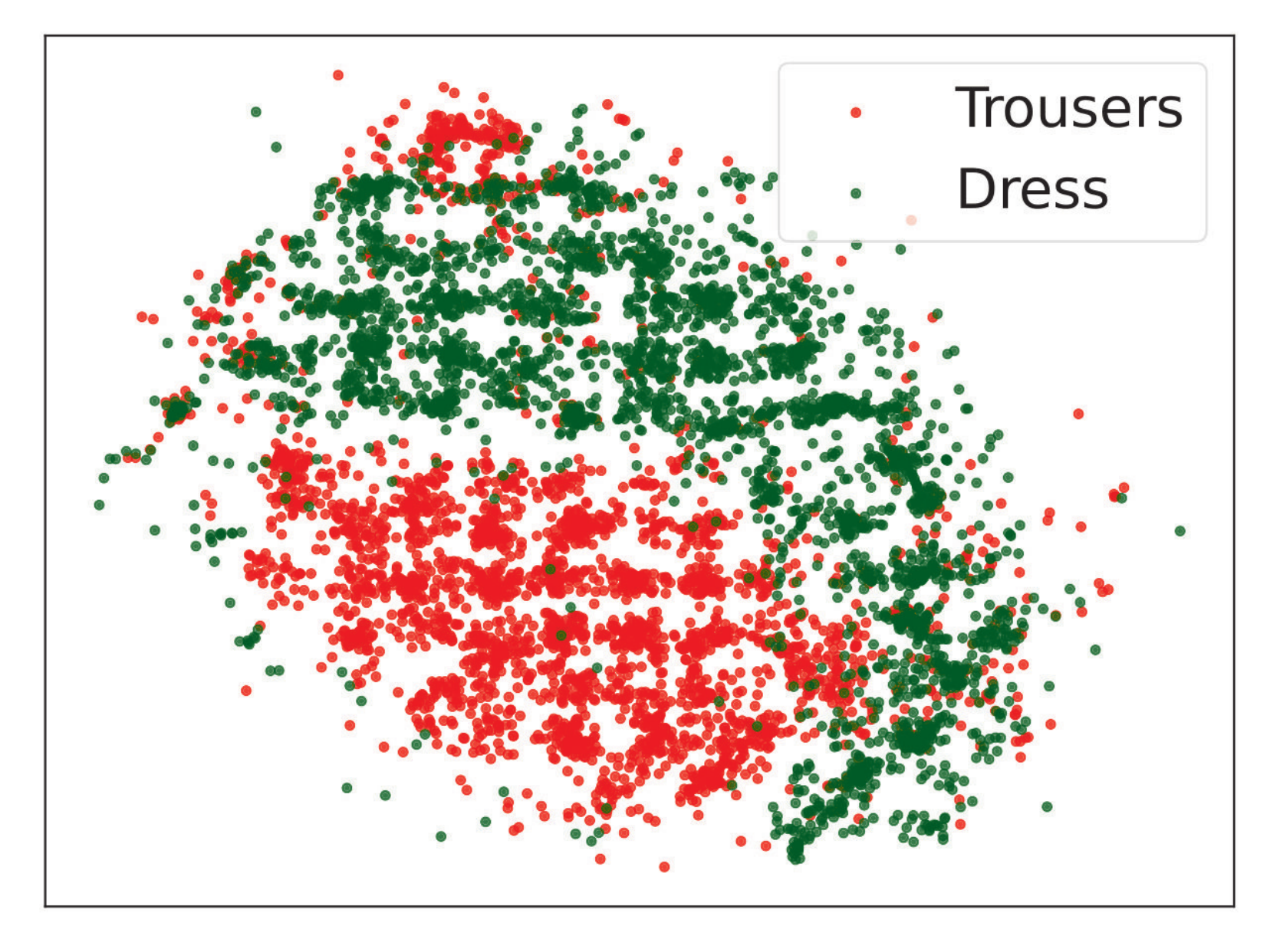}}
\subfigure[Fair-NeRV]{\label{fig:ex4-d}\includegraphics[height=1in,width=1in]{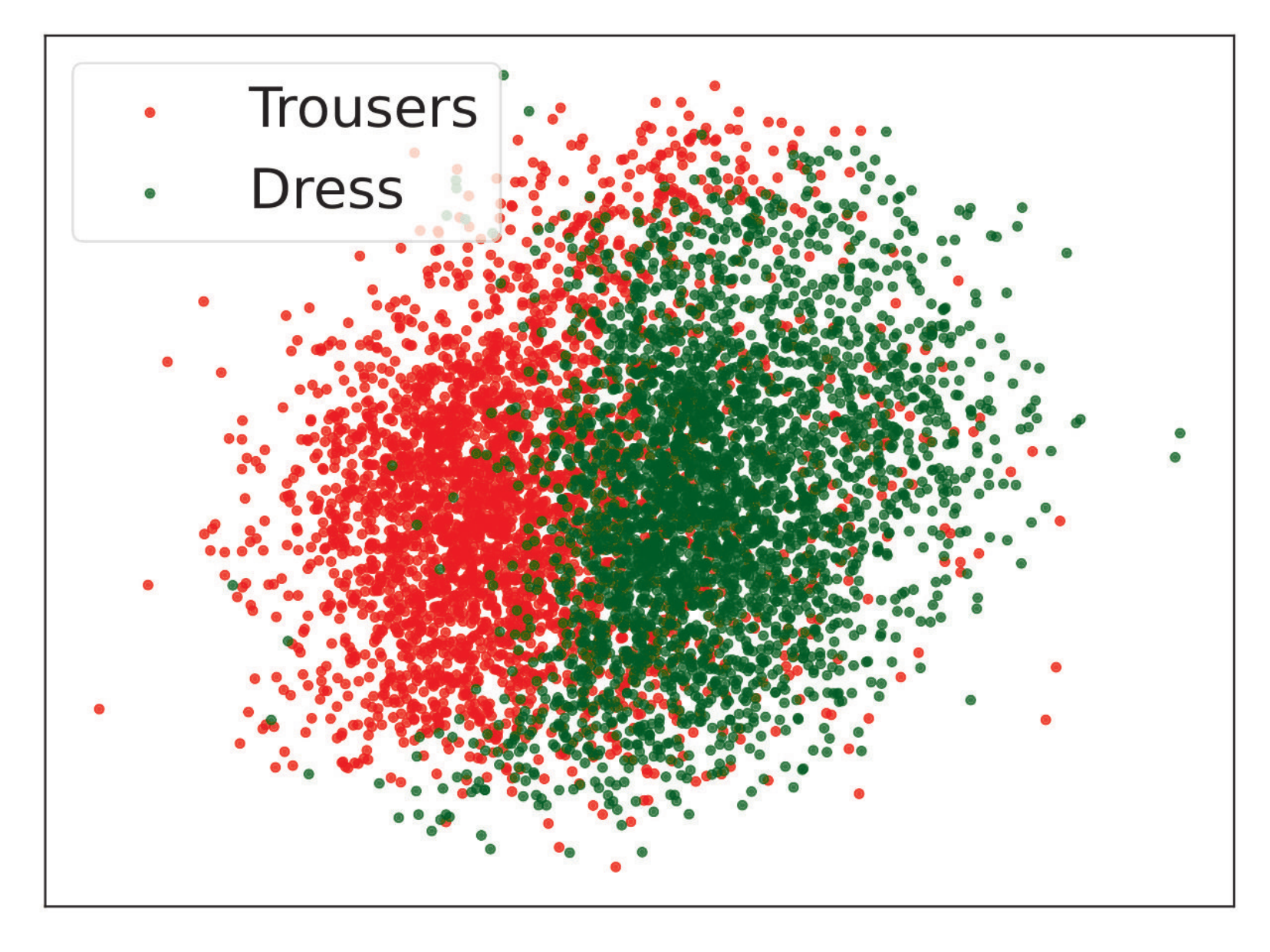}}
\subfigure[IMAPCE]{\label{fig:ex4-e}\includegraphics[height=1in,width=1in]{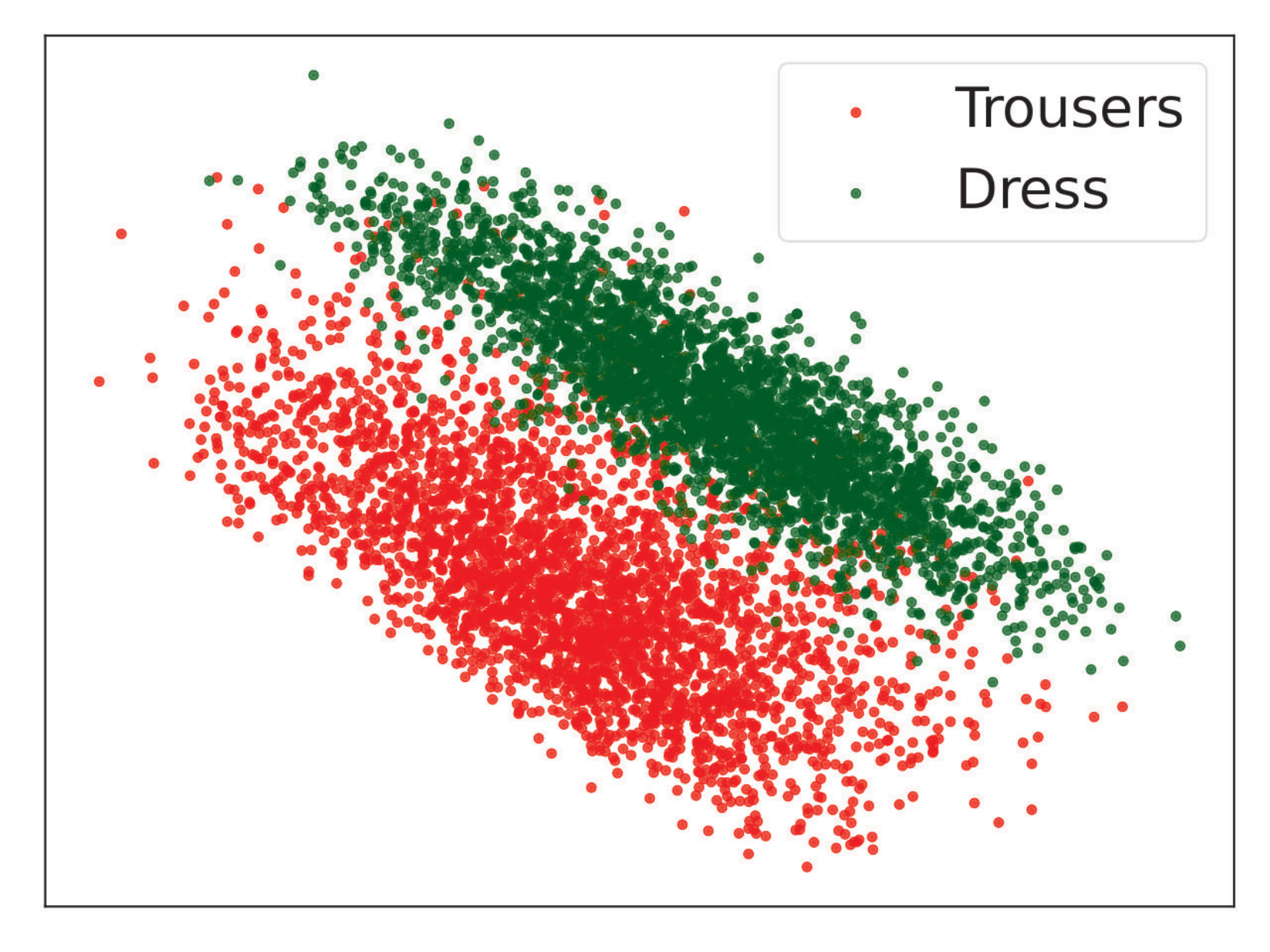}}
\caption{Top row shows complex MNIST + FMNIST experiments  using MNIST as prior (a), while bottom row complex CIFAR-100 + FMNIST experiments using CIFAR-100 as prior (f). In both cases, embeddings by cPCA (b,g), ct-SNE (c,h) and Fair-Nerve (d,i) appear significantly more mixed w.r.t their class, as opposed to the IMAPCE ones (e,j) which exhibit better segregation.}
\label{fig:Complex_data_projections}
\end{figure*}

\subsection{Sample-based prior}
\label{subsec:sample}
\subsubsection{MNIST + Fashion-MNIST (FMNIST)}
We created some complex data by combining instances from MNIST (which we consider the background part) and FMNIST (which we consider as the reference part) datasets. The task in this case is to compute two-dimensional embeddings that remove the information associated with the MNIST data and provide separation according to the complementary structure defined by the FMNIST labels. We constructed a series of complex data (6,000 samples) as follows. We first selected two different FMNIST classes and randomly sampled 6,000 images of those classes. We then superimposed those images with randomly chosen MNIST images. We repeated this procedure for a few combinations of FMNIST classes in order to create different series of complex data. An example of a complex data instance and its construction is shown in Figure \ref{fig:ex3-a}. The superimposed 28x28 images, as well as the MNIST images serving as the background, are flattened to 784 dimensional vectors before their processing. As background data we randomly select 1,000 MNIST samples which are different from the ones used for the complex data formation, but adequately capture the MNIST variation we wish to remove from the embeddings. While we can employ this prior data for both IMAPCE and cPCA, ct-SNE and Fair-NeRV are limited to label priors. To remove the MNIST information using ct-SNE and Fair-NeRV, we utilize the ground truth labels of the MNIST instances that were used for the complex data construction. Finally, for IMAPCE we set $\alpha = 1$ and $\mu = 10^5$.

Figures \ref{fig:ex3-b}-\ref{fig:ex3-e} show complex MNIST + FMNIST data embeddings (with "Sandal", "Sneaker" FMNIST ground truth labels) generated by all methods. We observe that ct-SNE, cPCA and Fair-NeRV separate a small portion of "Sandal" embeddings from the "Sneaker" ones. However, there is significant overlap between the majority of "Sandal" and "Sneaker" instances. On the other hand, IMAPCE computes embeddings that provide the clearest separation with respect to their FMNIST class as their overlap is smaller than all other methods.

To quantify the separability according to the FMNIST labels, we trained (75\%) and tested (25\%) a Linear classifier on the classification of the calculated 2D complex data embeddings with respect to these labels. The test-set accuracy is averaged over 10 random train-test splits and given in Table \ref{table:1}. IMAPCE not only achieves the best mean test-set accuracy and thus separation (also observed visually), but at the same time considers only a small amount of background MNIST information, as opposed to ct-SNE and Fair-NeRV which require all MNIST labels. We provide additional experiments for different MNIST + FMNIST complex data as well as SVM classifiers with accuracy, F1 and Recall scores in Section \ref{Supp:MFMNIST} of the Supplementary Material.

\subsubsection{CIFAR 100 + FMNIST}
Similar to the combination of MNIST with FMNIST data, we constructed more complex data by superimposing 6,000 images from CIFAR-100 \cite{krizhevsky2009learning} with images from FMNIST. CIFAR-100 has 100 classes, each of which has 600 color images. To perform the superimposition of 32x32 CIFAR-100 images with the 28x28 FMNIST images, we cropped the border of the CIFAR-100 images and subsequently converted them from color (RGB) to grayscale. Afterwards, we flattened each 28x28 superimposed image to a 784 dimensional vector. An example of a grayscale CIFAR-100 instance, FMNIST instance and their superimposition is shown in Figure \ref{fig:ex4-a}, while more examples are provided in Section \ref{Supp:C100-FMNIST} of the Supplementary Material. In this case, we consider the CIFAR-100 image data as background and the FMNIST data as the reference. Therefore, we wish to compute embeddings which exhibit cluster formation according to their FMNIST label. To achieve this, the CIFAR-100 label variation should be removed from the generated embeddings. For cPCA and IMAPCE, we define as background data 5,000 randomly selected CIFAR-100 images (not necessarily used during the dataset construction) which can capture the variance of the CIFAR-100 data. On the contrary, for ct-SNE and Fair-NeRV we provide the CIFAR-100 ground truth labels of the CIFAR-100 images we used when constructing the complex data. For IMAPCE, we set $\alpha = 1$ and $\mu = 10^7$.

Figures \ref{fig:ex4-b}-\ref{fig:ex4-e} show complex CIFAR-100 + FMNIST data embeddings (with "Trousers", "Dress" FMNIST ground truth labels) generated by all methods. We observe that cPCA completely fails to separate the embeddings according to their FMNIST class. While ct-SNE and Fair-NeRV achieve some sort of separation, there is considerable overlap of "Trousers" and "Dress" instances. On the contrary, the embeddings computed by IMAPCE achieve the best separation regarding their FMNIST class. 

To complement our qualitative observations, we trained a Linear classifier on the classification of the calculated 2D embeddings of all methods using identical specifications to the classifier of the MNIST + FMNIST case. The test-set accuracies are provided in Table \ref{table:1}, where the superiority of IMAPCE among all methods is demonstrated. This superiority is boosted by the fact that it is achieved under an unsupervised setting by considering only a small number (5,000 out of the total 60,000) of representative CIFAR-100 background instances that can effectively capture the CIFAR-100 structure that has to be factored out from the computed embeddings. On the contrary, both ct-SNE and Fair-NeRV need the specific CIFAR-100 labels of all the instances that were used when constructing the complex data. 

Finally, in terms of running time performance, IMAPCE has a comparable execution time to cPCA, while it is faster than both ct-SNE and Fair-NeRV across all experimental setups as Table~\ref{tab:time} shows. Due to the large feature space of the sample-based prior experiments, an 1hr cutoff time was set for Fair-NeRV (and the final embeddings computed within this cutoff time were used for the experiments) to facilitate hyperparameter search. More experiments for different combinations of FMNIST classes are included in Section \ref{Supp:C100-FMNIST} of the Supplementary Material. Additionally, SVM classifiers are trained on the 2D embeddings and accuracy, F1 and Recall scores are computed for all methods and are also included in that Section of the Supplementary Material.

\begin{figure*}[h]
\centering
% \vspace*{0.75cm}
\subfigure[]{\label{fig:ex5-a}\includegraphics[height=1.3in,width=1.3in]{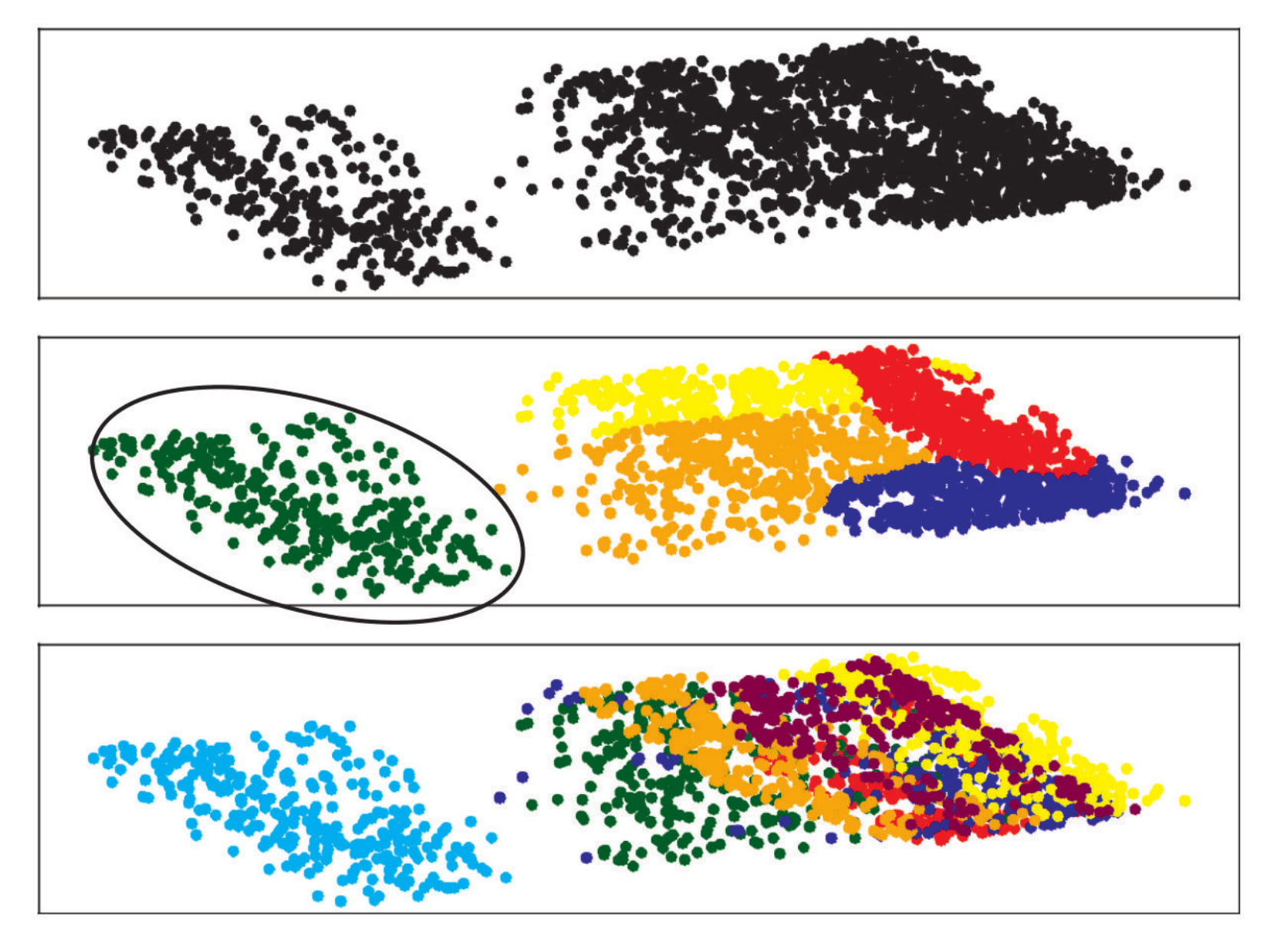}}
\subfigure[]{\label{fig:ex5-b}\includegraphics[height=1.3in,width=1.3in]{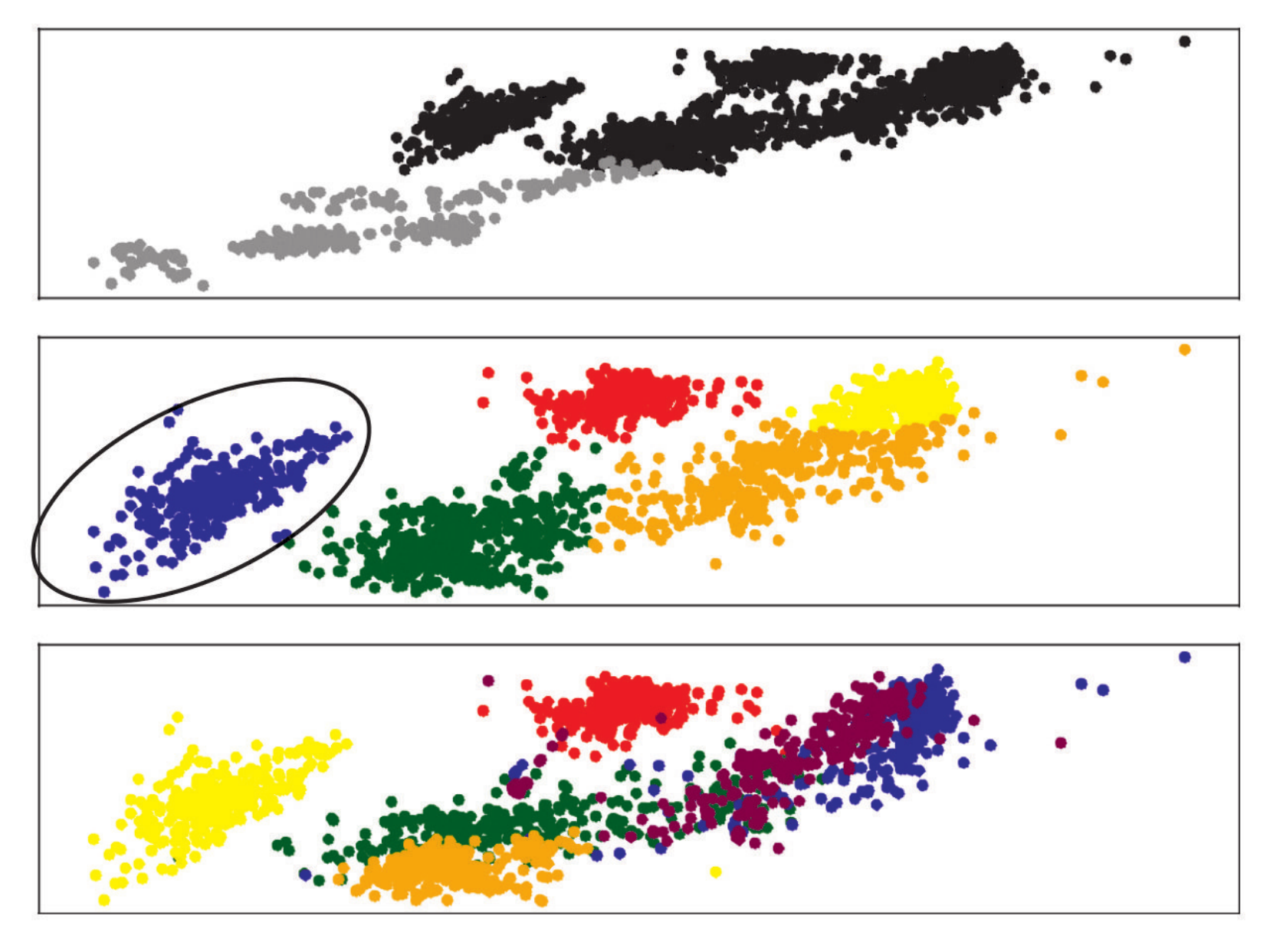}}
\subfigure[]{\label{fig:ex5-c}\includegraphics[height=1.3in,width=1.3in]{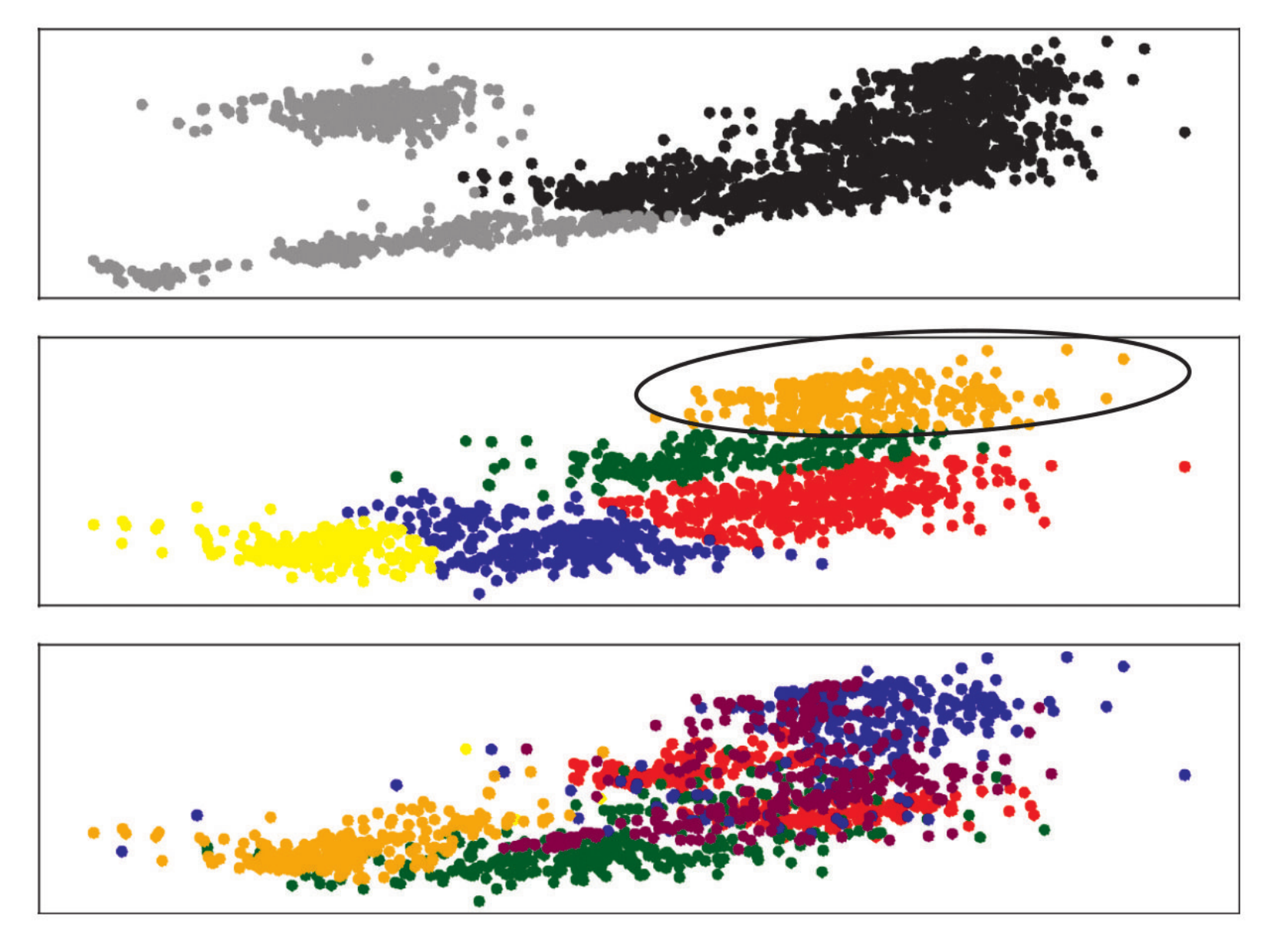}}
\subfigure[]{\label{fig:ex5-d}\includegraphics[height=1.3in,width=1.3in]{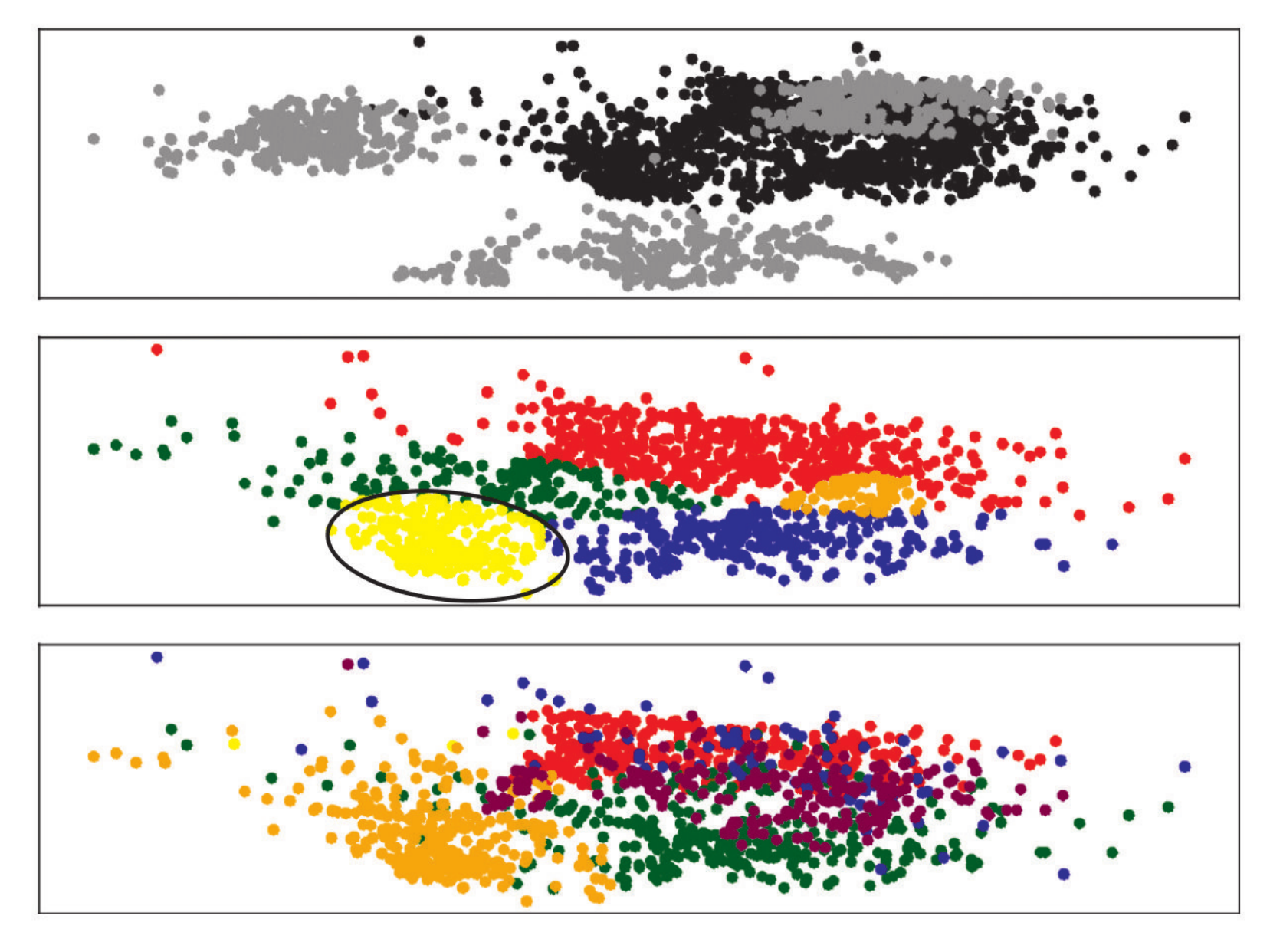}}
% \subfigure[Iteration 5]{\label{fig:ex4-e}\includegraphics[height=1.5in,width=1.5in]{AAAI_submission/Image_segmentation_data/Proj4_yellow.pdf}}
% \subfigure[Iteration 6]{\label{fig:ex4-f}\includegraphics[height=1.5in,width=1.5in]{AAAI_submission/Image_segmentation_data/Proj5_green.pdf}}

\caption{Iterative exploration of UCI image segmentation data by IMAPCE. Each column (subfigure) corresponds to an iteration of the exploration process. In every column (subfigure), the upper plot shows the data projections where prior data are colored in grey while the unexplored subset in black. The middle plot shows the clustering of the unexplored points according to DPGMM and the most distinct cluster encircled. The bottom plot illustrates the unexplored points colored according to their ground truth label.}
\label{fig:Iterative}
\end{figure*}

\subsection{Subset-based prior}
\label{subsec:subset}
In this setup, prior data are subsets of the original data that share the same class (and are thus similar). The goal is to generate embeddings that reveal the cluster structure of the remaining samples (unexplored subset) of these datasets. By employing our structure extraction framework with IMAPCE or cPCA, we sequentially obtain new sets of informative embeddings that gradually unveil the underlying cluster structure of all high-dimensional data. In this case, we did not compare against ct-SNE and Fair-NeRV as they both require data label availability for the whole dataset, which contradicts the present subset-based analysis setup.

We performed the iterative visual exploration of Image Segmentation data from the UCI machine learning repository \cite{UCI} under various prior data assumptions. This dataset consists of 2,310 samples and 19 attributes and includes 330 instances from seven different classes, namely "sky", "grass", "path", "foliage", "cement", "brickface", "window". Assuming no prior data and selecting $\alpha = 1$, $\mu = 10^5$, $s= 75$, we sequentially apply IMAPCE on UCI Image segmentation data. The obtained projections are shown in Figure \ref{fig:Iterative}, where each subfigure consists of three subplots. The top row illustrates the IMAPCE data projections for a specific iteration of the process. Grey points correspond to the prior data, while black points correspond to the unexplored subset of samples. The middle row shows the results of a DPGMM clustering on the unexplored points where the most distinct cluster is encircled. The bottom row shows the unexplored points coloured according to their ground truth class.

The first data projection is shown in Figure \ref{fig:ex5-a}. The upper plot consists of solely black points as there is no prior data. All data are clustered in the middle subplot and the green cluster is the most distinct. Its points are considered very similar and are stored for the evaluation stage. Subsequently, these points define the prior data and are removed from the unexplored data. Afterwards, the second iteration takes place and new informative embeddings are calculated and shown in Figure \ref{fig:ex5-b}. Separation of data samples that were previously overlapping is now observed and indicative of an informative data projection. The grey points of the upper subplot correspond to the prior data while the black ones refer to the unexplored points, which are clustered in the middle subplot. The most distinct cluster is the blue one and its points are incorporated in the prior data and removed from the unexplored data. This iterative process is repeated until no clusters of the unexplored points are formed (only outliers are left). Gradual exploration of the whole dataset contributes to the extraction of new and meaningful underlying structure. For brevity and demonstration reasons, only the first four iterations are illustrated in Figures \ref{fig:ex5-a}-\ref{fig:ex5-d} and the rest of the exploration analysis is omitted.

To quantitatively compare IMAPCE and cPCA, we ran several experiments on UCI Image Segmentation ($\alpha = 1$, $\mu = 10^5$, $s= 75$) with different initial subsets of prior data. Performance evaluation took place after the exploration of the data had finished. During the evaluation stage, the quality of the most distinct clusters (which are stored along the exploration process) is measured with respect to their ground truth labels using the Jaccard \cite{jaccard} and NMI scores. Both scores are widely used in the DR literature for evaluating cluster quality. Detailed results for IMAPCE and cPCA are given in Table \ref{table:UCI_ImSeg}. We consider the case of no prior data and also experiment by setting as prior data subsets of data samples from every ground truth class. Overall, IMAPCE clearly outperforms cPCA on both scores under all prior data assumptions. The superiority of IMAPCE on the Jaccard and NMI scores indicates that it promotes enhanced cluster segregation in comparison to cPCA, owing to the inclusion of the kurtosis term. We provide equivalent quantitative results for 10,000 randomly selected MNIST instances in Section \ref{Supp:Iter_Expl} of the Supplementary Material.
\begin{table}[ht]
  \caption{Jaccard (averaged over all classes) and NMI scores for the evaluation of IMAPCE and cPCA on the visual exploration of the UCI Image Segmentation data under different prior data assumptions. Ten random initialisations were used for IMAPCE.}
\centering
  \label{tab:uci}
  \begin{tabular}{llrl}
    \toprule
    Method & Prior class & Mean Jaccard &  NMI\\
    \midrule
     cPCA & - & 0.44 & 0.46\\ 
    IMAPCE & - &$\mathbf{0.63} \pm 0.05$ & $\mathbf{0.67} \pm 0.06$ \\ \midrule
    
cPCA  & "Brickface" & 0.49 & 0.60\\ 
 IMAPCE & "Brickface" &$ \mathbf{0.63} \pm 0.04$ & $\mathbf{0.67} \pm 0.06$ \\ \midrule
 
  cPCA & "Cement" & 0.47 & 0.60\\ 
 IMAPCE & "Cement" &$ \mathbf{0.64} \pm 0.07$ & $\mathbf{0.69} \pm 0.06$\\ \midrule
 
  cPCA  & "Foliage" & 0.47 & 0.61\\ 
 IMAPCE & "Foliage" &$\mathbf{0.65} \pm 0.05$ & $\mathbf{0.72} \pm 0.05$\\ \midrule
 
cPCA  & "Grass" & 0.50 & 0.48\\ 
 IMAPCE & "Grass" &$\mathbf{0.59} \pm 0.08$ & $\mathbf{0.64} \pm 0.08$\\ \midrule
 
 cPCA  & "Window" & 0.50 & 0.57\\ 
 IMAPCE & "Window" &$\mathbf{0.68} \pm 0.06$ & $\mathbf{0.74} \pm 0.04$\\ \midrule
  
 cPCA & "Sky" & 0.35 & 0.43\\ 
 IMAPCE & "Sky" &$\mathbf{0.57} \pm 0.05$ & $\mathbf{0.61} \pm 0.07$\\ \midrule
 
 cPCA  & "Path" & 0.43 & 0.56\\ 
 IMAPCE & "Path" &$\mathbf{0.59} \pm 0.06$ & $\mathbf{0.65} \pm 0.07$\\
 
    \bottomrule
  \end{tabular}
  \label{table:UCI_ImSeg}
\end{table}

\section{Conclusion}
\label{Conclusion}
In this work we proposed IMAPCE, a dimensionality reduction method to generate low-dimensional embeddings that reveal previously unknown underlying structures while at the same time filtering out any unwanted prior information from the original data. A novel bi-objective formulation is introduced that jointly optimizes a projection pursuit index with a contrastive-PCA loss over the Stiefel manifold. We validated the performance of our proposed method on a set of diverse datasets for three different case scenarios of prior knowledge removal. IMAPCE demonstrated competitive performance against other compared methods in both cluster separation and removal of unwanted knowledge without requiring explicit data labels, while also being computationally efficient in practice.

%%%%%%%%%%%%%%%%%%%%%%%%%%%%%%%%%%%%%%%%%%%%%%%%%%%%%%%%%%%%%%%%%%%%%%%%

%%% Use this environment to include acknowledgements (optional).
%%% This will be omitted in doubleblind mode.

\begin{ack}
Xenophon Evangelopoulos acknowledges financial support from the Leverhulme Trust via the Leverhulme Research Centre for Functional Materials Design. This work was also supported by a studentship from the School of Electrical Engineering, Electronics and Computer Science, at the University of Liverpool, UK.
\end{ack}

%%%%%%%%%%%%%%%%%%%%%%%%%%%%%%%%%%%%%%%%%%%%%%%%%%%%%%%%%%%%%%%%%%%%%%%%

%%% Use this command to include your bibliography file.
\bibliography{m948}

%%%%%%%%%%%%%%%%%%%%%%%%%%%%%%%%%%%%%%%%%%%%%%%%%%%%%%%%%%%%%%%%%%%%%%
\clearpage
\appendix
\section{Dirichlet Process Gaussian Mixture Model} \label{Supp:DPGMM}
\label{Dirichlet Process Gaussian Mixture Model}
A Gaussian mixture model with $K$ components can be described by
\begin{equation}
    \label{eq:gmm}
    p(\mathbf{z} | \{\theta_l\}_{l=1}^K) = \sum_{l=1}^K \mathbf{w}_l\mathcal{N}(\mathbf{z}|\mathbf{m}_l,\mathbf{C}_l^{-1}),
\end{equation}
where $\theta_l = \{ \mathbf{w}_l,\mathbf{m}_l,\mathbf{C}_l^{-1} \}$ is the set of parameters for component $l$, $\mathbf{w}_l$ are the mixing weights of the Gaussians (or clusters here) satisfying $\sum_{i=l}^K\mathbf{w}_l = 1$, $\mathbf{m}_l$ is the mean vector for cluster $l$, and $\mathbf{C}_l^{-1}$ is its inverse covariance matrix. The latter ones are modeled by a joint Normal/Wishart distribution
\begin{equation}
    \label{eq:gmm_meancov}
    (\mathbf{m}_l,\mathbf{C}_l^{-1}) \sim \mathcal{N}\mathcal{W}(\mathbf{\xi},\rho,\beta,W)
\end{equation}
Here, $\xi$ corresponds to a prior of the mean $\textbf{m}_l$ of l-th cluster, and $\rho$ is a scalar representing how strongly we believe this prior. $\textbf{W}$ is a prior of the precision matrix (inverse Covariance matrix) $\textbf{C}_l^{-1}$ of l-th cluster, and $\beta$ represents how strongly we believe this prior. The DPGMM model follows a hierarchical structure where the data is modeled first as in Eq.~\eqref{eq:gmm}, the component parameters in the second stage and finally the various hyperparameters of those~\cite{rasmussen}.

\section{Hyperparameters' choice for ct-SNE and Fair-NeRV} \label{Supp:hypers}
Both ct-SNE and Fair-NeRV come with a few hyperparameters than need to be selected before running the experiments. ct-SNE's hyperparameters correspond to perplexity and $\acute\beta$. These were selected performing a grid search with perplexity 50, 100, 200 and $\acute\beta$ $\in$ $\{10^{-2}, 10^{-3}, \dots , 10^{-7}\}$. The combination of perplexity and $\beta$ values that generated the embeddings which achieved the best cumulative performance (Linear test-set accuracy for both cases of Complex data and mean NLS for adult and synthetic data) were selected. We note that for the smaller adult and synthetic data we used a summarization strength $\theta = 0$, as suggested by the authors, while for both cases of the larger and higher-dimensional Complex data we used $\theta = 0.5$ to speed up the optimization convergence.

Fair-NeRV has five hyperparamaters, each of which can take values within a prespecified range. More specifically,  $\tau^{\in} \in [0, 1]$, $\tau^{\notin} \in (\tau^{\in}, 1]$, $\beta \in [0, 1]$, $\gamma \in [0, 1]$ and $\omega \in [0.5, 0.99]$. We performed uniform sampling of 20 sets of hyperparameters and, similarly to ct-SNE, we kept the set which achieved the best cumulative performance across all tasks and datasets.

\section{UCI Adult Data Embeddings} \label{Supp:Adult}
\begin{figure*}[ht]
\centering
% \vspace*{0.75cm}
\subfigure[cPCA]{\label{fig:ex6-a}\includegraphics[height=1.2in,width=1.5in]{UCI_Adult_data/uci_Adult_cPCA_ethnicity.pdf}}
\subfigure[ct-SNE]{\label{fig:ex6-b}\includegraphics[height=1.2in,width=1.5in]{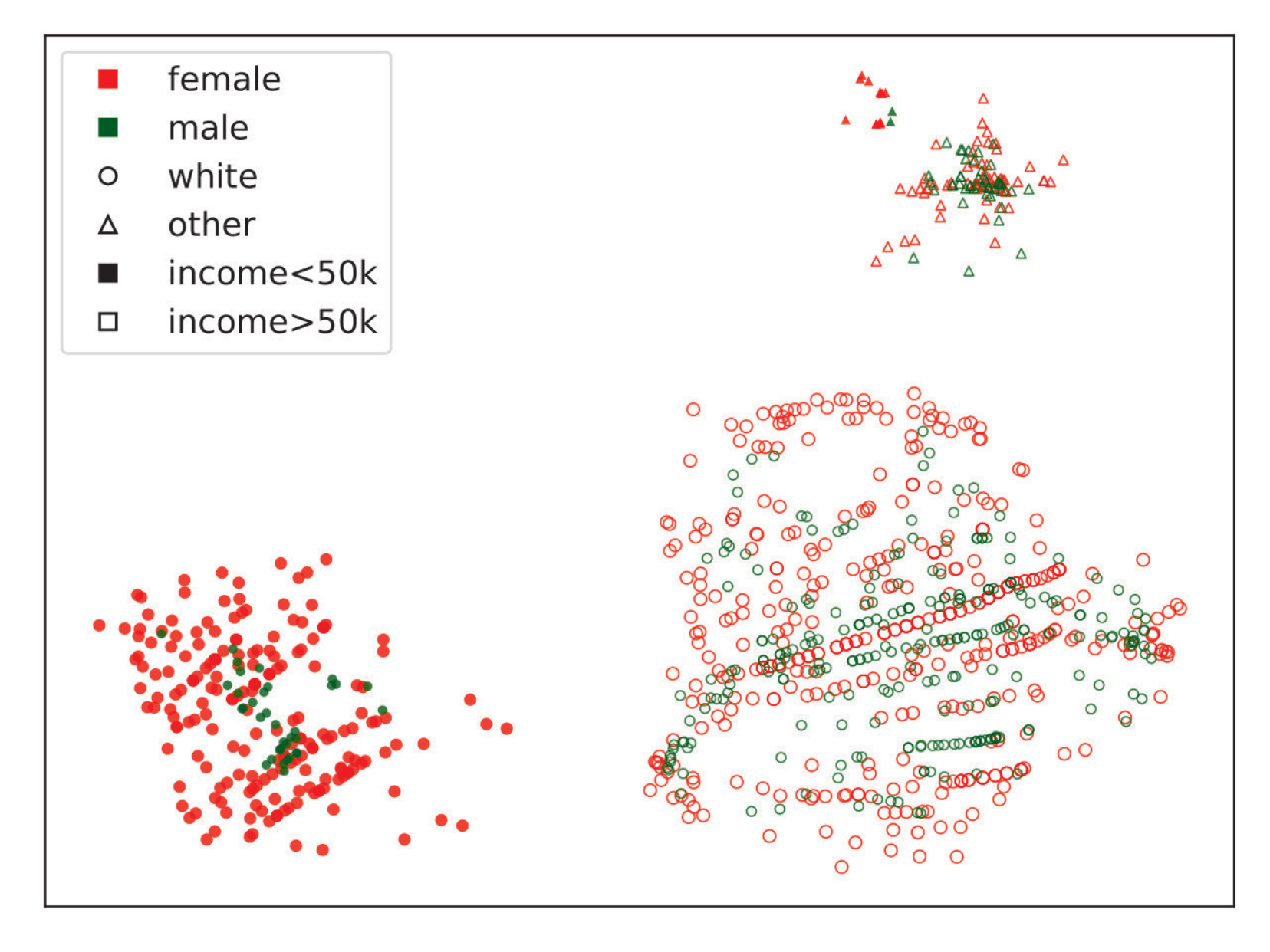}}
\subfigure[Fair-NeRV]{\label{fig:ex6-c}\includegraphics[height=1.2in,width=1.5in]{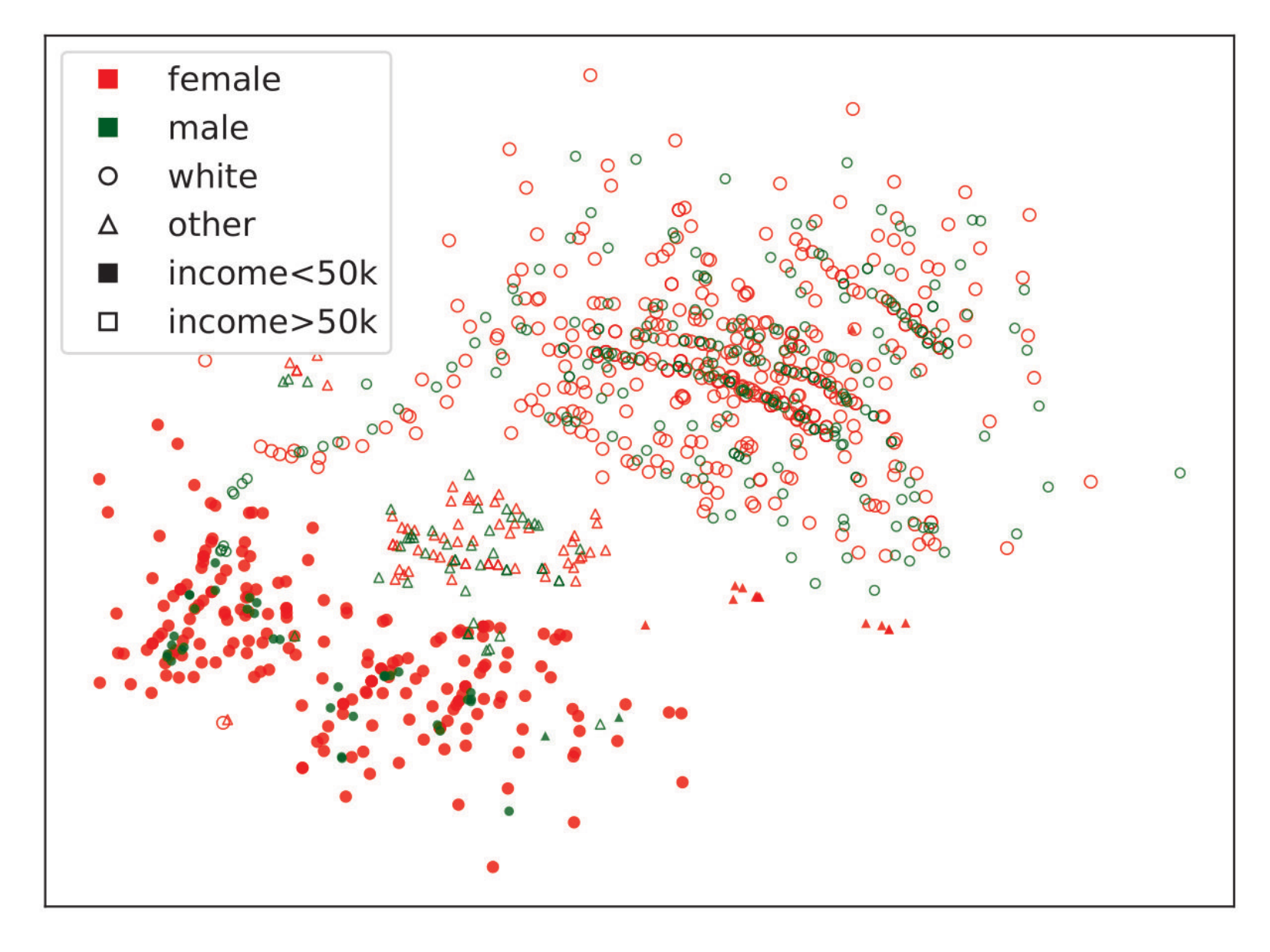}}
\subfigure[IMAPCE]{\label{fig:ex6-d}\includegraphics[height=1.2in,width=1.5in]{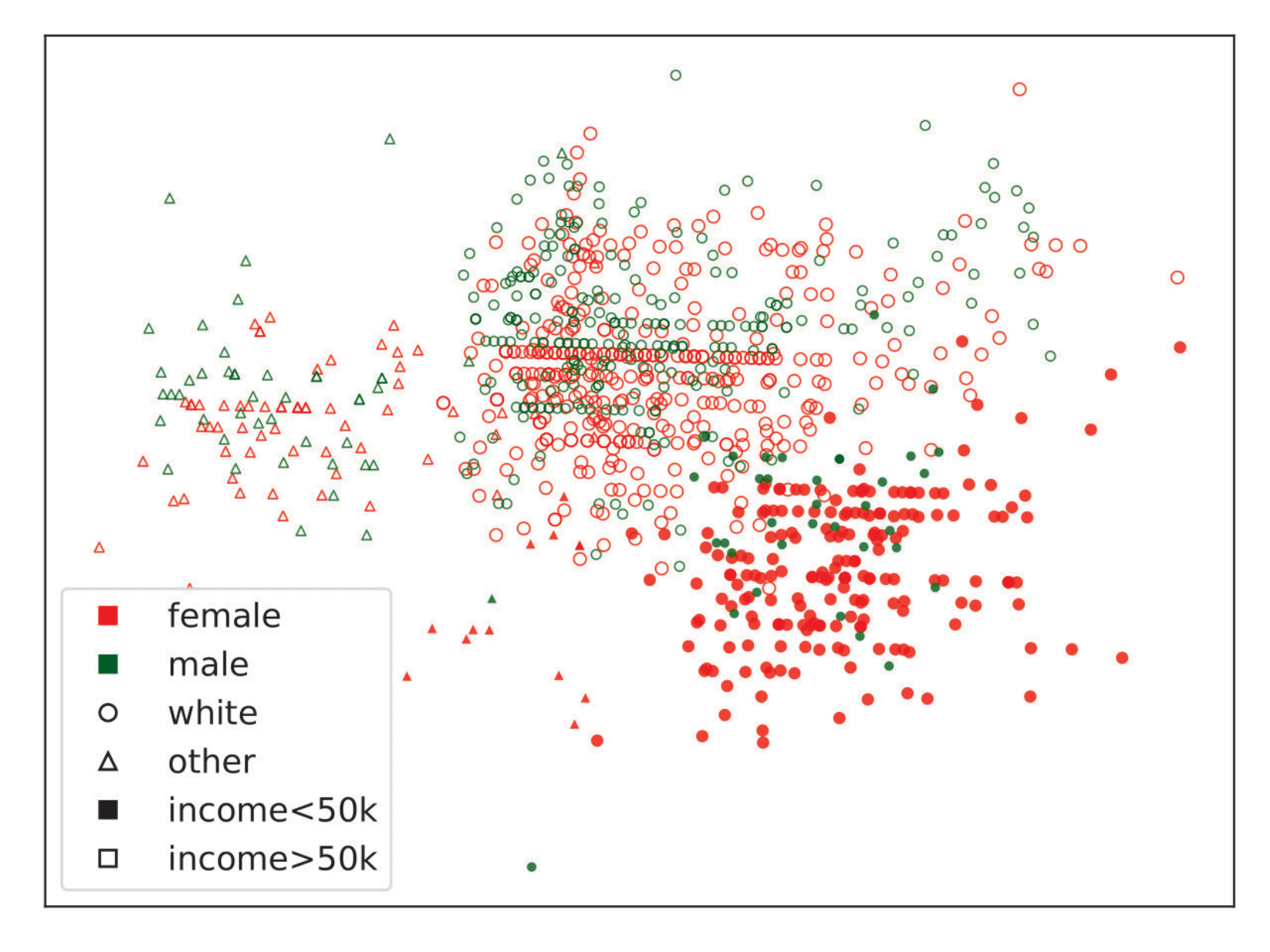}}
\subfigure[cPCA]{\label{fig:ex6-e}\includegraphics[height=1.2in,width=1.5in]{UCI_Adult_data/uci_Adult_cPCA_ethnicity.pdf}}
\subfigure[ct-SNE]{\label{fig:ex6-f}\includegraphics[height=1.2in,width=1.5in]{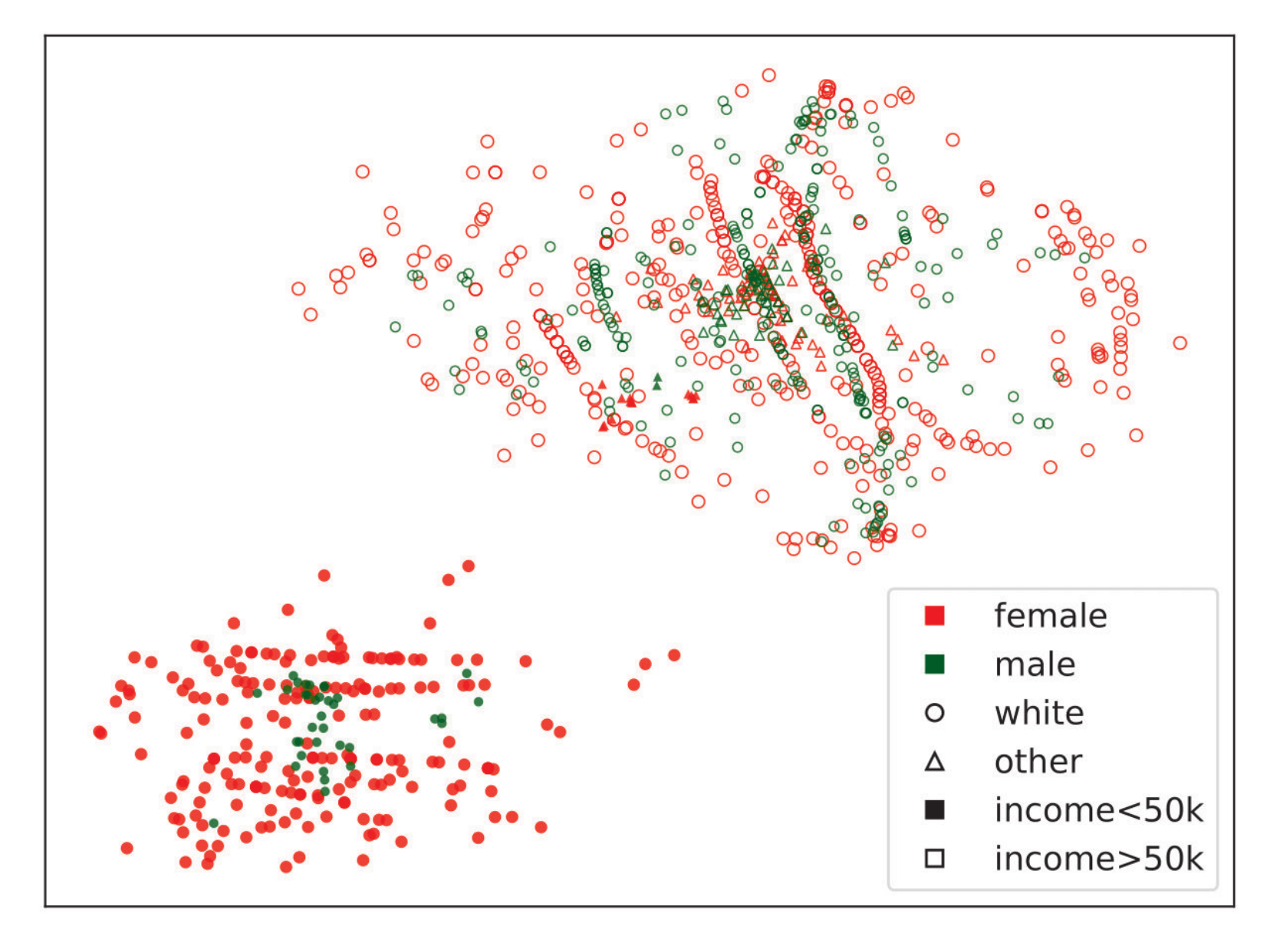}}
\subfigure[Fair-NeRV]{\label{fig:ex6-g}\includegraphics[height=1.2in,width=1.5in]{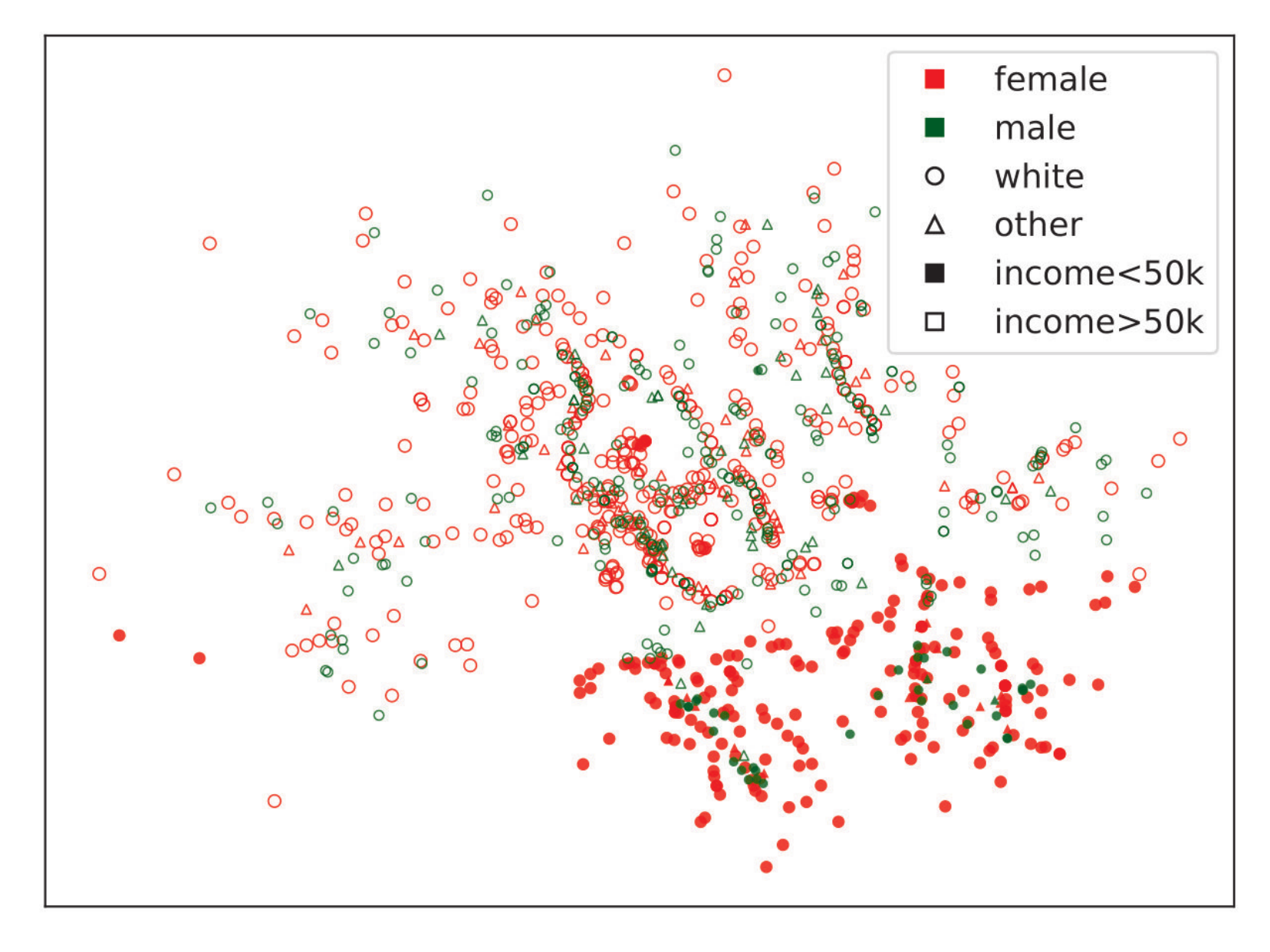}}
\subfigure[IMAPCE]{\label{fig:ex6-h}\includegraphics[height=1.2in,width=1.5in]{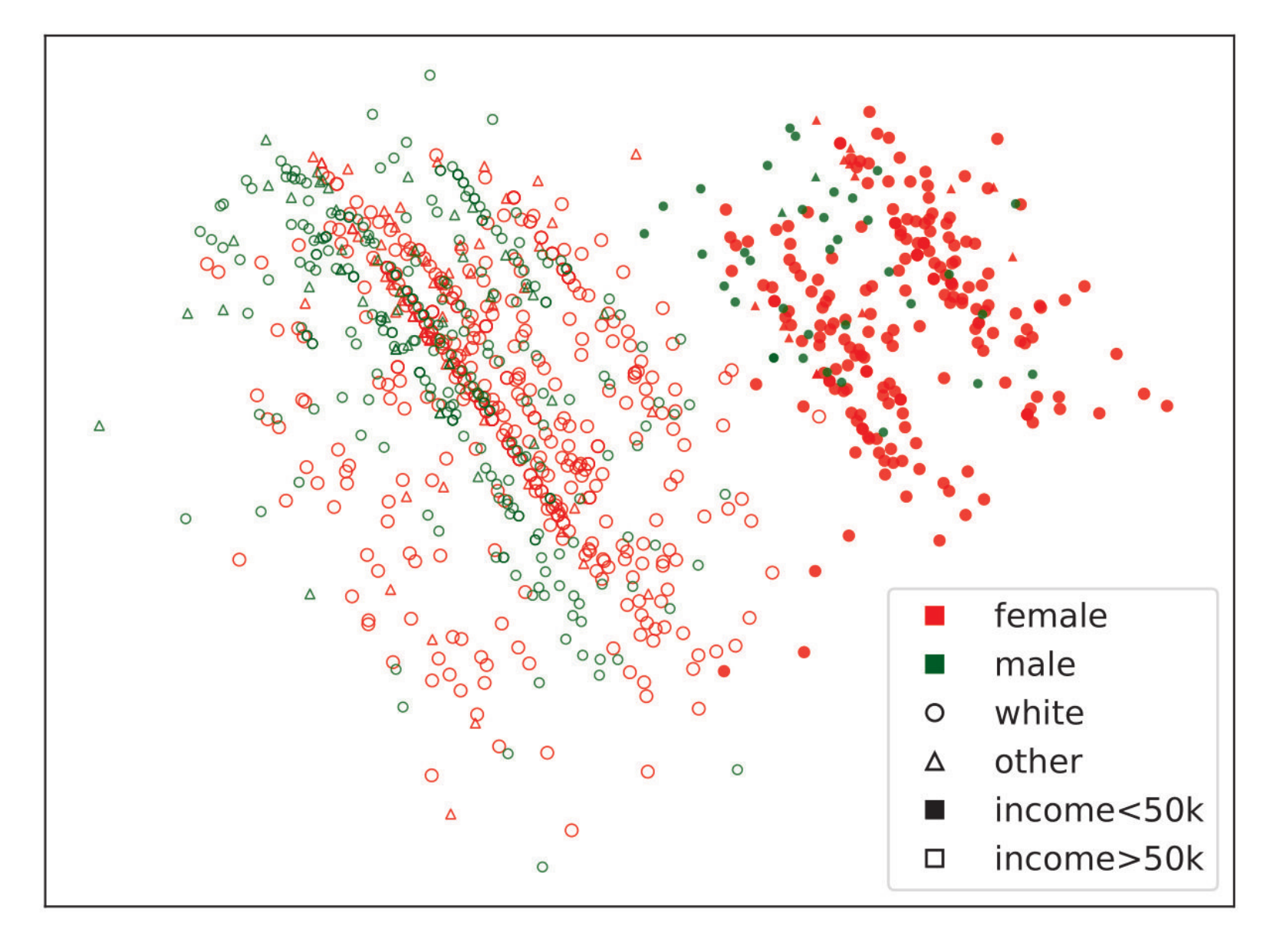}}

\caption{Visualisations of UCI adult data projections computed by cPCA, ct-SNE, Fair-NeRV and IMAPCE (a), (b), (c), (d) were computed using the gender attribute as prior. (e), (f), (g), (h) were computed using the combination of gender and ethnicity attributes as prior.}
\label{fig:UCI_adult_extra}
\end{figure*}

Using the gender attribute as prior, embeddings for cPCA, ct-SNE, Fair-NeRV and IMAPCE are shown in Figures \ref{fig:ex6-a}-\ref{fig:ex6-d}. cPCA embeddings are mixed in terms of ethnicity, gender and income, not exhibiting any apparent cluster structure. On the other hand, ct-SNE, Fair-NeRV and IMAPCE embeddings have mixed genders (an indication that prior structure has been factored out) and are clustered according to income (filled and unfilled markers) as well as ethnicity (circles and triangles), thus revealing complementary structure.

Using the combination of gender and ethnicity features as prior, we computed the embeddings of cPCA, ct-SNE, Fair-NeRV and IMAPCE and visualise them in Figures \ref{fig:ex6-e}-\ref{fig:ex6-h}. We observe that cPCA produces the same uninformative set of embeddings regardless of the selection of the prior attribute. ct-SNE, Fair-NeRV and IMAPCE compute embeddings which are clustered according to their income attribute (illustrating its structure), while their clusters consist of mixed gender and ethnicity points (indicating that the structure associated with the prior knowledge is removed). Overall, IMAPCE seems to achieve the clearest separation with respect to income.

\section{Complex MNIST+FMNIST Embeddings} \label{Supp:MFMNIST}
\begin{figure*}[ht]
\centering
% \vspace*{0.75cm}
\subfigure[cPCA]{\label{fig:ex8-a}\includegraphics[height=1.5in,width=1.7in]{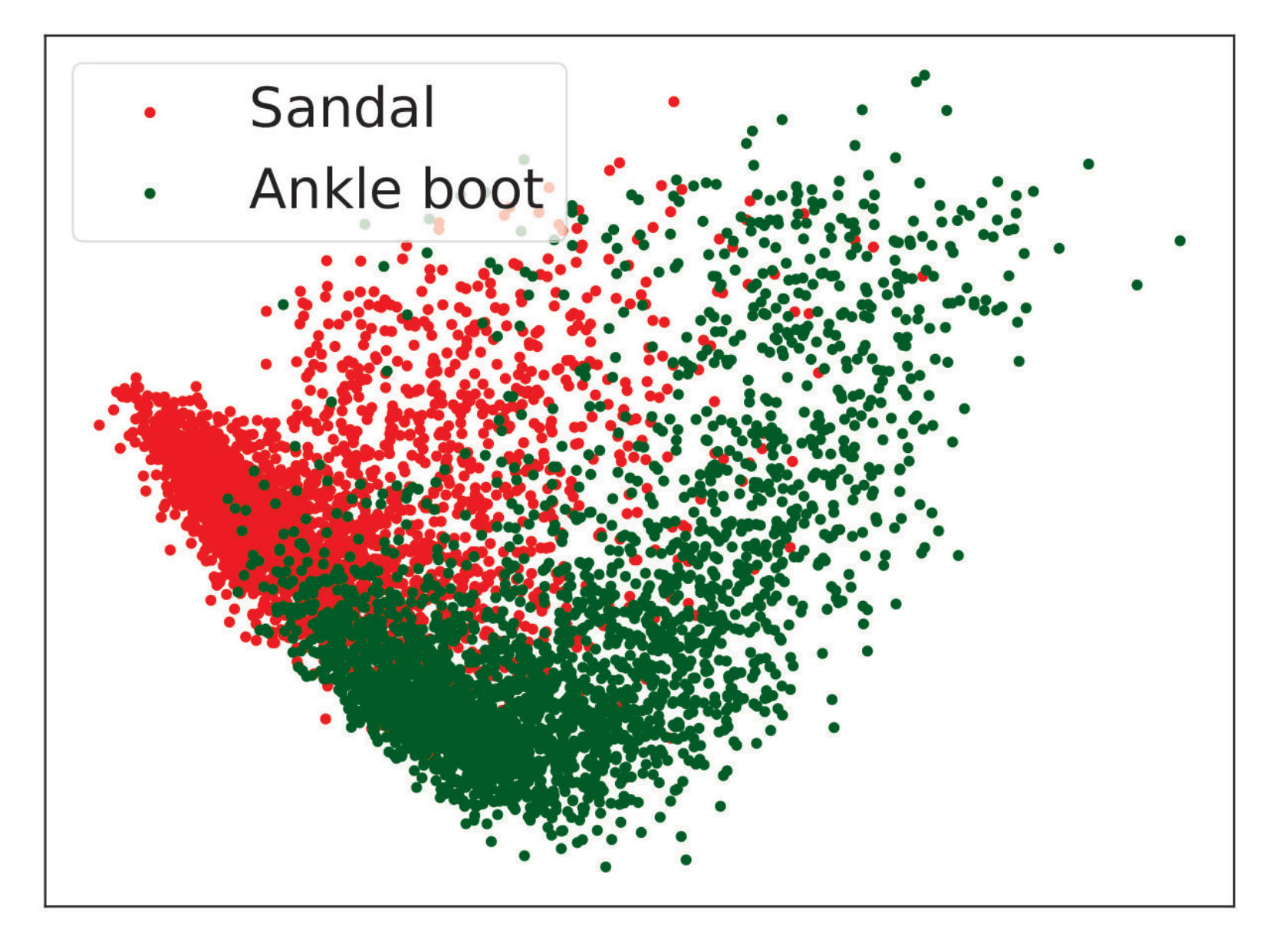}}
\subfigure[ct-SNE]{\label{fig:ex8-b}\includegraphics[height=1.5in,width=1.7in]{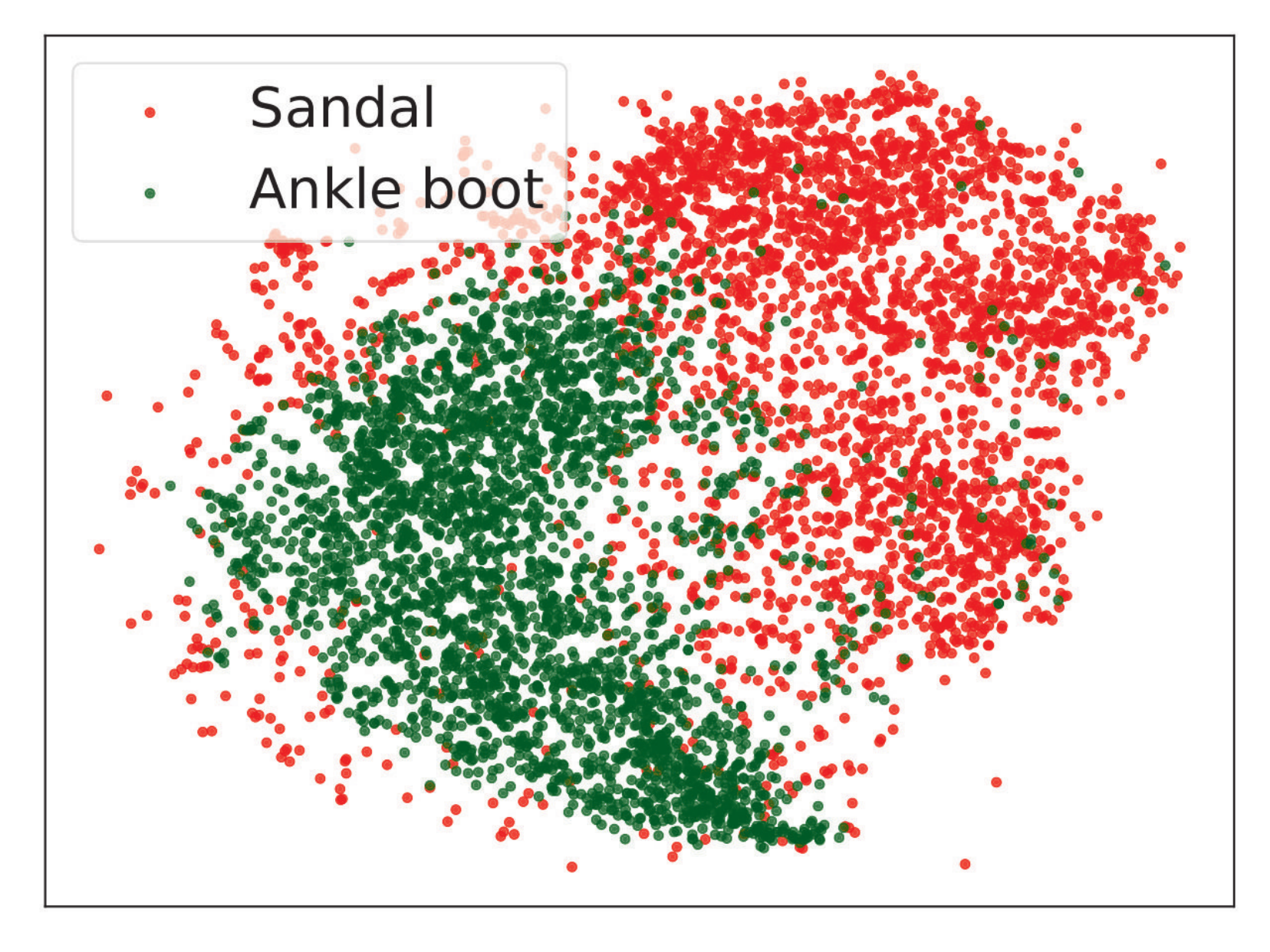}}
\subfigure[Fair-NeRV]{\label{fig:ex8-c}\includegraphics[height=1.5in,width=1.7in]{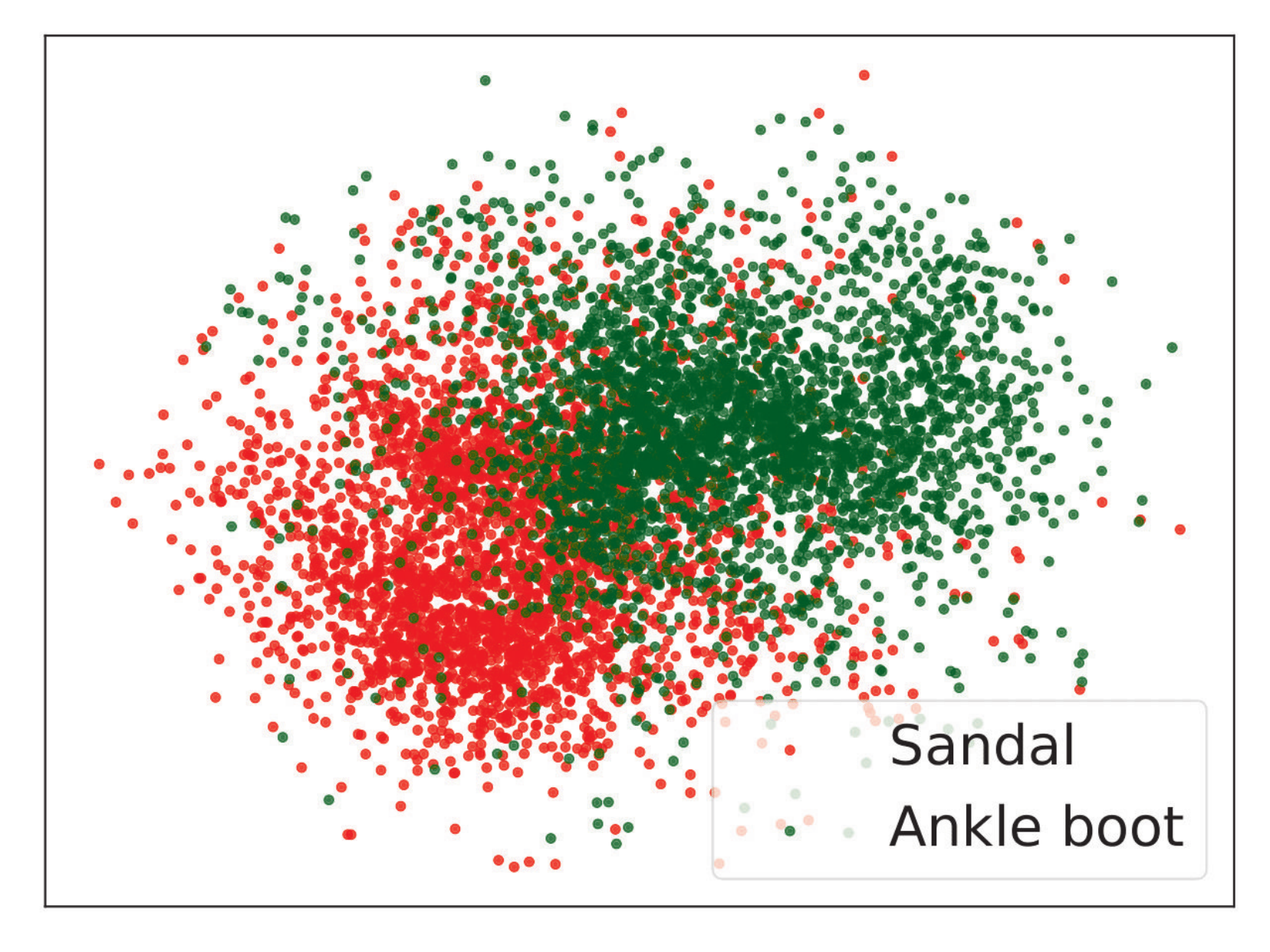}}
\subfigure[IMAPCE]{\label{fig:ex8-d}\includegraphics[height=1.5in,width=1.7in]{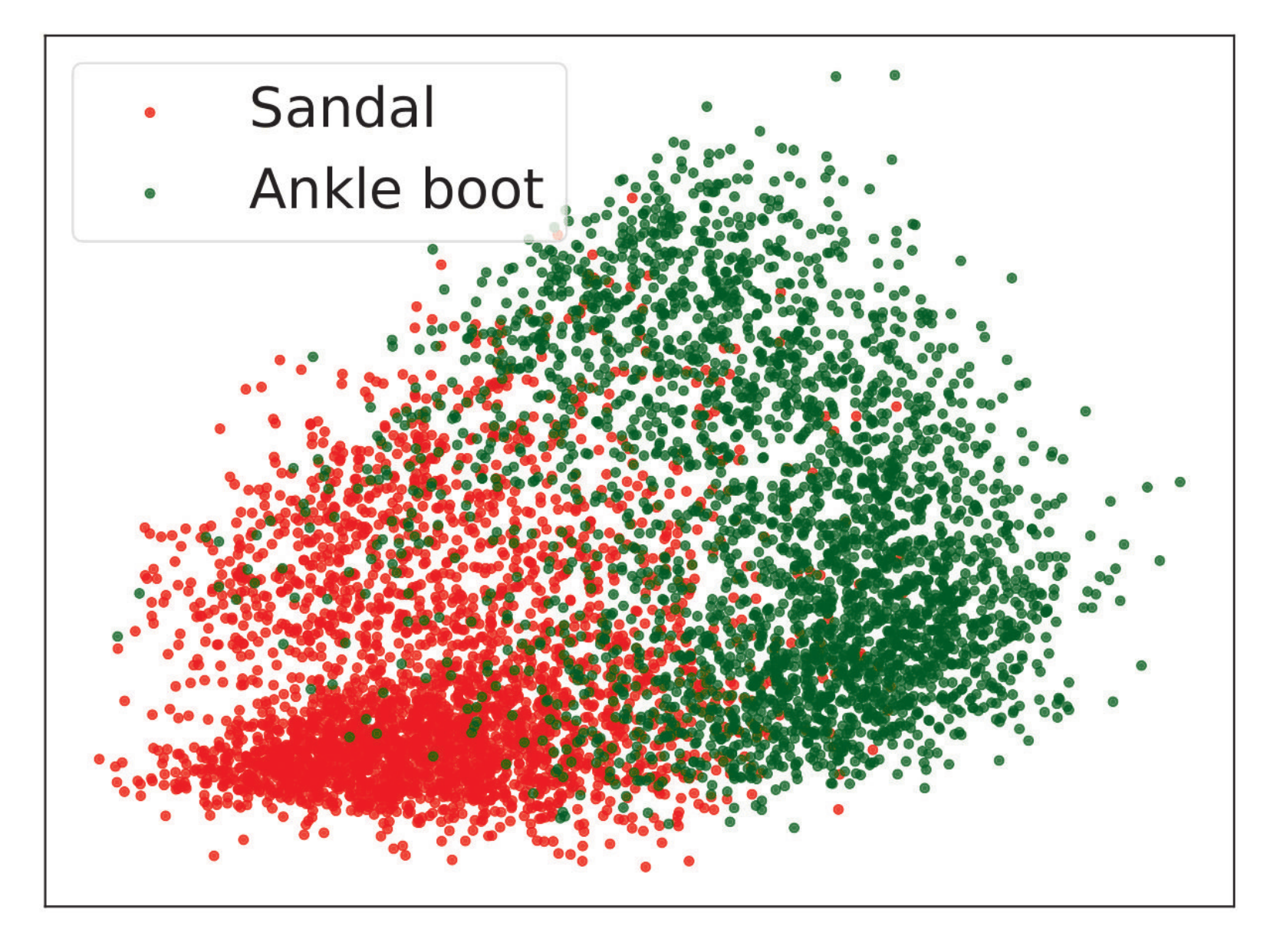}}
\subfigure[cPCA]{\label{fig:ex8-e}\includegraphics[height=1.5in,width=1.7in]{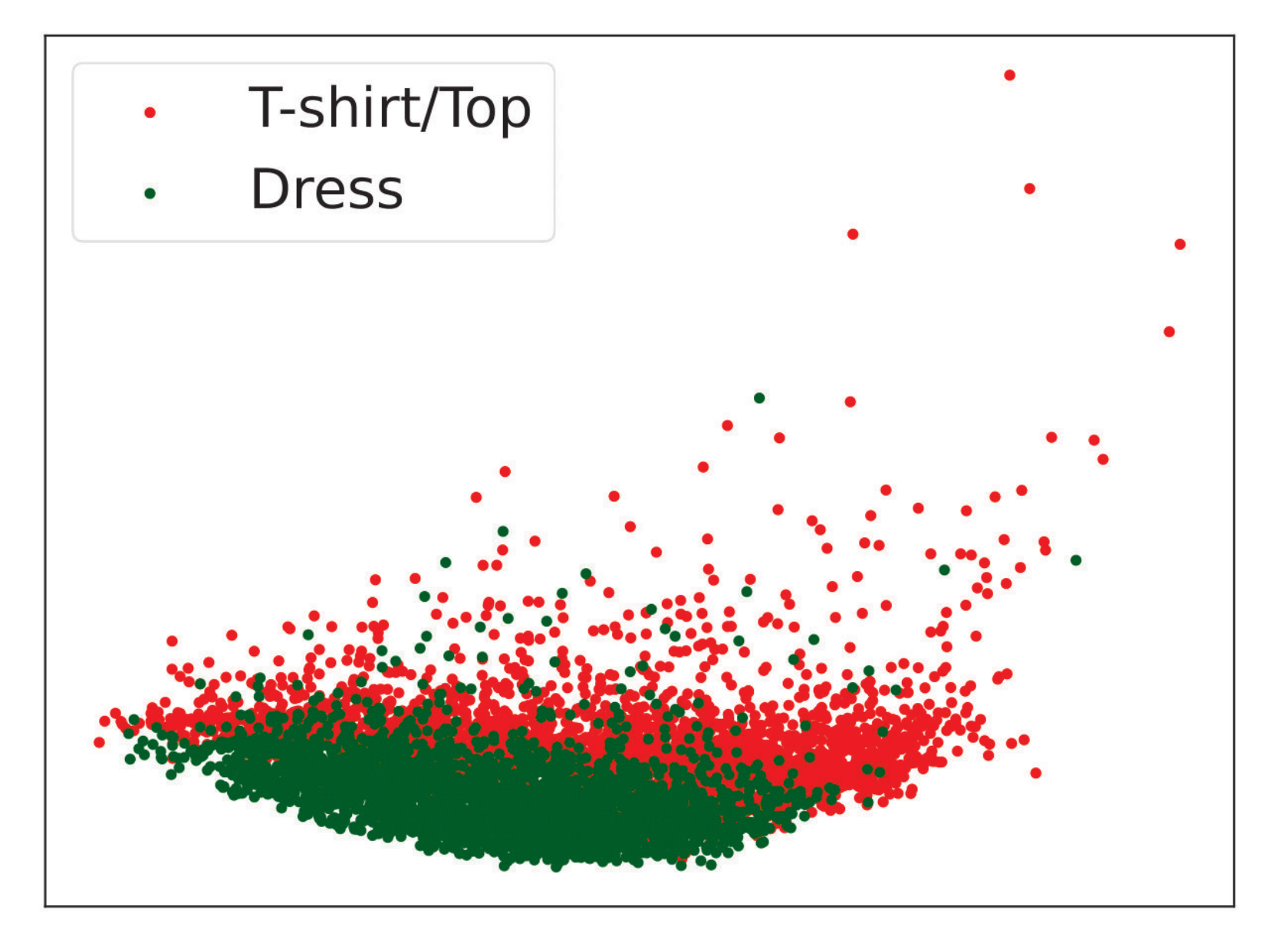}}
\subfigure[ct-SNE]{\label{fig:ex8-f}\includegraphics[height=1.5in,width=1.7in]{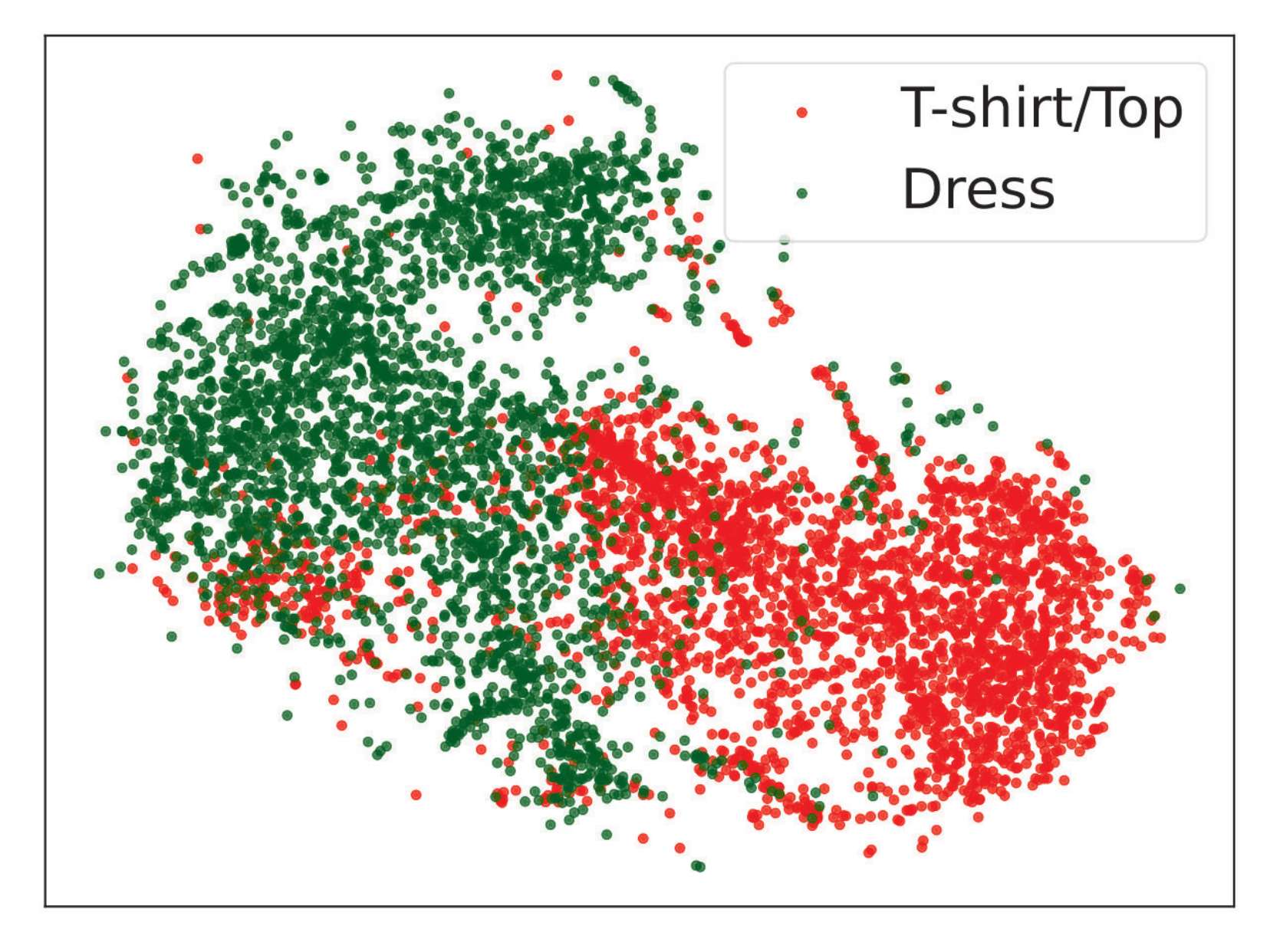}}
\subfigure[Fair-NeRV]{\label{fig:ex8-g}\includegraphics[height=1.5in,width=1.7in]{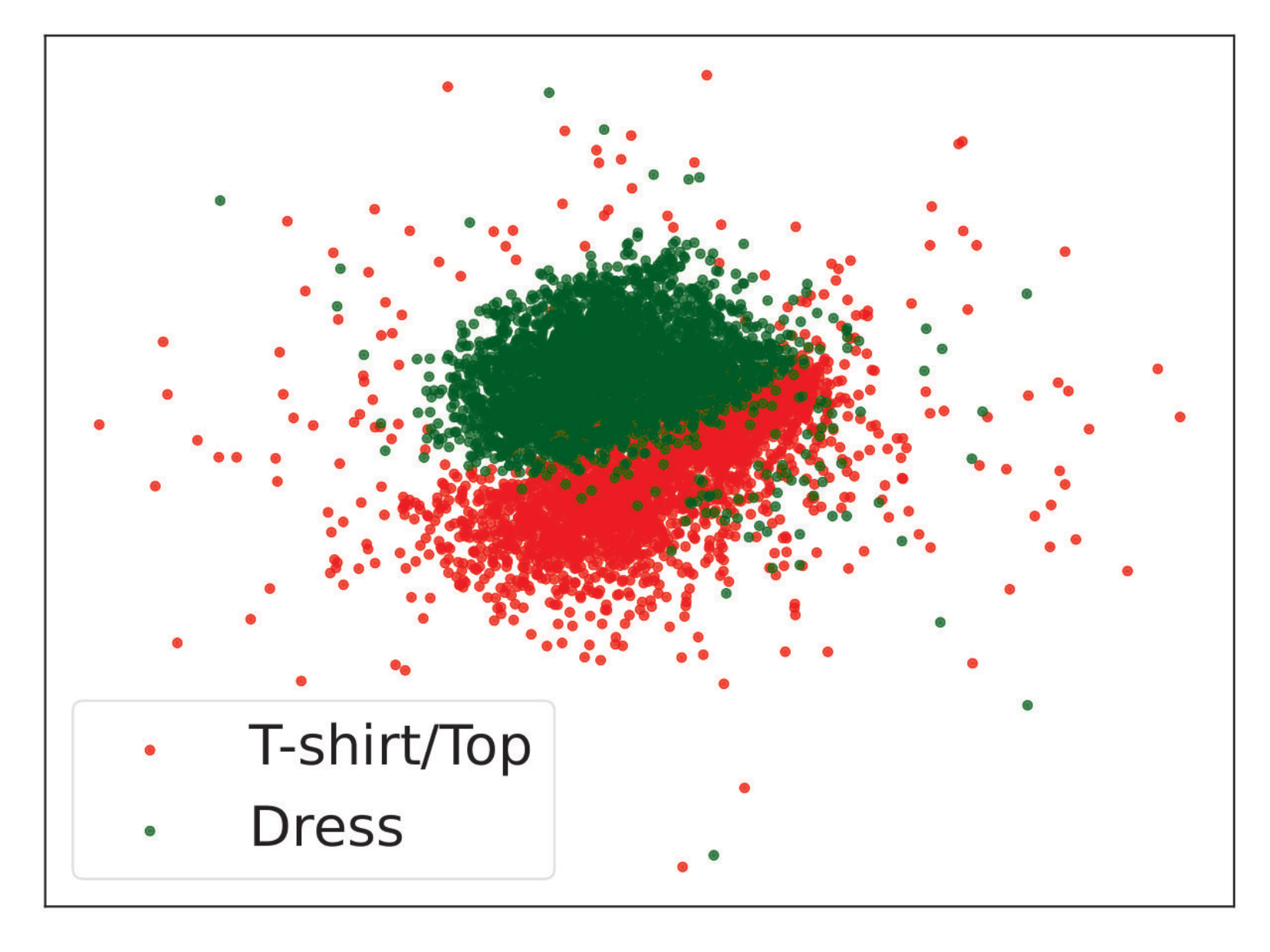}}
\subfigure[IMAPCE]{\label{fig:ex8-h}\includegraphics[height=1.5in,width=1.7in]{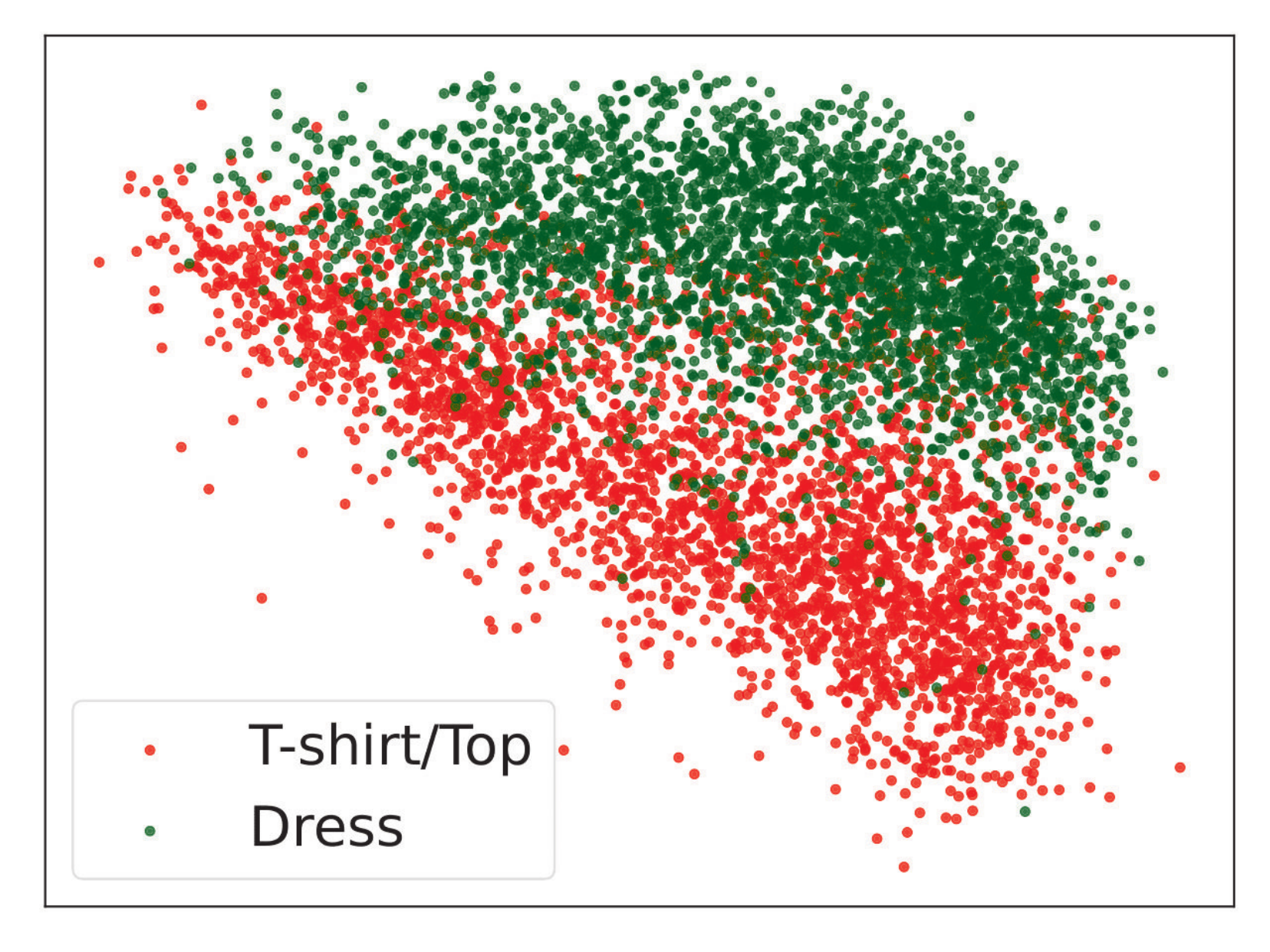}}
\subfigure["5"]{\label{fig:ex8-i}\includegraphics[height=1.3in,width=1.5in]{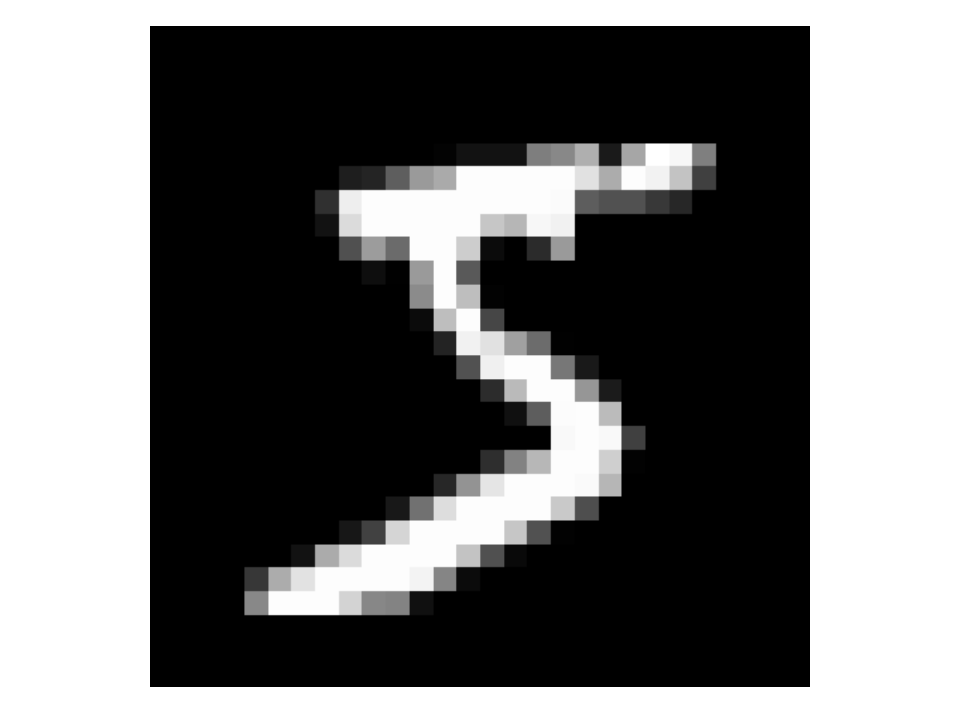}}
\subfigure["Sneaker"]{\label{fig:ex8-j}\includegraphics[height=1.3in,width=1.5in]{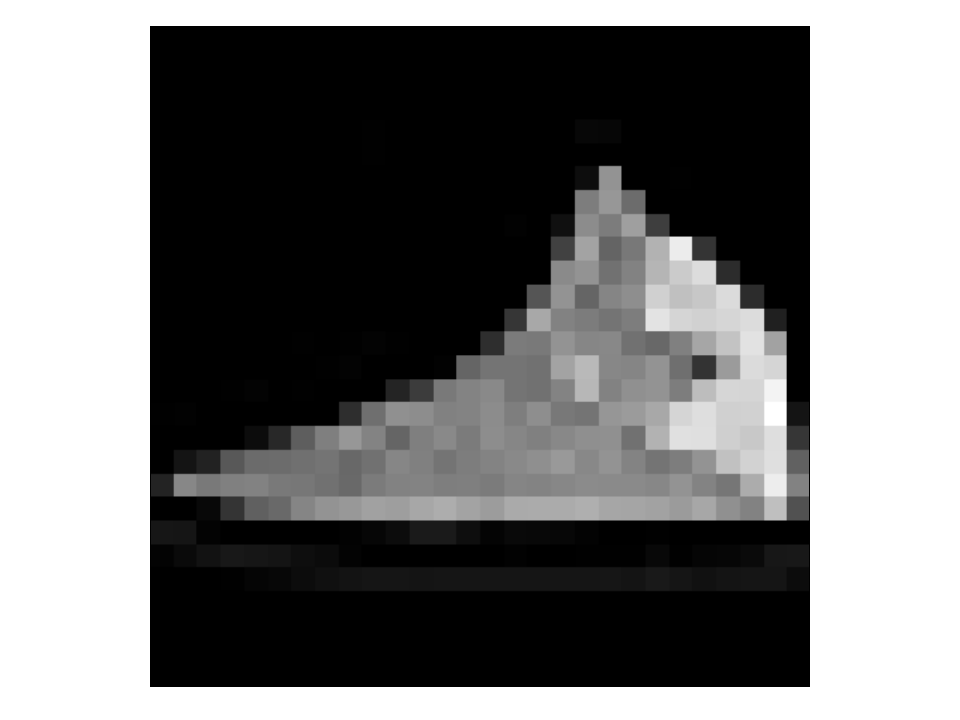}}
\subfigure["5" + "Sneaker"]{\label{fig:ex8-k}\includegraphics[height=1.3in,width=1.5in]{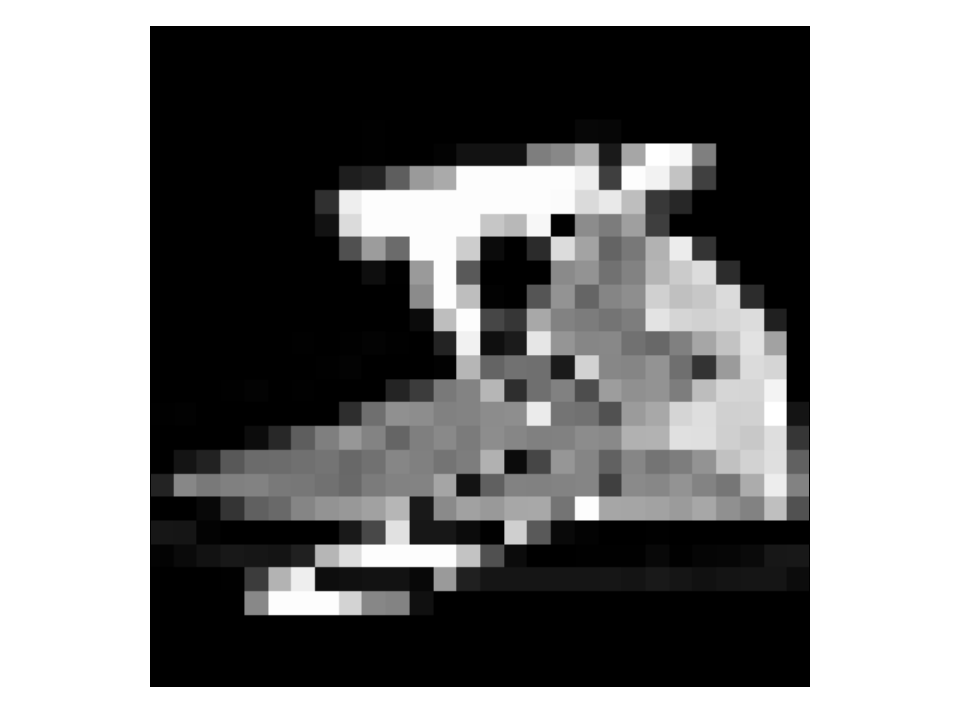}} \\ 
\subfigure["3"]{\label{fig:ex8-l}\includegraphics[height=1.3in,width=1.5in]{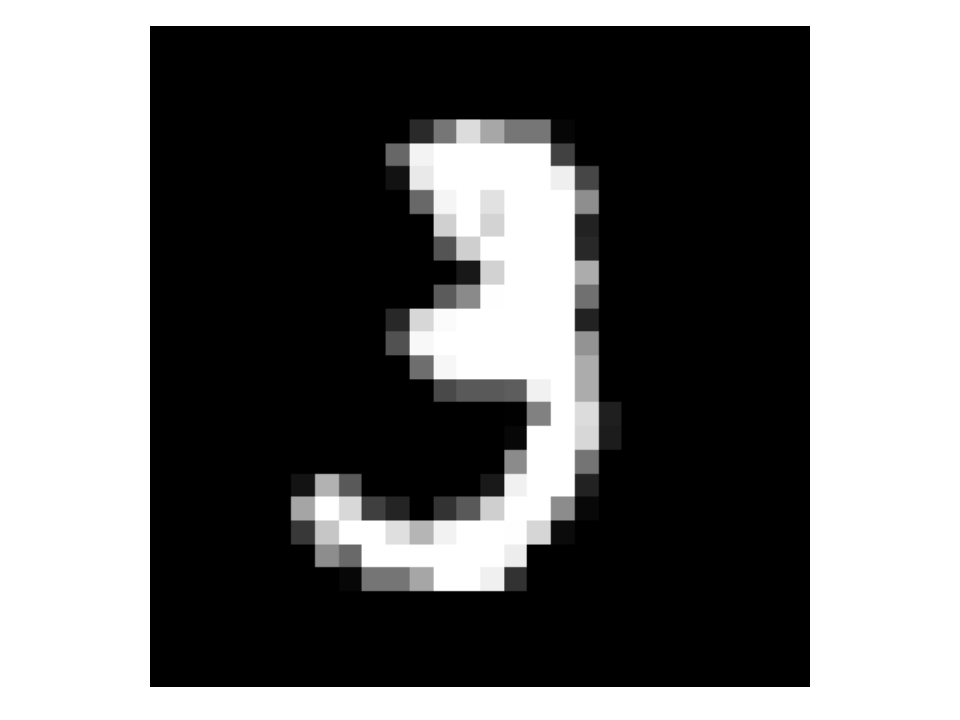}}
\subfigure["Sandal"]{\label{fig:ex8-m}\includegraphics[height=1.3in,width=1.5in]{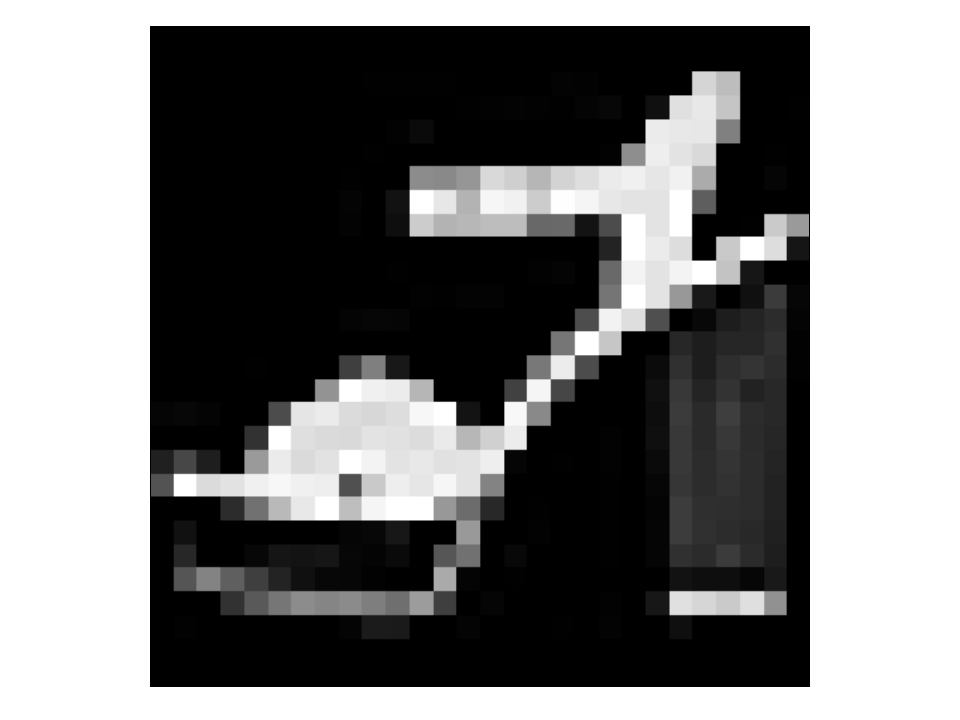}}
\subfigure["3" + "Sandal"]{\label{fig:ex8-n}\includegraphics[height=1.3in,width=1.5in]{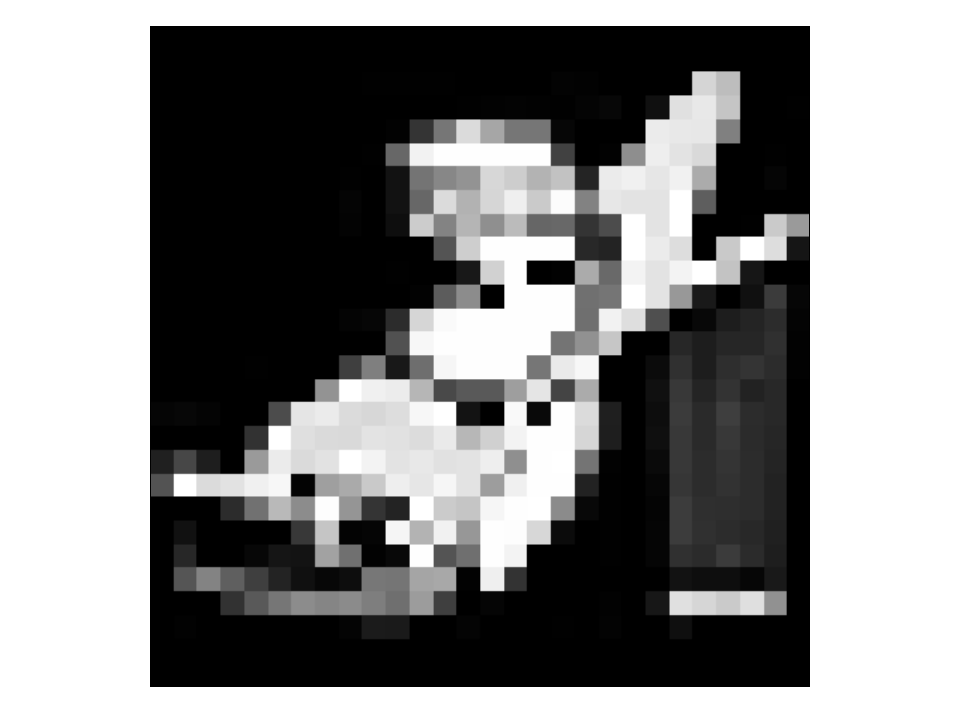}}
\caption{Superimposed MNIST + FMNIST embeddings computed by all methods for different combinations of FMNIST classes. (a), (b), (c), (d) for "Sandal"-"Ankle boot" case, (e), (f), (g), (h) for "Tshirt"-"Dress". (i), (l) MNIST instances, (j), (m) FMNIST instances, (k), (n) their superimposition results.}
\label{fig:MNIST-FMNIST_embeddings}
\end{figure*}

We constructed two more complex MNIST + FMNIST data cases by i) superimposing FMNIST instances of "Sandal" and "Ankle boot" with random MNIST digits and ii) FMNIST instances of "Tshirt" and "Dress" with random MNIST digits. We provide additional MNIST, FMNIST instances and their superimposition results in Figure \ref{fig:MNIST-FMNIST_embeddings}. 

As also mentioned previously, we want to compute embeddings that are clustered according to their FMNIST part, while not exhibiting any cluster formation w.r.t. their MNIST part. For cPCA and IMAPCE, we define as prior data 1,000 randomly MNIST images (not used when constructing the complex data). These can capture the structure of the MNIST part we wish to remove from the embeddings. ct-SNE and Fair-NeRV, which require the prior information in terms of discrete labels, are provided with the specific MNIST labels of the instances we used when constructing the respective complex data. 

Embeddings computed by all methods are shown in Figure \ref{fig:MNIST-FMNIST_embeddings}. We observe that, all methods provide embeddings which are to some extent separated according to their FMNIST classes. However, in both cases, IMAPCE's embeddings seem to have the lowest overlap with respect to their FMNIST class.

Linear and SVM classifiers were trained (ten random train 75\% -test 25 \% splits) on the FMNIST classification of the 2D embeddings for each case. The average test-set accuracy of the linear classifier for each method and case is shown in Table \ref{table:mnist-fmnist-linear}, while the average test-set accuracy, F1 and Recall for the SVM classifiers are shown in Table \ref{table:mnist-fmnist-svm}. IMAPCE achieves the best accuracy in both cases, while taking comparable training time with ct-SNE and cPCA and much lower than Fair-NeRV. Finally, as also mentioned previously, IMAPCE does not consider the MNIST labels used in the construction of the complex data, but rather a few representative MNIST instances which adequately capture the general MNIST variation we wish to remove from the embeddings.

\begin{table}[ht]
\caption{Test-set (averaged over ten random train-test splits) accuracy for Linear classification of the complex MNIST + FMNIST 2D embeddings computed by IMAPCE, cPCA, ct-SNE and FAIR-NeRV with respect to their FMNIST ground truth labels.}
\begin{center}
\begin{tabular}{c c c c c} 
 \toprule
 FMNIST classes & cPCA & ct-SNE & Fair-NeRV & IMAPCE \\ 
 \midrule
 "Tshirt"-"Dress" & 0.83 & 0.85 & 0.86 & \textbf{0.88} \\
  \midrule
 "Sandal"-"Ankle boot" & 0.87 & 0.85 & 0.81 & \textbf{0.89} \\
\bottomrule
\end{tabular}
\label{table:mnist-fmnist-linear}
\end{center}
\end{table}

\section{Complex CIFAR-100+FMNIST Embeddings} \label{Supp:C100-FMNIST}
We constructed two additional complex data cases by combining CIFAR-100 images with FMNIST images. More specifically, we superimposed i) 6,000 random FMNIST samples having "T-shirt/Top" and "Shirt" classes with randomly chosen CIFAR-100 samples and ii) 6,000 random FMNIST samples having "T-shirt/Top" and "Coat" classes with randomly selected CIFAR-100 samples. 

\begin{figure*}[ht]
\centering
% \vspace*{0.75cm}
\subfigure[cPCA]{\label{fig:ex7-a}\includegraphics[height=1.5in,width=1.7in]{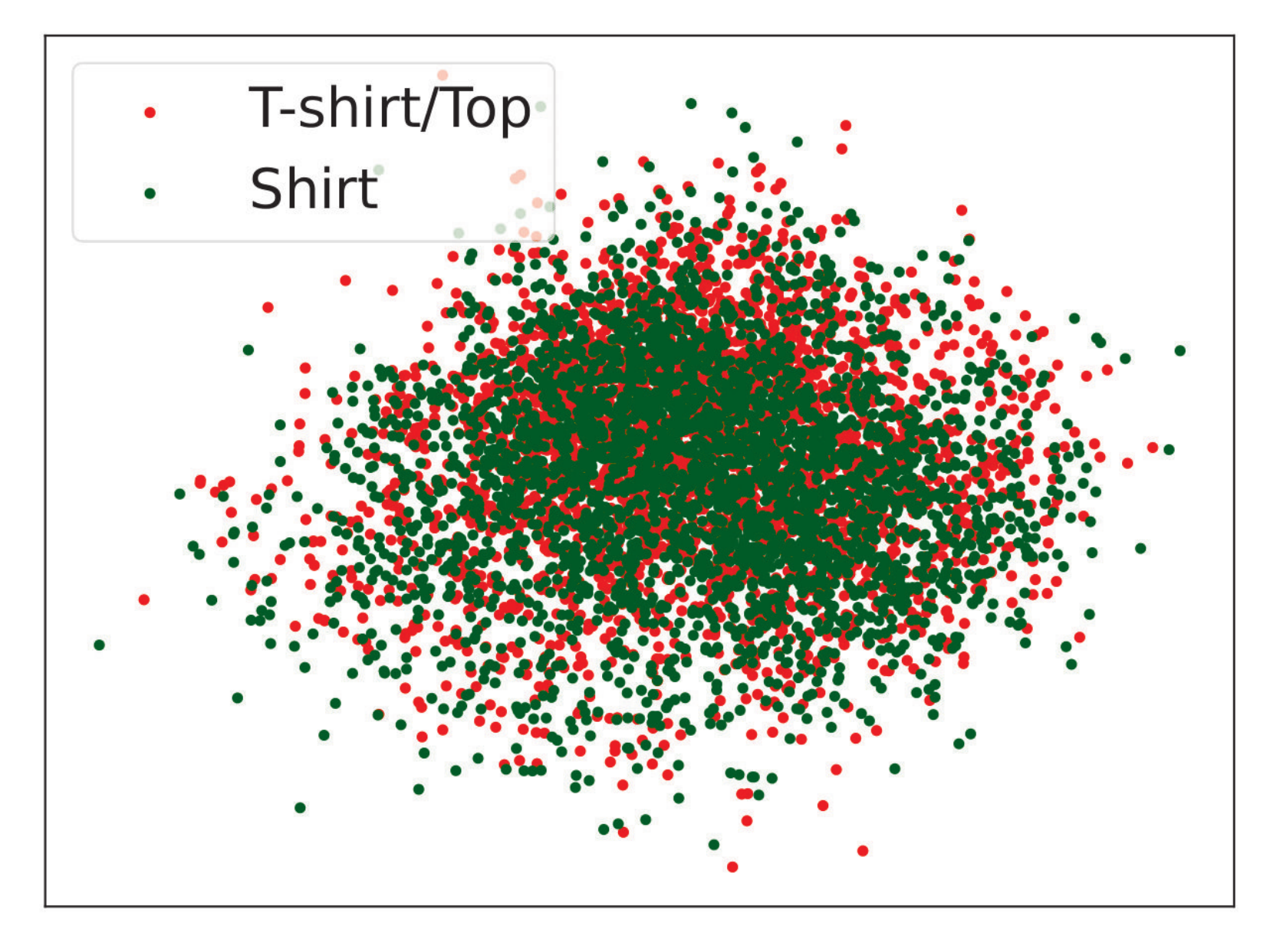}}
\subfigure[ct-SNE]{\label{fig:ex7-b}\includegraphics[height=1.5in,width=1.7in]{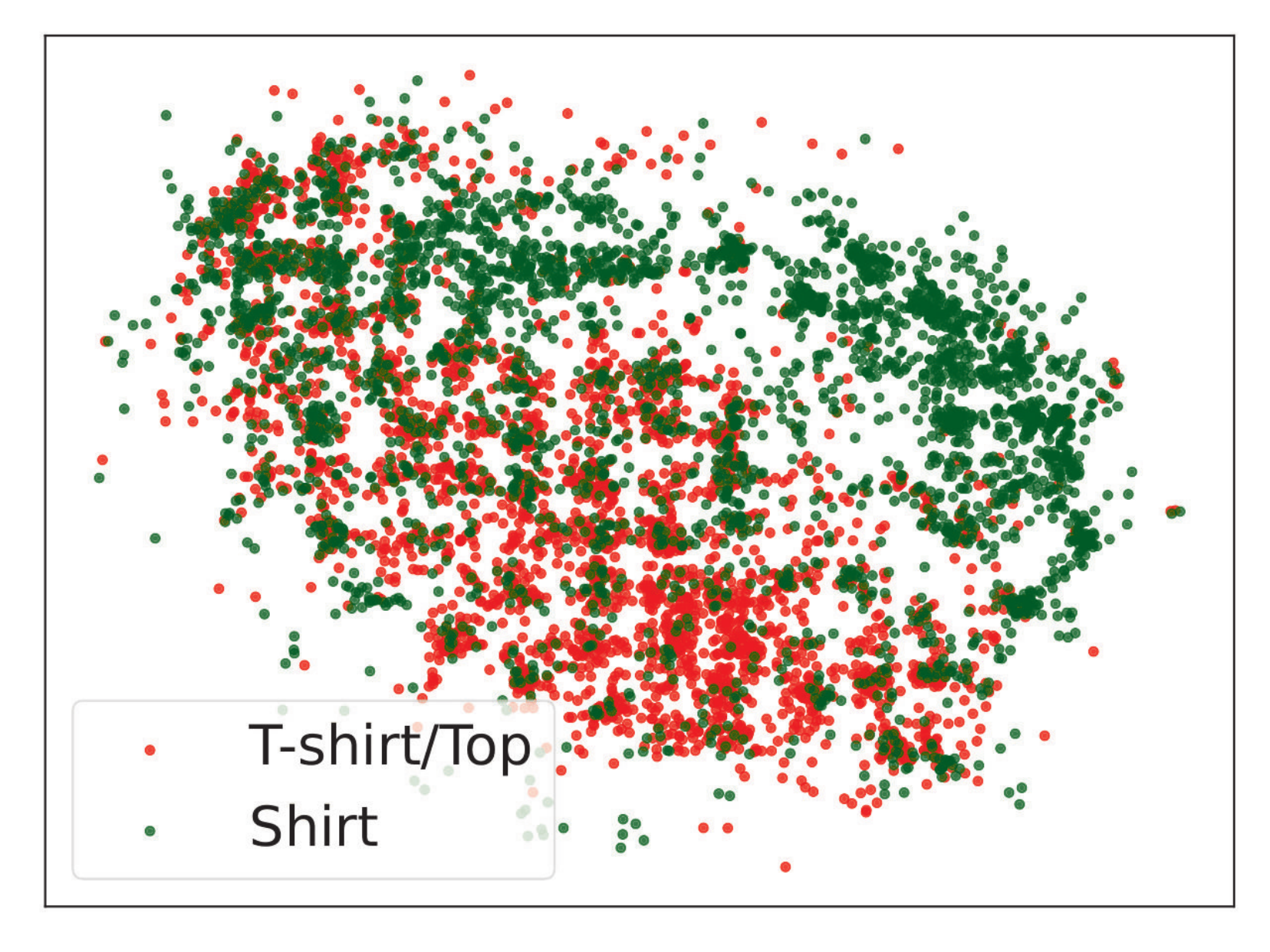}}
\subfigure[Fair-NeRV]{\label{fig:ex7-c}\includegraphics[height=1.5in,width=1.7in]{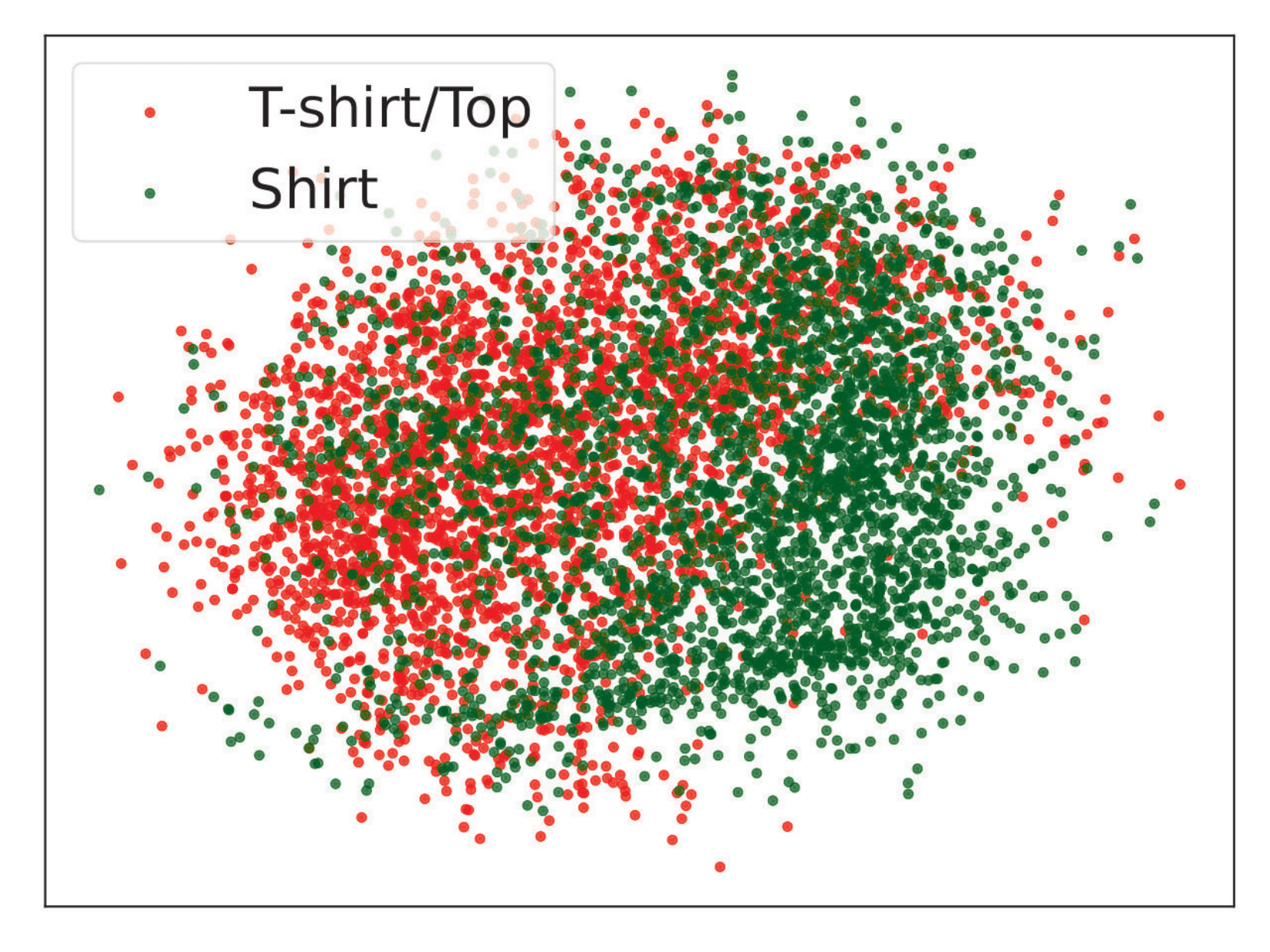}}
\subfigure[IMAPCE]{\label{fig:ex7-d}\includegraphics[height=1.5in,width=1.7in]{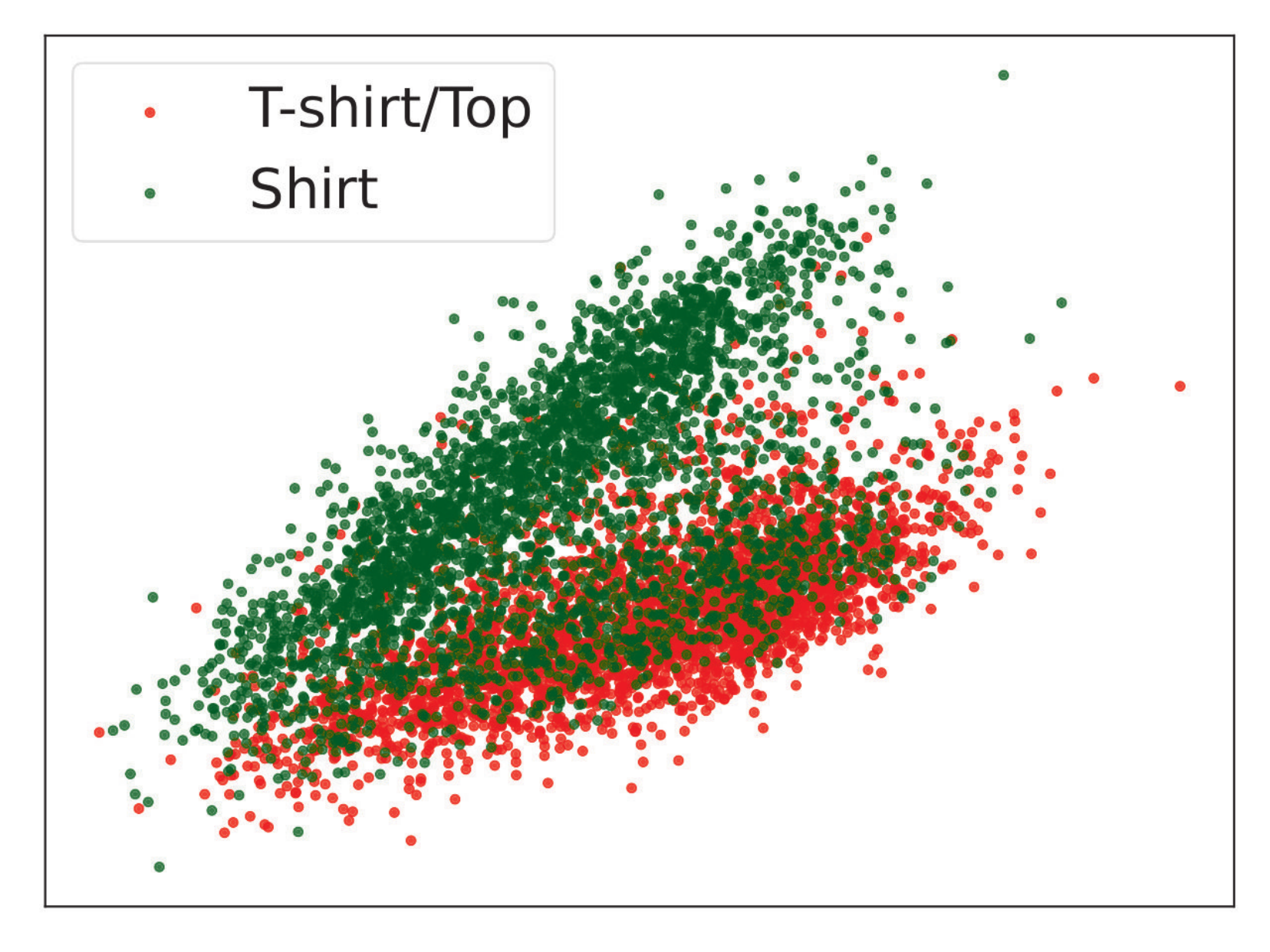}}
\subfigure[cPCA]{\label{fig:ex7-e}\includegraphics[height=1.5in,width=1.7in]{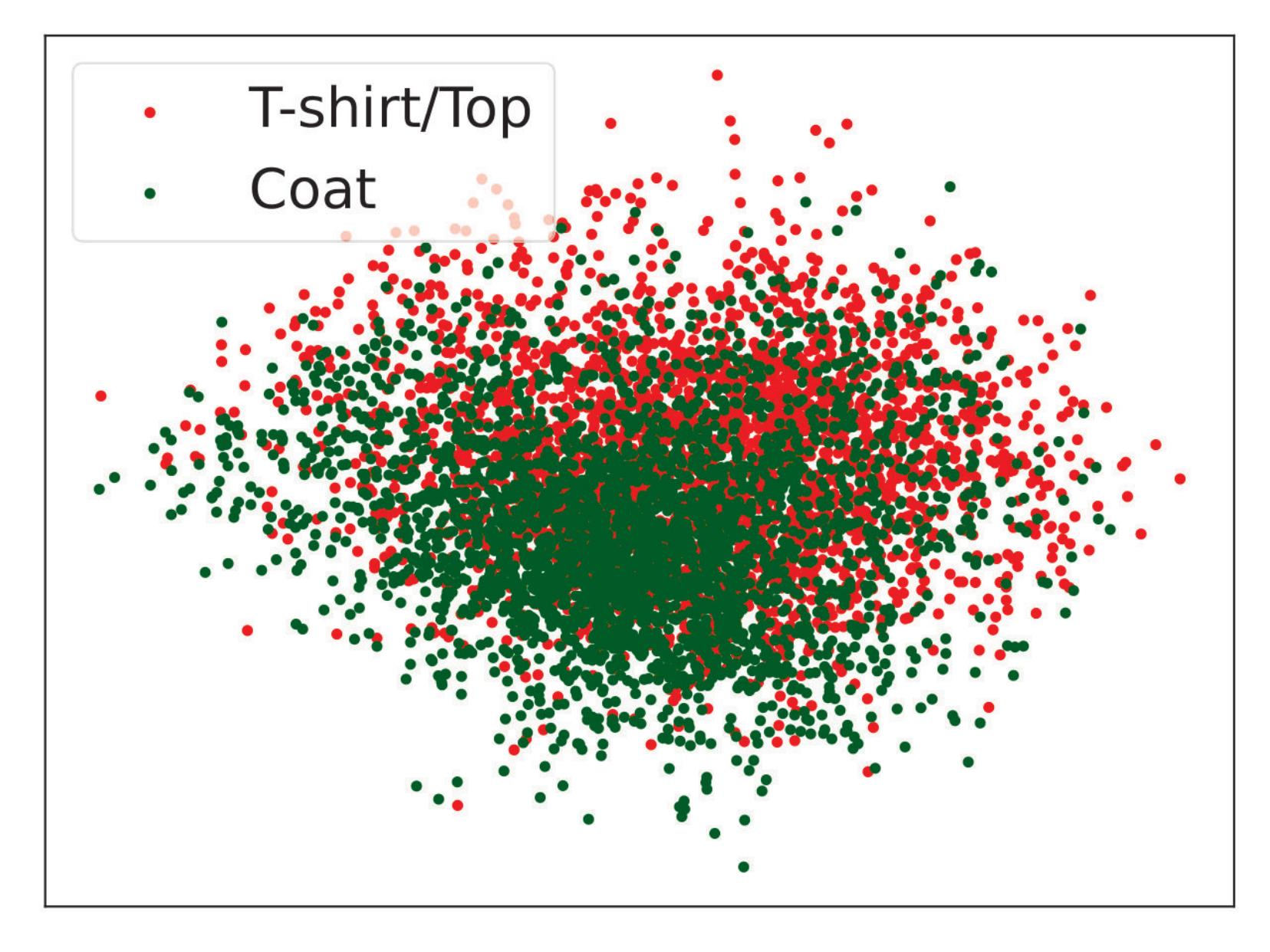}}
\subfigure[ct-SNE]{\label{fig:ex7-f}\includegraphics[height=1.5in,width=1.7in]{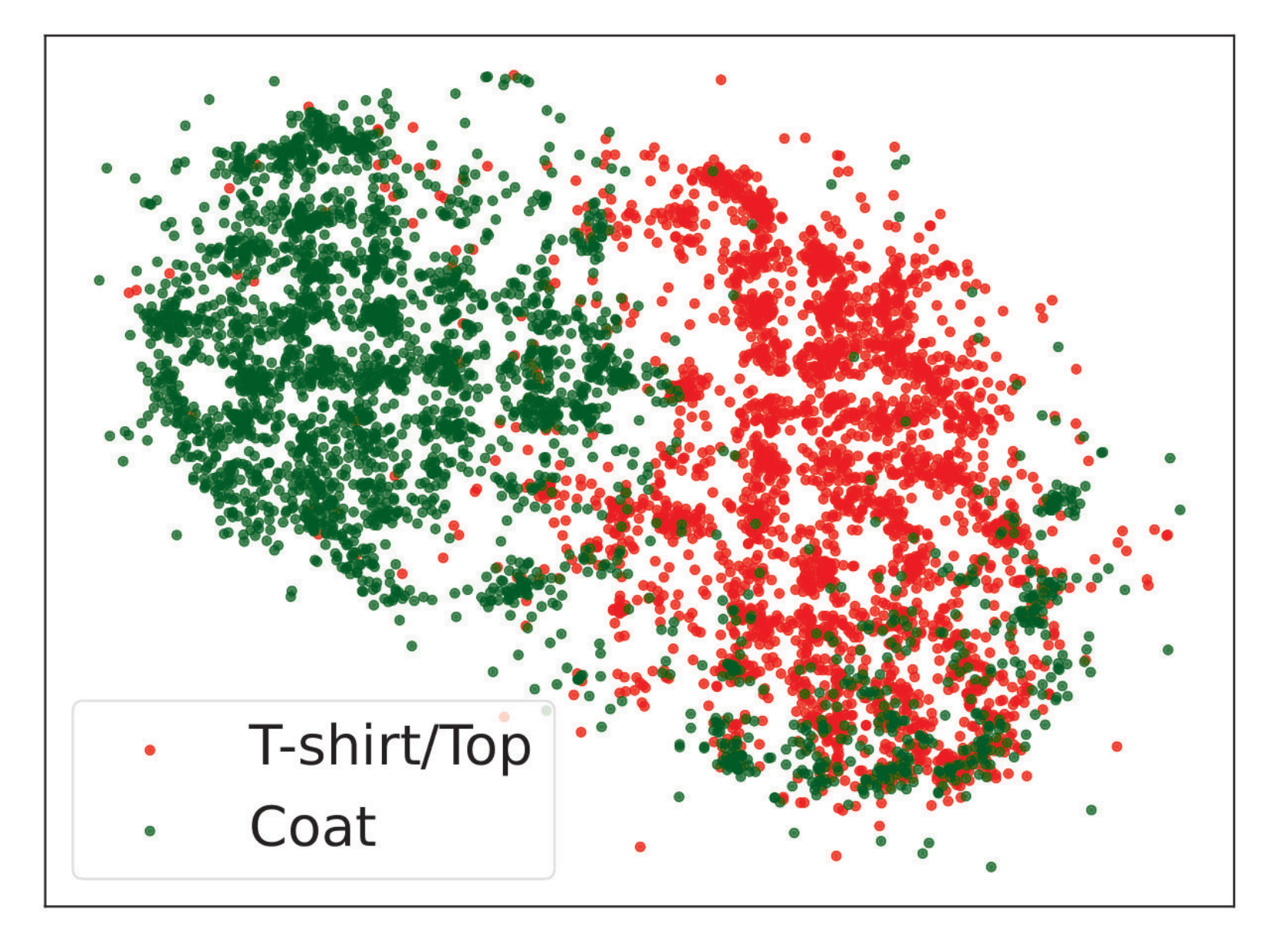}}
\subfigure[Fair-NeRV]{\label{fig:ex7-g}\includegraphics[height=1.5in,width=1.7in]{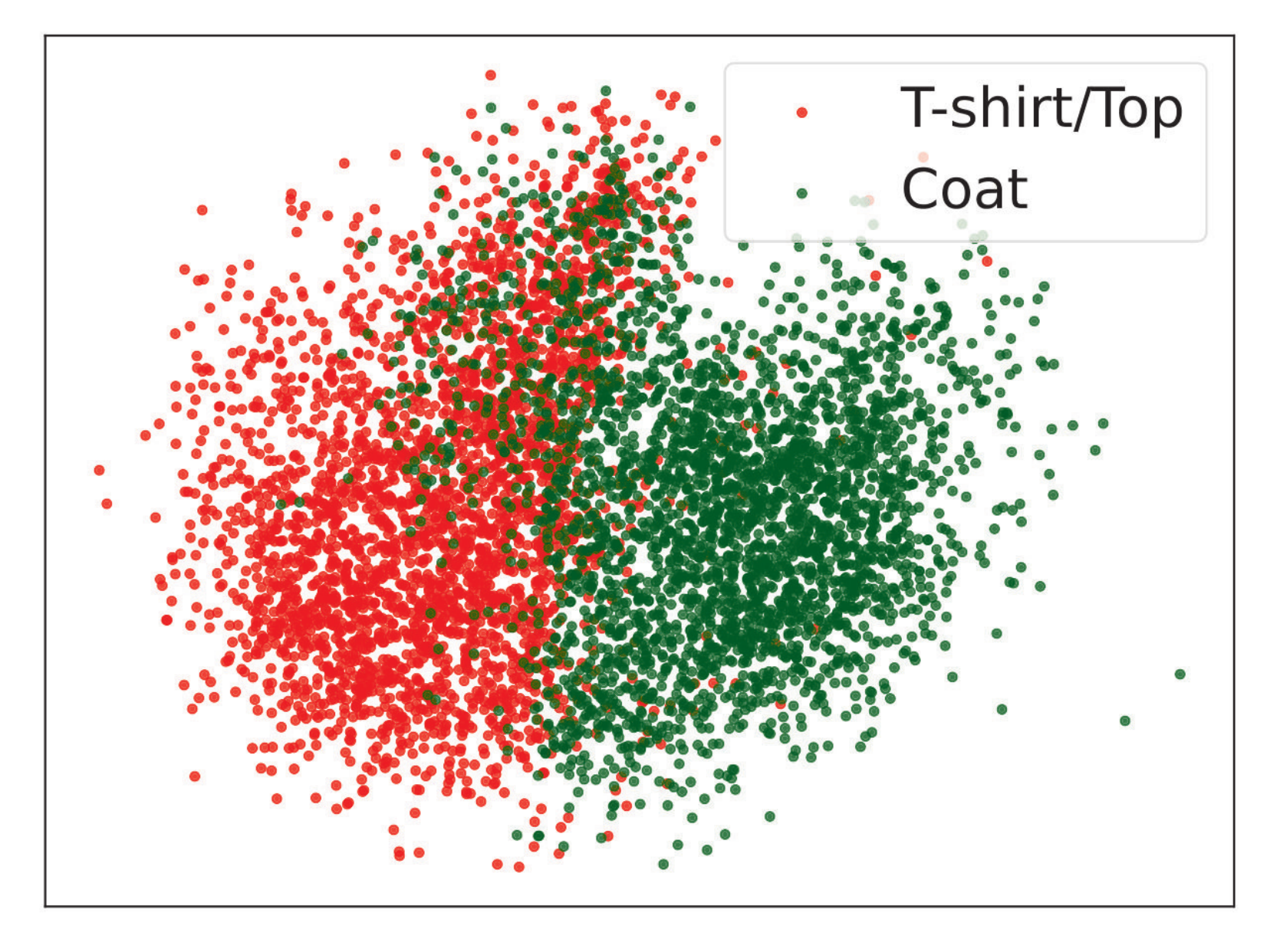}}
\subfigure[IMAPCE]{\label{fig:ex7-h}\includegraphics[height=1.5in,width=1.7in]{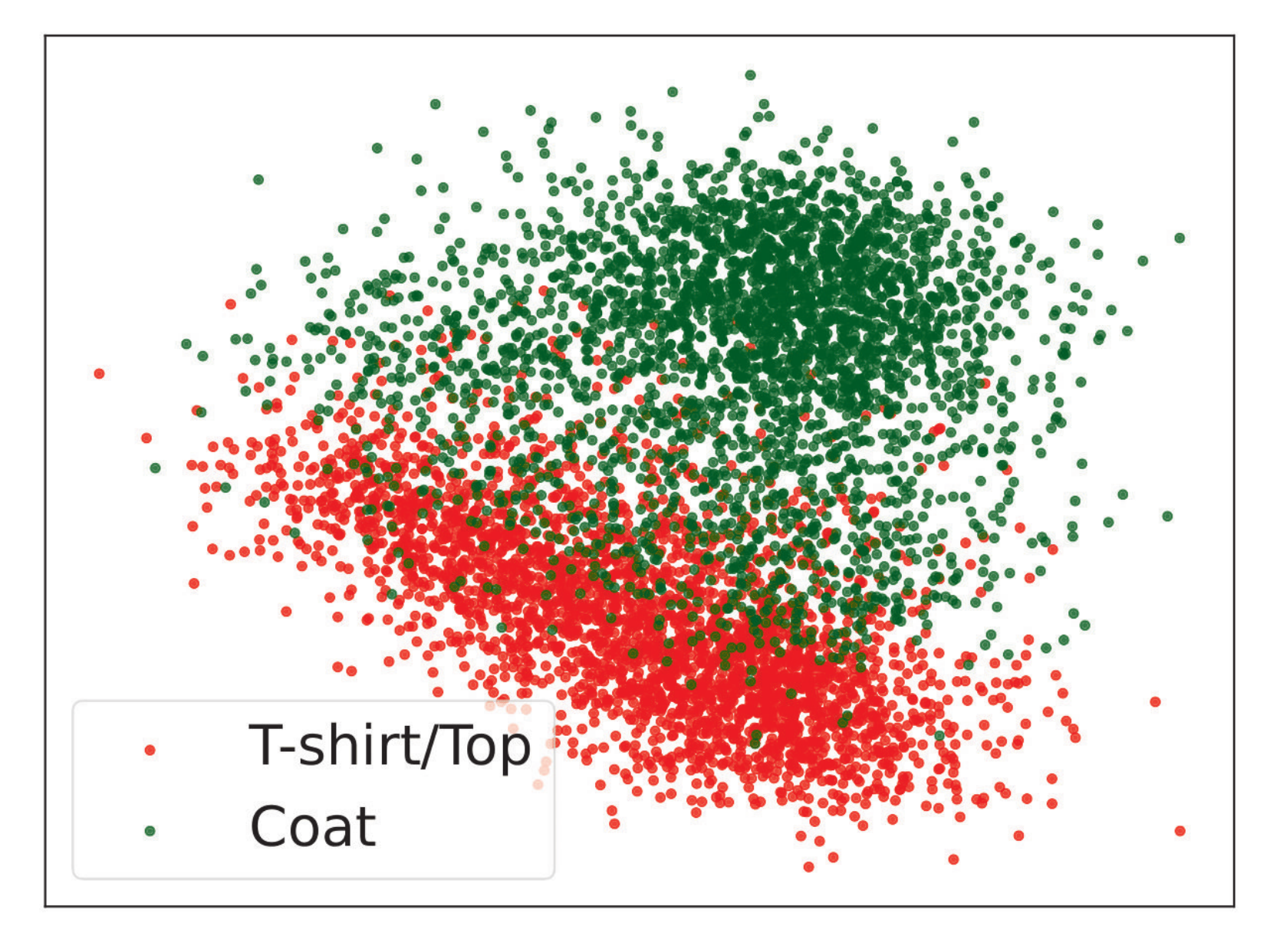}}
\subfigure["Cow"]{\label{fig:ex7-i}\includegraphics[height=1.3in,width=1.5in]{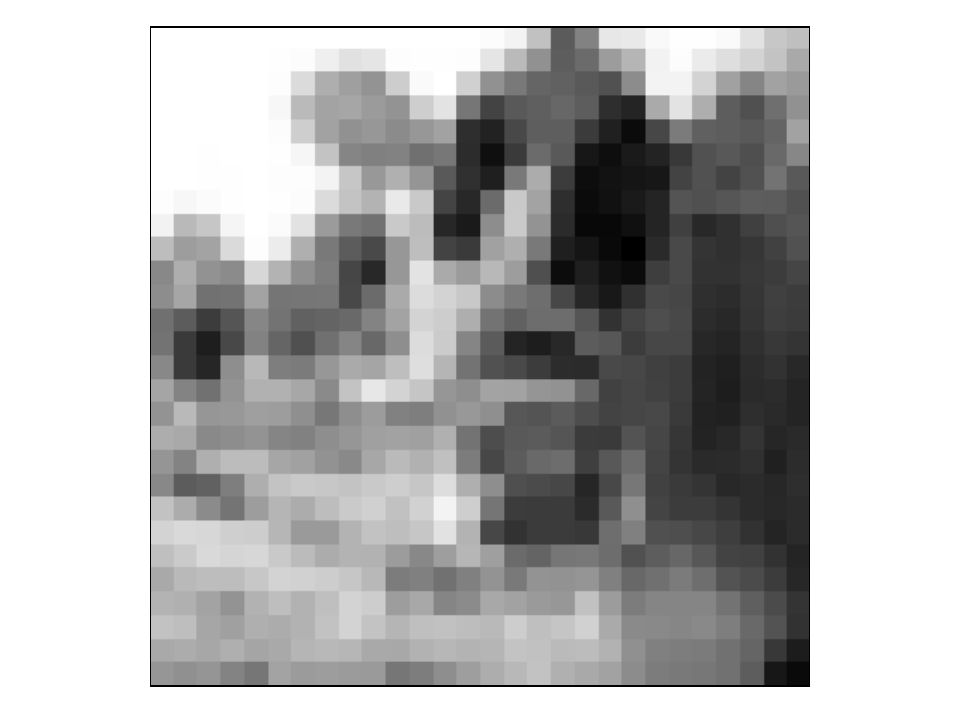}}
\subfigure["Dress"]{\label{fig:ex7-j}\includegraphics[height=1.3in,width=1.5in]{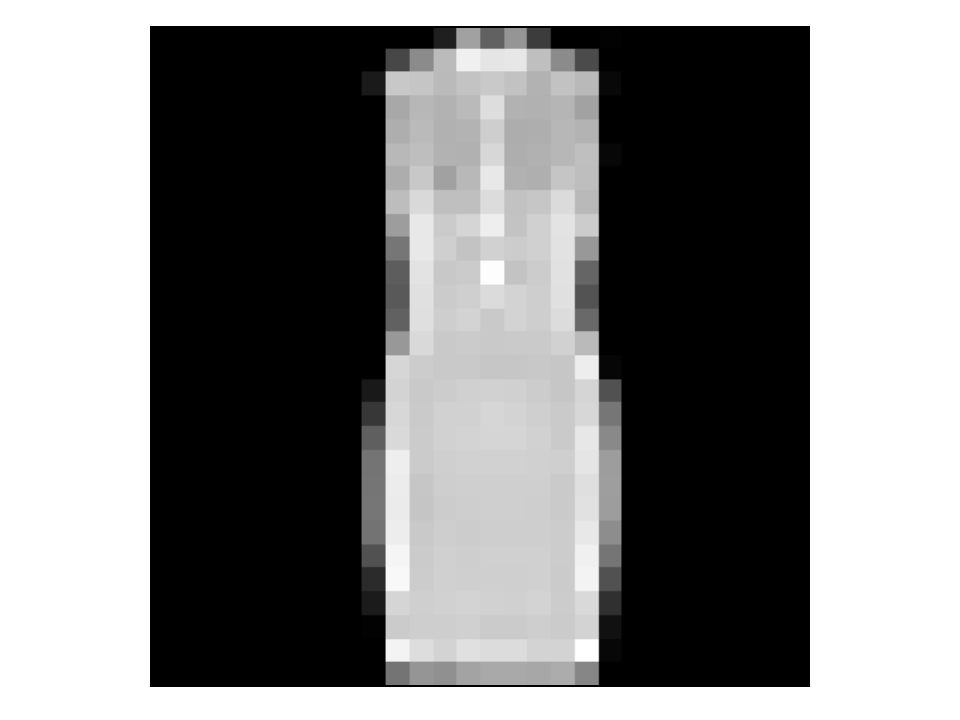}}
\subfigure["Cow" + "Dress"]{\label{fig:ex7-k}\includegraphics[height=1.3in,width=1.5in]{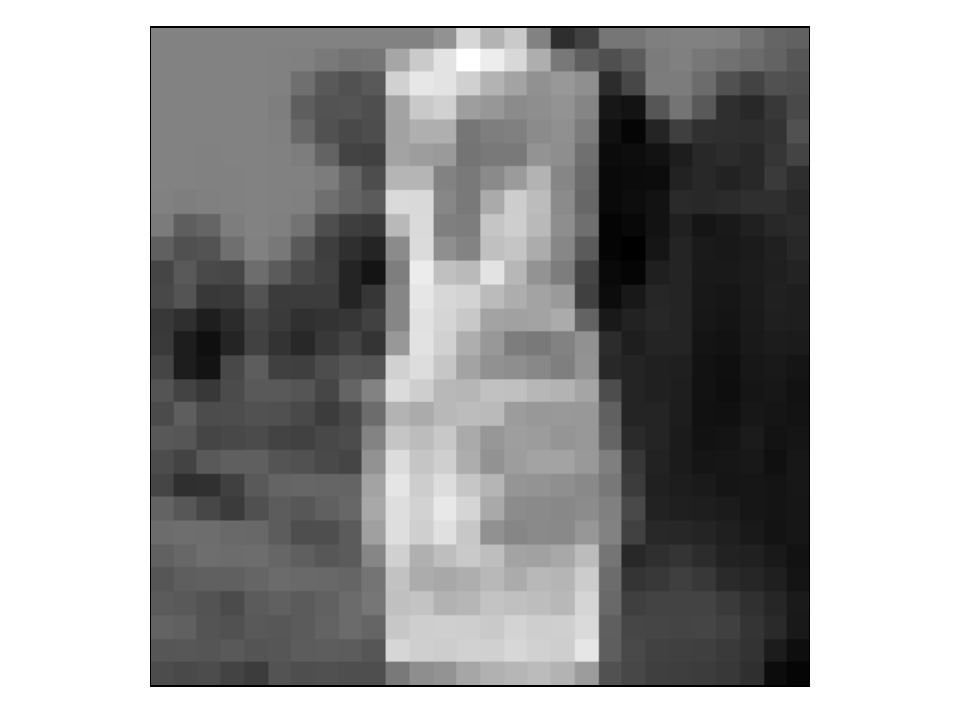}} \\
\subfigure["Bicycle"]{\label{fig:ex7-l}\includegraphics[height=1.3in,width=1.5in]{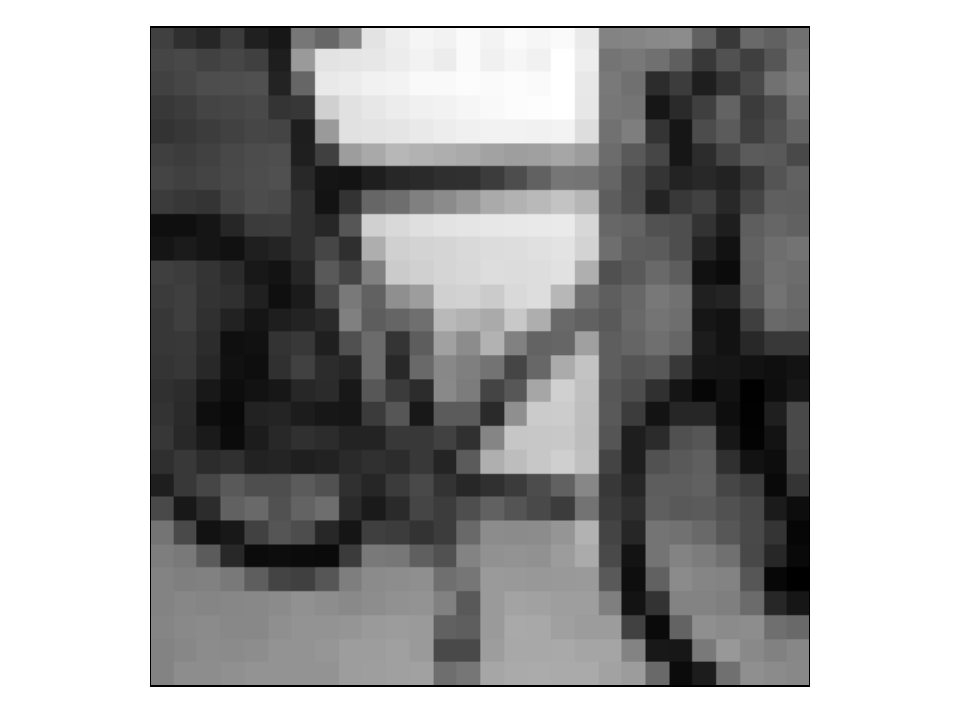}}
\subfigure["Trousers"]{\label{fig:ex7-m}\includegraphics[height=1.3in,width=1.5in]{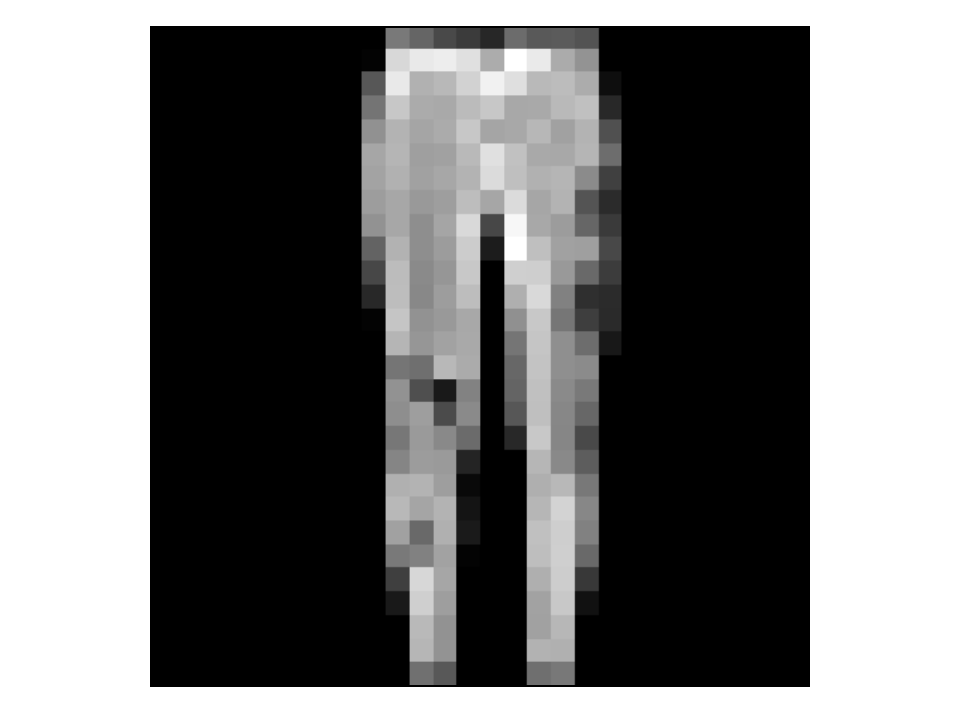}}
\subfigure["Bicycle" + "Trousers"]{\label{fig:ex7-n}\includegraphics[height=1.3in,width=1.5in]{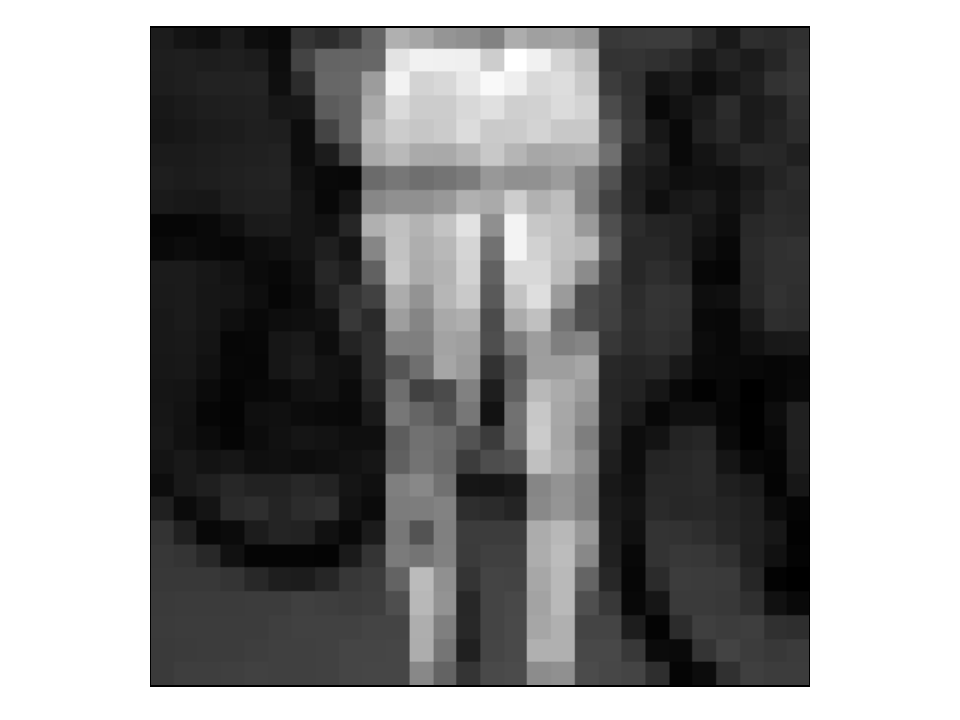}}
\caption{Superimposed CIFAR-100 + FMNIST embeddings computed by all methods for different combinations of FMNIST classes. (a), (b), (c), (d) for "T-shirt/Top"-"Shirt" case, (d), (e), (f), (g) for "T-shirt/Top"-"Coat" case, (i), (j) CIFAR-100 instances, (l), (m) FMNIST instances and (k), (n) their superimposition results.}
\label{fig:CIFAR-FMNIST-embeddings}
\end{figure*}

The generated complex data embeddings for both cases are illustrated in Figure \ref{fig:CIFAR-FMNIST-embeddings}. We observe that, cPCA completely fails to separate the embeddings with respect to their FMNIST class in both cases. While ct-SNE and Fair-NeRV illustrate some separation, there is considerable overlap of embeddings with different FMNIST classes. On the contrary, the respective segregation of the IMAPCE embeddings is evident.

To quantify the embeddings' separability according to their FMNIST class, we trained (75\% train, 25\% test) Linear and SVM classifiers over ten random train-test splits on the FMNIST classification of the 2D embeddings computed by each method. We report the average (over the ten random splits) test-set classification accuracy in Table \ref{table:cifar-fmnist-linear} for the Linear classifier and the average test-set classification accuracy, F1 and Recall in Table \ref{table:cifar-fmnist-svm} for the SVM classifier. Our visual observations are verified as IMAPCE achieves the best scores among all methods for both the linear and the SVM classifiers across all cases. This performance is boosted by the fact that it is achieved under an unsupervised setting by considering a few (5,000 out of the total 60,000) CIFAR-100 instances as background. On the contrary, both ct-SNE and Fair-NeRV need the specific CIFAR-100 labels of all the instances that were used when constructing the complex data. Finally, as also argued previously , the superior performance of IMAPCE is boosted by its computational efficiency.

\begin{table}[ht]
\caption{Test-set (averaged over ten random train-test splits) accuracy scores for Linear classification of the 2D complex CIFAR-100 + FMNIST embeddings of cPCA, ct-SNE, Fair-NeRV and IMAPCE with respect to their FMNIST classes.}
\begin{center}
\begin{tabular}{c c c c c} 
 \toprule
 FMNIST classes & cPCA & ct-SNE & Fair-NeRV & IMAPCE \\ 
 \midrule 
 Tshirt-Shirt & 0.54 & 0.70 & 0.71 & \textbf{0.78} \\ 
 \midrule 
 Tshirt-Coat & 0.64 & 0.86 & 0.87 & \textbf{0.90} \\
\bottomrule
\end{tabular}
\label{table:cifar-fmnist-linear}
\end{center}
\end{table}

\section{Iterative Exploration of MNIST} \label{Supp:Iter_Expl}
We performed the iterative visual exploration of 10,000 randomly selected MNIST samples using both IMAPCE and cPCA, identically to the UCI Image Segmentation dataset. We ran multiple experiments assuming no prior data as well as prior data coming from every ground truth class. Quantitative results are provided in Table \ref{table:MNIST} where IMAPCE outperforms cPCA with respect to both Jaccard and NMI under every prior data assumption.

\section{Ablation Study} \label{Supp:Abb}
To further examine the effect of the kurtosis term on the performance of IMAPCE, we ran several experiments using different $\mu$ (hyperparameter that controls the contribution of the kurtosis term along the optimization of the IMAPCE) values on the synthetic, UCI adult and both types of Complex data. 

For the synthetic data, we used the first four dimensions as prior (as we did in 4.1.1) and wanted to compute embeddings that do not exhibit any cluster formation according to the labels of those dimensions, while unveiling the label information of dimensions five and six. The first column of Table \ref{table:syn-adult} shows the NLS (averaged over 10, 20, \dots, 100 neighborhood sizes) of the IMAPCE embeddings computed with respect to the labels of dimensions one to four, for different $\mu$ values. The remaining columns refer to the attributes (and their combinations) that were used as prior when computing the IMAPCE embeddings for the UCI adult dataset. In this case, the NLS is computed using using the labels of the specific attributes as in 4.1.2 . 
\begin{table}[ht]
\caption{Jaccard (averaged over all classes) and NMI scores for evaluation of IMAPCE and cPCA on the exploration of 10,000 MNIST samples. Ten different random initializations were used for IMAPCE.}
\begin{center}
\begin{tabular}{c c c c} 
 \hline
 Method & Prior class & Jaccard  &  NMI \\ 
 \hline 
IMAPCE & - &$ \mathbf{0.41} \pm 0.01$ & $\mathbf{0.41} \pm 0.02$ \\
cPCA & - & 0.3 & 0.22 \\ 
\hline
IMAPCE & 0 &$ \mathbf{0.38} \pm 0.01$ & $\mathbf{0.39} \pm 0.03$ \\
cPCA & 0 & 0.21 & 0.14 \\ 
\hline
IMAPCE & 1 &$ \mathbf{0.33} \pm 0.02$ & $\mathbf{0.35} \pm 0.02$ \\
cPCA & 1 & 0.16 & 0.09 \\ 
\hline
IMAPCE & 2 &$ \mathbf{0.37} \pm 0.03$ & $\mathbf{0.38} \pm 0.02$ \\
cPCA & 2 & 0.31 & 0.22 \\ 
\hline
IMAPCE & 3 &$ \mathbf{0.34} \pm 0.02$  & $\mathbf{0.39} \pm 0.01$ \\
cPCA & 3 & 0.25 & 0.19 \\ 
\hline
IMAPCE & 4 &$ \mathbf{0.44} \pm 0.03$  & $\mathbf{0.42} \pm 0.03$ \\
cPCA & 4 & 0.27 & 0.19 \\ 
\hline
IMAPCE & 5 &$ \mathbf{0.43} \pm 0.03$  & $\mathbf{0.43} \pm 0.02$ \\
cPCA & 5 & 0.26 & 0.28 \\ 
\hline
IMAPCE & 6 &$ \mathbf{0.32} \pm 0.02$  & $\mathbf{0.35} \pm 0.02$ \\
cPCA & 6 & 0.26 &  0.2\\ 
\hline
IMAPCE & 7 &$ \mathbf{0.4} \pm 0.03$  & $\mathbf{0.38} \pm 0.02$ \\
cPCA & 7 & 0.24 & 0.2 \\ 
\hline
IMAPCE & 8 &$ \mathbf{0.41} \pm 0.03$  & $\mathbf{0.41} \pm 0.03$ \\
cPCA & 8 & 0.27 & 0.19 \\ 
\hline
IMAPCE & 9 &$ \mathbf{0.42} \pm 0.03$  & $\mathbf{0.39} \pm 0.04$ \\
cPCA & 9 & 0.27 & 0.19 \\ 
\hline
\end{tabular}
\label{table:MNIST}
\end{center}
\end{table}

Furthermore, we computed IMAPCE embeddings using different $\mu$ values for both types of Complex data in a similar way as in 4.2.1 and 4.2.2. The test-set accuracy, F1 and Recall scores of SVM classifiers trained on the classification of the 2D embeddings with respect to their FMNIST are reported in Table \ref{table:abb-mfmnist} for MNIST + FMNIST case and Table \ref{table:abb-cifar-fmnist} for CIFAR-100 + FMNIST case. 

We observe that for smaller and lower-dimensional datasets, such as the synthetic and the UCI adult, lower $\mu$ values achiever better NLS which are indicative of the desirable prior information removal. As $\mu$ starts increasing, NLS get worse, indicating that the prior information is not factored out to a good extent. We believe this is happening because the kurtosis term dominates the optimization in these cases. On the other hand, for larger and higher-dimensional datasets, such as both cases of Complex data, IMAPCE benefits from the use of larger $\mu$ values as better performance is achieved. 

Finally, our empirical suggestion for selecting $\mu$ achieves the best performance in the majority of datasets and prior knowledge types. We believe this happens because we select $\mu$ so that the kurtosis term contributes to the optimization of the complex loss but does not dominate it. As a result, both prior knowledge can be factored out (via the background variance-based loss part) to a good extent from the embeddings and meaningful separation, which reveals complementary structure, can take place (via the kurtosis term).

\begin{table*}[ht]
\caption{NLS (averaged over 10, 20, \dots, 100 neighborhood sizes) of the IMAPCE embeddings computed using different $\mu$ values. Dimensions one to four are used as prior for the synthetic data, while the respective attribute or combination of attributes as prior for the UCI adult data.}
\begin{center}
\begin{tabular}{c c c c c} 
 \toprule
 & Synthetic & Ethnicity & Gender & Ethnicity-Gender  \\ 
\midrule
$\mu=0$ & 0.44 & 0.20 & 0.37 & 0.47 \\ 
$\mu=10$ & 0.44 & 0.20 & \textbf{0.38} & \textbf{0.50} \\
$\mu=200$ & \textbf{0.45} & \textbf{0.21} & \textbf{0.38} & 0.49 \\
$\mu=1e3$ & 0.01 & 0.19 & 0.00 & 0.20 \\
$\mu=1e5$ & 0.00 & 0.17 & 0.00 & 0.17 \\ 
\bottomrule
\end{tabular}
\label{table:syn-adult}
\end{center}
\end{table*}

\begin{table*}[ht]
\caption{Test-set (averaged over ten random train-test splits) accuracy, F1 and Recall scores for SVM classification of the complex MNIST + FMNIST 2D embeddings computed by IMAPCE, cPCA, ct-SNE and FAIR-NeRV with respect to their FMNIST ground truth labels.}
\begin{center}
\begin{tabular}{c|c c c|c c c|c c c} 
 \toprule
 & & "Sandal"-"Sneaker" & &  & "Tshirt"-"Dress" &  & &  "Sandal"-"Ankle boot" &  \\ 
\midrule
Method & Acc. & F1 Score & Recall & Acc. & F1 Score & Recall & Acc. & F1 Score & Recall \\ 
\midrule
cPCA & 0.75 & 0.75 & 0.76 & 0.85 & 0.85 & 0.84 & 0.89 & 0.89 & 0.88  \\ 
ct-SNE & 0.77 & 0.76 & \textbf{0.91} & \textbf{0.90} & \textbf{0.92} & 0.90 & \textbf{0.92} & \textbf{0.92} & \textbf{0.92}  \\
Fair-NeRV & 0.80 & 0.80 & 0.78 & 0.88 & 0.88 & \textbf{0.91} & 0.82 & 0.82 & 0.83  \\
IMAPCE & \textbf{0.81} & \textbf{0.81} & 0.80 & 0.88 & 0.88 & \textbf{0.91} & 0.90 & 0.90 & 0.89\\ 
\bottomrule
\end{tabular}
\label{table:mnist-fmnist-svm}
\end{center}
\end{table*}

\begin{table*}[ht]
\caption{Test-set (averaged over ten random train-test splits) accuracy, F1 and Recall scores for SVM classification of the complex CIFAR-100 + FMNIST 2D embeddings computed by IMAPCE, cPCA, ct-SNE and FAIR-NeRV with respect to their FMNIST ground truth labels.}
\begin{center}
\begin{tabular}{c|c c c|c c c|c c c} 
 \toprule
 &  & "Tshirt"-"Shirt" &  & &  "Trousers"-"Dress" & & & "Tshirt"-"Coat" & \\ 
\midrule
Method & Acc. & F1 Score & Recall & Acc. & F1 Score & Recall & Acc. & F1 Score & Recall \\ 
\midrule
cPCA & 0.54 & 0.54 & 0.56 & 0.58 & 0.57 & 0.55 & 0.65 & 0.65 & 0.65   \\ 
ct-SNE & 0.73 & 0.73 & 0.66 & 0.90 & 0.90 & 0.96 & 0.88 & 0.88 & 0.84 \\
Fair-NeRV & 0.72 & 0.72 & 0.68 & 0.85 & 0.85 & 0.92 & 0.89 & 0.88 & 0.83  \\
IMAPCE & \textbf{0.79} & \textbf{0.79} & \textbf{0.73} & \textbf{0.96} & \textbf{0.96} & \textbf{0.98} & \textbf{0.93} & \textbf{0.93} & \textbf{0.94}  \\ 
\bottomrule
\end{tabular}
\label{table:cifar-fmnist-svm}
\end{center}
\end{table*}

\begin{table*}[ht]
\caption{Test-set (averaged over ten random train-test splits) accuracy, F1 and Recall scores for SVM classification of the complex MNIST + FMNIST 2D embeddings computed by IMAPCE using different $\mu$ values with respect to their FMNIST ground truth labels. * denotes the $\mu$ value we selected according to our tuning.}
\begin{center}
\begin{tabular}{c|c c c|c c c|c c c} 
 \toprule
 & & "Sandal"-"Sneaker" & &  & "Tshirt"-"Dress" &  & &  "Sandal"-"Ankle boot" &  \\ 
\midrule
& Acc. & F1 Score & Recall & Acc. & F1 Score & Recall & Acc. & F1 Score & Recall \\ 
\midrule
$\mu=0$ & 0.7 & 0.79 & 0.70 & 0.85 & 0.85 & 0.84 & 0.89 & 0.89 & 0.89   \\ 
$\mu=10$ & 0.75 & 0.75 & 0.81 & 0.85 & 0.85 & 0.84 & 0.89 & 0.89 & 0.89 \\
$\mu=1e3$ & 0.76 & 0.76 & 0.81 & 0.86 & 0.86 & 0.85 & 0.89 & 0.89 & 0.89 \\
$\mu=1e5$* & \textbf{0.81} & \textbf{0.81} & \textbf{0.80} & \textbf{0.88} & \textbf{0.88} & 0.91 & \textbf{0.90} & \textbf{0.90} & \textbf{0.89}  \\ 
$\mu=1e7$ & 0.74 & 0.74 & 0.77 & \textbf{0.88} & \textbf{0.88} & \textbf{0.94} & 0.84 & 0.84 & 0.87  \\ 
\bottomrule
\end{tabular}
\label{table:abb-mfmnist}
\end{center}
\end{table*}

\begin{table*}[ht]
\caption{Test-set (averaged over ten random train-test splits) accuracy, F1 and Recall scores for SVM classification of the complex CIFAR-100 + FMNIST 2D embeddings computed by IMAPCE using different $\mu$ values with respect to their FMNIST ground truth labels. * denotes the $\mu$ value we selected according to our tuning.}
\begin{center}
\begin{tabular}{c|c c c|c c c|c c c} 
 \toprule
 &  & "Tshirt"-"Shirt" &  & &  "Trousers"-"Dress" & & & "Tshirt"-"Coat" & \\ 
\midrule
 & Acc. & F1 Score & Recall & Acc. & F1 Score & Recall & Acc. & F1 Score & Recall \\ 
\midrule
$\mu=0$ & 0.79 & 0.79 & 0.70 & 0.74 & 0.73 & 0.79 & 0.90 & 0.90 & 0.88   \\ 
$\mu=10$ & 0.79 & 0.79 & 0.69 & 0.74 & 0.73 & 0.79 & 0.90 & 0.90 & 0.88 \\
$\mu=1e3$ & 0.79 & 0.79 & 0.70 & 0.74 & 0.74 & 0.79 & 0.90 & 0.90 & 0.88 \\
$\mu=1e5$ & \textbf{0.80} & \textbf{0.80} & 0.71 & 0.86 & 0.86 & 0.94 & 0.90 & 0.90 & 0.89  \\ 
$\mu=1e7$* & 0.79 & 0.79 &\textbf{0.73} & \textbf{0.96} & \textbf{0.96} & \textbf{0.98} & \textbf{0.93} & \textbf{0.93} & \textbf{0.94}  \\ 
\bottomrule
\end{tabular}
\label{table:abb-cifar-fmnist}
\end{center}
\end{table*}
\end{document}